\documentclass{article} % For LaTeX2e
\usepackage{arxiv_preprint,times}

\usepackage{amsmath,amsfonts,bm}

\def\eqref#1{equation~\ref{#1}}
\def\1{\bm{1}}

\DeclareMathAlphabet{\mathsfit}{\encodingdefault}{\sfdefault}{m}{sl}
\SetMathAlphabet{\mathsfit}{bold}{\encodingdefault}{\sfdefault}{bx}{n}

\usepackage[utf8]{inputenc}
\usepackage[T1]{fontenc}
\usepackage{algorithm}   % loads `float`; must precede hyperref
\usepackage{algorithmic}
\usepackage{hyperref}
\usepackage{url}
\usepackage{booktabs}
\usepackage{amsmath}
\usepackage{amssymb}
\usepackage{amsfonts}
\usepackage{amsthm}

\usepackage{graphicx}
\usepackage{wrapfig}
\usepackage{placeins}
\usepackage{adjustbox}
\usepackage{multirow}
\usepackage{makecell}
\usepackage{xcolor}
\usepackage{colortbl}
\definecolor{inkgreen}{RGB}{15,90,55}
\hypersetup{
  colorlinks=true,
  linkcolor=inkgreen,
  citecolor=inkgreen,
  urlcolor=inkgreen
}
\usepackage{pifont}
\usepackage{enumitem}
\usepackage[protrusion=true,expansion=false]{microtype}
\usepackage{oppo_report}

\definecolor{headergray}{RGB}{245,245,245}
\definecolor{oursrow}{RGB}{233,238,250}   % refined periwinkle: \method{} rows
\colorlet{rowblue}{oursrow}               % alias used by generated tables

\newcommand{\method}{TRACE}            % TRajectory-robust Admission and Coverage-aware Evidence ordering
\newcommand{\lmp}{LIP}                 % Layout-derived Interaction Prior
\newcommand{\nwe}{NEO}                 % Nested Evidence Ordering
\newcommand{\awc}{NCR}                 % Native-token Coverage Repair
\newcommand{\kvr}{MKC}                 % Monotone KV Contraction (KV-cache realization of the reusable visual-state lifecycle)

\newcommand{\rhocur}{\rho_{\mathrm{cur}}}
\newcommand{\rhohist}{\rho_{\mathrm{hist}}}
\newcommand{\keepc}{c}                 % current-frame keep ratio
\newcommand{\keeph}{h}                 % history keep ratio
\newcommand{\keepr}{r}                 % single-step keep ratio

\newcommand{\pub}[1]{$_{\text{#1}}$}   % venue subscript after method names
\newcommand{\cmark}{\ding{51}}
\newcommand{\xmark}{\ding{55}}
\newcommand{\pmark}{$\triangle$}       % satisfiable only with modification

\newcommand{\cnum}[1]{\raisebox{-0.2ex}{\scalebox{1.35}{\textbf{\ding{\numexpr171+#1\relax}}}}}

\title{TRACE: Trajectory-robust Admission with Evidence Ordering for Efficient GUI Agents}

\author{%
Yuhao Wang$^{1}$,\
Mu Qiao$^{1}$,\
Xindong Zhang$^{2}$,\
Yunzhi Zhuge$^{1}$,\
Lei Zhang$^{2,3}$,\
Huchuan Lu$^{1}$\\[0.4em]
{\mdseries\footnotesize\makebox[\textwidth][c]{$^{1}$Dalian University of Technology, $^{2}$OPPO Research Institute, $^{3}$The Hong Kong Polytechnic University}}%
}

\begin{document} 

\maketitle

\begin{abstract}
GUI agents accumulate high-resolution screenshots as the trajectory unfolds, increasing inference latency and memory usage.
Training-free visual token pruning can reduce this cost, but cache reuse introduces a fundamental constraint.
Once tokens are discarded, the corresponding visual evidence cannot be recovered without re-encoding.
Pruning therefore becomes an \textit{irreversible admission decision} that must remain useful for unknown future targets while preserving coverage of operable regions under tight budgets.
To address these challenges, we propose \textbf{\method{}}, a training-free framework for \emph{\textbf{T}rajectory-\textbf{r}obust \textbf{A}dmission and \textbf{C}overage-aware \textbf{E}vidence ordering}.
Specifically, we combine a query-independent layout-derived interaction prior with instruction relevance and feature novelty to rank visual evidence according to both potential future utility and diversity.
Then, we reserve part of the budget for native visual tokens distributed across the screen, repairing missing spatial coverage without breaking the ordering.
Together, these mechanisms produce a nested token order, allowing retained visual evidence to shrink monotonically across budgets while remaining reusable throughout the trajectory.
Finally, our monotone KV contraction incrementally contracts retired frames into compact session state, avoiding repeated visual encoding or pruning.
Extensive experiments across six GUI benchmarks and diverse models verify the effectiveness of our proposed \method{} under tight budgets.
The source code will be released.
\end{abstract}

\section{Introduction}
\label{sec:intro}

\begin{figure}[t]
    \centering
    \includegraphics[width=0.99\linewidth]{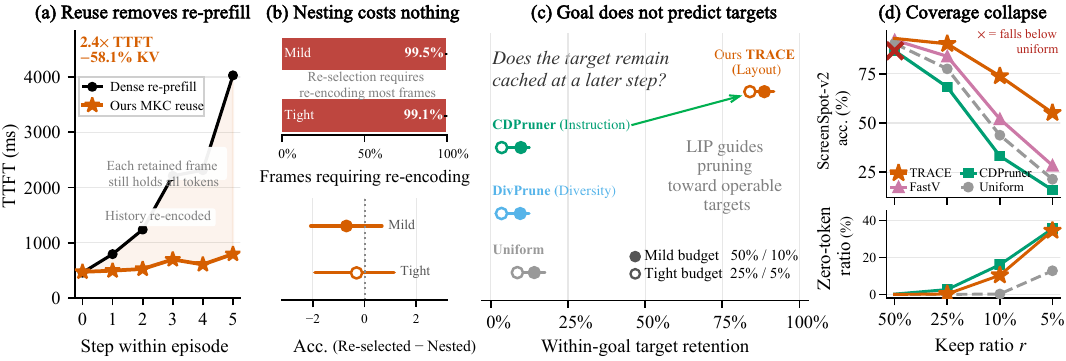}
    \vspace{-0mm}
    \caption{%
    \textbf{Motivation of our proposed TRACE.} %
    \textbf{(a)}~Reusing the cache removes repeated prefill and lowers latency. %
    \textbf{(b)}~Independent re-selection demands re-encoding on most frames, while the nested order costs no accuracy. %
    \textbf{(c)}~Instruction-based pruning misses the target in later steps while layout-based pruning retains it. %
    \textbf{(d)}~Concentrating the budget on salient regions loses spatial coverage.}
    \label{fig:evidence}
\end{figure}

Multimodal large language models (MLLMs) are increasingly deployed as agents that complete multi-step tasks by perceiving and acting on graphical user interfaces (GUIs)~\citep{cheng2024seeclick,guiowl2025}.
At each step, a GUI agent processes the high-resolution screenshot together with context accumulated from earlier interactions~\citep{tian2025mmina}.
As the trajectory unfolds, visual computation grows to dominate the inference cost.
Each historical screenshot carries a large number of visual tokens and recurs across successive steps, inflating both computation and memory~\citep{xie2024osworld}.
As shown in Fig.~\ref{fig:evidence} (a), cache reuse mitigates the temporal redundancy across frames by limiting past computation~\citep{zheng2024sglang}.
However, this mechanism operates at frame granularity and leaves substantial spatial redundancy within each retained screenshot.
The core problem therefore becomes deciding which visual tokens enter the reusable state of each frame.

Training-free visual token pruning has emerged as an effective way to reduce this redundancy.
For example, FastV~\citep{chen2024fastv} prunes tokens by attention distribution.
DivPrune~\citep{alvar2025divprune} and CDPruner~\citep{zhang2025cdpruner} select for feature diversity or instruction relevance.
Besides, PruMerge~\citep{shang2025prumerge} merges redundancy into synthetic summaries.
These approaches are effective because the selection can be recomputed whenever the query changes.
Multi-step GUI serving breaks this premise once visual state becomes reusable, because each screenshot is encoded and written into the session cache only at its first prefill.
Later steps read history from that cache rather than pixels, so tokens dropped at the single write are absent for every future query and can only be recovered by re-encoding the frame.
\textbf{Visual token pruning thus shifts from revisable selection to irreversible write-time commitment of visual evidence.}
Under this commitment, each frame should be admitted once before prefill, reused while current, and contracted when it becomes history.
Contraction can only drop tokens from the committed state, so the historical keep sets should nest inside the current one.
Pruning therefore must output a nested order whose prefixes realize every budget.
As shown in Fig.~\ref{fig:evidence} (b), re-selection demands re-encoding on most frames, whereas enforcing the nested order costs no accuracy.
We term this regime \emph{lifecycle-aware visual pruning} and formulate its decision as \emph{write-time visual evidence commitment under trajectory uncertainty}.

This formulation exposes two distinct challenges.
The first is trajectory uncertainty.
The task goal is known at admission, yet the fine-grained targets along the trajectory remain unknown.
Write-time admission must therefore commit evidence before those later targets appear.
Among the signals available at that moment, feature diversity is query-independent yet does not separate operable regions from background.
By contrast, the interface layout offers operable regions before any demand is observed.
As shown in Fig.~\ref{fig:evidence} (c), when a screen recurs under the same goal, instruction-selected tokens miss the later target, whereas layout-selected tokens still cover it.
Beyond present-step cues, trajectory-robust admission should therefore also incorporate this layout prior.
Independently, the second challenge is spatial coverage under tight budgets.
Biased token pruning concentrates on a few salient regions, leaving other operable regions with zero support and thus unavailable to the agent.
As shown in Fig.~\ref{fig:evidence} (d), such biased keeps leave a growing uncovered area and eventually fall below uniform sampling as the budget tightens~\citep{deng2025scope,xu2026score}.
Beyond importance concentration, tight-budget admission must therefore also preserve spatial coverage.

Motivated by these observations, we propose \textbf{\method{}}, a \emph{\textbf{T}rajectory-\textbf{r}obust \textbf{A}dmission and \textbf{C}overage-aware \textbf{E}vidence ordering} framework for efficient GUI agents.
Specifically, \method{} comprises four key components: \textbf{Layout-derived Interaction Prior (\lmp{})}, \textbf{Nested Evidence Ordering (\nwe{})}, \textbf{Native-token Coverage Repair (\awc{})}, and \textbf{Monotone KV Contraction (\kvr{})}.
First, \lmp{} maps layout detections into an interaction prior, enabling admission to favor operable regions before later targets appear.
Second, \nwe{} fuses this prior with instruction relevance and feature novelty into one nested order, so every tighter budget is realized as a prefix rather than by re-selection.
Third, \awc{} restores spatial coverage with native tokens while preserving that nested order.
Finally, \kvr{} contracts each retiring frame to its history prefix and replays the retained native rows in the next prefill.
Together, these components commit visual evidence once at admission and contract it monotonically thereafter.
In this way, \method{} realizes lifecycle-aware visual pruning under trajectory uncertainty for efficient GUI agents.
Extensive experiments across six GUI benchmarks covering single-step and multi-step settings, including multiple model scales and a cross-family backbone, verify the effectiveness of \method{}.
Overall, our contributions can be summarized as follows.
\begin{itemize}
\item We formulate \emph{lifecycle-aware visual pruning} as \emph{write-time visual evidence commitment under trajectory uncertainty}. Once visual state is reusable, pruning becomes an irreversible admission with nested keep sets across budgets, exposing two challenges of committing evidence before later targets appear and preserving spatial coverage under tight budgets.
\item We propose \method{}, a training-free framework that commits visual evidence once and contracts it monotonically thereafter. \lmp{} injects a query-independent layout prior, \nwe{} builds one nested evidence order, \awc{} restores native-token spatial coverage, and \kvr{} turns the admitted order into reusable session state without re-encoding historical frames.
\item Extensive experiments across six GUI benchmarks verify the effectiveness of \method{} under single-step and multi-step settings, multiple model scales, and a cross-family backbone.
\end{itemize}

\section{Related Work}
\label{sec:related}
\subsection{Efficient GUI Agents}
GUI agents ground actions in an accumulating stream of high-resolution screenshots~\citep{hong2024cogagent,guiowl2025}, inflating both computation and memory.
Existing methods fall into \textit{training-based designs} and \textit{training-free designs}.
\textbf{Training-based Designs.}
This paradigm learns cheaper perception by retraining the architecture or the history policy.
For example, CogAgent~\citep{hong2024cogagent} couples a low-resolution backbone with a high-resolution cross-attention module.
ReVision~\citep{abaskohi2026revision} learns to cut temporal visual redundancy along the trajectory.
While effective, these designs demand extra training compute, and the learned efficiency cannot transfer across architectures.
These costs motivate training-free designs on frozen models.
\textbf{Training-free Designs.}
Training-free methods cut cost on a frozen model via \textit{input-side} pruning or \textit{cache-side} compression.
On the \textit{input side}, AQuaUI~\citep{li2026aquaui} partitions screenshots with adaptive quadtrees, and related methods prune high-resolution screens or historical frames by spatio-temporal cues~\citep{xu2026guipruner,histprune2026}.
On the \textit{cache side}, GUI-KV~\citep{huang2025guikv} combines spatial saliency with temporal redundancy scoring.
ST-Lite~\citep{zhou2026stlite} couples component-centric saliency with trajectory-aware gating.
Despite these improvements, keeps are re-scored at every step, so adjacent keeps need not nest and a retired frame must be re-encoded or re-ranked.
In contrast, \method{} admits a nested keep once and contracts it as monotone session state efficiently.
\subsection{Visual Token Pruning}
Visual token pruning accelerates MLLM inference by removing redundant visual tokens, and existing methods either discard native tokens directly or aggregate them into synthetic ones.
\textbf{Pruning-based Methods.}
FastV~\citep{chen2024fastv} ranks visual tokens by attention statistics inside the language model.
DivPrune~\citep{alvar2025divprune} emphasizes feature dispersion to reduce redundancy.
CDPruner~\citep{zhang2025cdpruner} selects tokens that are both diverse and relevant to the instruction.
Nevertheless, selection discards tokens irreversibly, which motivates token merging as an alternative paradigm.
\textbf{Merging-based Methods.}
This paradigm aggregates discarded patches into compact synthetic summaries.
VisionZip~\citep{yang2025visionzip} merges visual tokens to extend the effective context length.
PruMerge+~\citep{shang2025prumerge} combines pruning with merging to adapt the visual token count.
Although effective, the selections of both paradigms drift across steps and demand rows that a contracted cache no longer holds.
Moreover, neither is tailored to GUI streams, where score-driven keeps collapse onto a few salient regions and sacrifice the spatial coverage that dense interfaces require.
In contrast, \method{} injects an interaction prior before prefill, repairs the spatial collapse at admission, and contracts the admitted keep into monotone session state for efficient serving.

\section{Method}
\label{sec:method}

\begin{figure*}[t]
\centering
\includegraphics[width=1\linewidth]{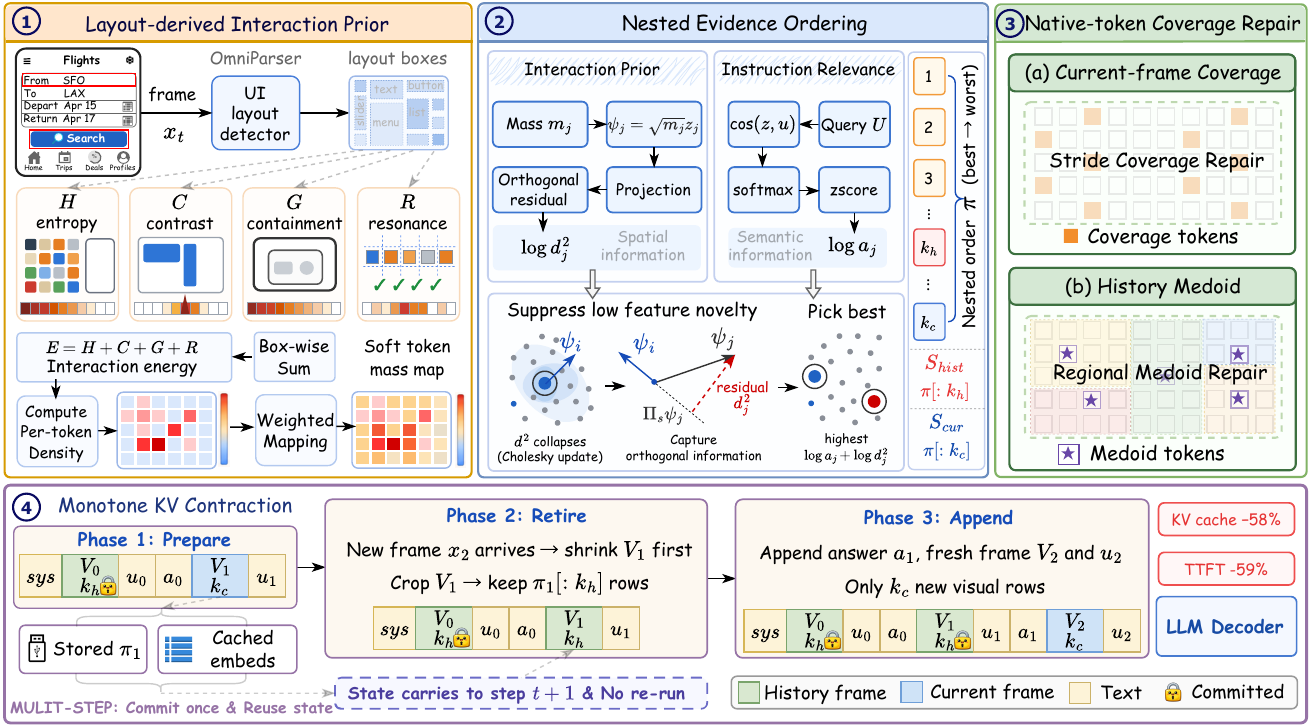}
\vspace{-5mm}
\caption{\textbf{Overview of the proposed \method{} framework.}
\textbf{(a)~Layout-derived Interaction Prior} maps UI detections into an interaction prior.
\textbf{(b)~Nested Evidence Ordering} fuses this prior with instruction relevance and feature novelty into one nested order.
\textbf{(c)~Native-token Coverage Repair} refills the spatial gaps with native coverage and medoid tokens.
\textbf{(d)~Monotone KV Contraction} crops each retiring frame to its history prefix and restores retained rows in the next prefill.
Together, they enable trajectory-robust admission with evidence ordering for efficient GUI agents.}
\label{fig:overview}
\vspace{-2mm}
\end{figure*}

\subsection{Problem Formulation}
\label{sec:method-setup}

In multi-step GUI interaction, an agent observes a growing sequence of screenshots as an episode unfolds. 
At step $t$, the agent's visual encoder maps screenshot $x_t$ into $N$ raster-ordered tokens $\mathbf{E}_t\in\mathbb{R}^{N\times D}$, where $D$ is the feature dimension.
Under the stateless serving paradigm~\citep{li2026aquaui}, the model repeatedly prefills the current frame together with up to $H$ historical frames, which inflates both latency and memory.
To avoid this cost, we propose a reusable visual-state lifecycle in which each frame is encoded once and only contracts thereafter, so the retained positions must be decided before the first LLM prefill.
At this stage, the selector operates on the non-visual prefix $T_t$, the current visual tokens $\mathbf{E}_t$, and the admitted history, denoted collectively by $\mathcal{A}_t$.
For each frame $t$, let $S_t^{\mathrm{cur}}$ denote the positions retained while the frame is current, and let $S_t^{\mathrm{hist}}$ denote the smaller subset retained after retirement.
Formally, the reusable lifecycle imposes three constraints:
\begin{equation}
\text{(i)}\ S_t^{\mathrm{cur}}=\sigma(\mathcal{A}_t),\quad
\text{(ii)}\ S_t^{\mathrm{hist}}\subseteq S_t^{\mathrm{cur}},\quad
\text{(iii)}\ S_t^{\mathrm{cur}},S_t^{\mathrm{hist}}\subseteq\{1,\ldots,N\}.
\label{eq:lifecycle}
\end{equation}
Here, $\sigma$ denotes the selector used before prefill.
Condition (i) confines pruning to information available at this stage, condition (ii) enforces monotone retirement, and condition (iii) preserves original tokens to avoid synthetic merges that cannot later be dropped as cache rows.
These constraints make reuse realizable, yet leave open which tokens to admit while future utility remains unknown.
We therefore ground the evidence in the interface layout, whose operable regions are already visible.

\subsection{Layout-derived Interaction Prior}
\label{sec:method-a}
Relying solely on instruction relevance~\citep{zhang2025cdpruner} cannot anticipate later step demands, while uniform geometry~\citep{deng2025scope} does not separate operable regions from background.
To address this issue, we introduce the Layout-derived Interaction Prior (\lmp{}), which maps layout detections such as buttons and text fields into an interaction prior for the subsequent token pruning.
\\
\noindent\textbf{From Detections to Interaction Density.}
Specifically, to strengthen understanding of dense GUI screens, we detect layout boxes with OmniParser~\citep{lu2024omniparser}.
As illustrated by step~\cnum{1} in Fig.~\ref{fig:overview}, we compute entropy~$H$, contrast~$C$, containment~$G$ and resonance~$R$ for each box and form the interaction energy $E=H+C+G+R$.
Dividing by the number $n$ of active UI tokens covered by the box yields a per-token density $\rho=E/n$.
We then scatter $\rho$ onto the visual-token grid and max-normalize overlaps into a field $P$.
Then, we set $p=P/\lVert P\rVert_1$ so that only relative spatial shares enter the mapping below.
\noindent\textbf{From Density to Interaction Prior.}
The density $p$ marks where layout structure lies, but layout alone cannot decide admission under later steps.
A keep ranking from $p$ would over-commit to detections and shut out tokens that later relevance or coverage may still need.
We therefore turn $p$ into a soft token mass $m_j=1+\alpha Np_j$, which biases the subsequent ordering toward layout regions while keeping every token eligible.
Here, $\alpha\geq 0$ controls the strength of the bias, and this mass is the interaction prior passed downstream.
In this way, \lmp{} can inject the layout preference for the next stage to balance against instruction relevance and feature novelty.

\subsection{Nested Evidence Ordering}
\label{sec:method-b}
Pure relevance concentrates the budget on the strongest instruction matches~\citep{zhang2025cdpruner}, whereas diversity alone overlooks structurally important regions.
We therefore introduce Nested Evidence Ordering (\nwe{}), which fuses the interaction prior with instruction relevance and feature novelty into a nested order.
Then, every later budget can be extracted from the order's prefixes.
\\
\noindent\textbf{Relevance and Prior-weighted Features.}
Specifically, for each visual token $\mathbf e_j$, we define the normalized feature $\mathbf z_j=\mathbf e_j/\lVert\mathbf e_j\rVert_2$.
Besides, we normalize the instruction's token embeddings and prepend their mean to form the query matrix $\mathbf U$.
Then, we compute the relevance weight as follows:
\begin{equation}
    a_j =
    \mathrm{softmax}_j\!\left(
    \mathrm{zscore}\!\left(
    \max_{\mathbf u\in\mathbf U}\cos(\mathbf z_j,\mathbf u)
    \right)\right).
    \label{eq:likelihood}
\end{equation}
Here, the max keeps the strongest local match and the prepended mean anchors the score globally.
Besides the z-score, we also apply softmax to turn similarities into positive relative weights.
With the relevance in place, we next reweight each normalized feature with the interaction prior as $\boldsymbol\psi_j=\sqrt{m_j}\,\mathbf z_j$.
Here, layout-dense tokens matter more in the residual geometry, while the unit baseline in $m_j$ keeps every token eligible.
\noindent\textbf{Greedy Residual Ordering.}
To form a nested keep order, we apply the greedy relevance-weighted orthogonal residual selection with the above relevance and prior-weighted features.
To be specific, given a selected set $S$, it chooses the next token as follows:
\begin{equation}
j^\ast=\arg\max_{j\notin S}\bigl(\log a_j+\log d_j^2(S)\bigr),\qquad
d_j^2(S)=\lVert\boldsymbol\psi_j-\Pi_S\boldsymbol\psi_j\rVert_2^2 .
\label{eq:qd}
\end{equation}
Here, $\Pi_S$ is the orthogonal projector onto the features already in $S$.
With $\Psi_S=[\boldsymbol\psi_i]_{i\in S}$, we set $\Pi_S=\Psi_S(\Psi_S^{\top}\Psi_S)^{-1}\Psi_S^{\top}$ when $S$ is nonempty and $\Pi_\emptyset=\mathbf 0$.
The residual $d_j^2(S)$ then measures how much of $\boldsymbol\psi_j$ remains novel.
Adding $\log a_j$ balances that novelty against instruction relevance.
As illustrated by step~\cnum{2} in Fig.~\ref{fig:overview}, repeating this update produces a nested order $\pi=(\pi_1,\ldots,\pi_k)$:
\begin{equation}
S^{(k')}=\{\pi_1,\ldots,\pi_{k'}\}\subset S^{(k)}\quad\text{for }k'<k.
\label{eq:prefix}
\end{equation}
Retirement can therefore delete a suffix of $\pi$ rather than reselect the frame.
However, this order still leaves spatial placement uncontrolled.
A tight prefix may leave an entire region without any kept token.
Thus, the next stage restores the spatial coverage while preserving nesting and native tokens.

\subsection{Native-token Coverage Repair}
\label{sec:method-c}
Biased token pruning concentrates on a few strong regions and leaves other areas empty, whereas a sparse uniform lattice only partially restores the coverage~\citep{xu2026score}.
We therefore introduce Native-token Coverage Repair (\awc{}), which restores the spatial coverage with native tokens.
As shown by step~\cnum{3} in Fig.~\ref{fig:overview}, \awc{} selects different types of tokens for the current frame and history.
\\
\noindent\textbf{Current-frame Coverage Tokens.}
Specifically, the current frame uses the larger budget $k_c$, so repair can spread a few positions across the screen while keeping the strongest ordered evidence.
Concretely, with current dose $\rhocur\in[0,1]$, \awc{} selects $g_c=\lceil\rhocur k_c\rceil$ coverage tokens, protects the prefix $P_c=\{\pi_1,\ldots,\pi_{k_c-g_c}\}$, and takes the raster-ordered complement $U_c=(u_{(1)},\ldots,u_{(|U_c|)})$.
It then draws those coverage tokens from the complement by stride sampling with stride $s=\lceil|U_c|/g_c\rceil$.
These coverage tokens replace the unprotected tail and update $\pi$.
Retirement later shrinks the same frame to $k_h$, so history coverage is restored with medoid tokens rather than another stride sample.
\noindent\textbf{History Medoid Tokens.}
For history, with history dose $\rhohist\in[0,1]$, \awc{} sets $g_h=\lceil\rhohist k_h\rceil$ and protects a prefix of length $k_h-g_h$.
It then partitions the remaining space into $g_h$ near-equal regions.
From each region $R_b$, \awc{} keeps the medoid token $\mu_b$, which is the native token that minimizes the sum of squared feature distances to the other tokens in $R_b$.
The protected prefix and these medoid tokens jointly define the history subset, which stays nested inside the current keep.
In this way, \awc{} effectively restores spatial coverage while leaving the reusable cache realization to the next stage.

\subsection{Monotone KV Contraction}
\label{sec:method-kv}
Admission yields its advantage only if serving contracts a retired frame into the next prefill without a second visual encoding.
We therefore introduce Monotone KV Contraction (\kvr{}), which realizes the reusable visual-state lifecycle.
As illustrated by step~\cnum{4} in Fig.~\ref{fig:overview}, each step advances this state through three phases, shown here from past frame $V_0$ and current frame $V_1$.
\noindent\textbf{Phase 1. Prepare.}
The session holds the system prompt, past frame $V_0$ at budget $k_h$, the preceding text and answer $u_0$ and $a_0$, current frame $V_1$ at budget $k_c$, and instruction $u_1$.
Admission has already stored the nested order $\pi_1$ of $V_1$ and its native visual embeddings.
\noindent\textbf{Phase 2. Retire.}
When a new screenshot $x_2$ arrives, \kvr{} first shrinks $V_1$ before any append.
It truncates the cache at the boundary of $V_1$ and crops the stored embeddings to the first $k_h$ entries of $\pi_1$.
The operation is indexing alone and incurs no extra computation.
\noindent\textbf{Phase 3. Append.}
\kvr{} then restores those cropped rows with answer $a_1$, fresh frame $V_2$ and instruction $u_2$ in one merged LLM forward pass.
The restored rows reuse embeddings encoded when $V_1$ arrived, whereas only $V_2$ is newly encoded at budget $k_c$.
The resulting state leaves $V_0$ unchanged, places $V_1$ at history budget $k_h$, and makes $V_2$ the new current frame.
In this way, \kvr{} turns the admitted order into monotone session state.
It keeps the evidence native and encodes only the fresh frame at each step.
Overall, these components ensure efficient and robust GUI agents.

\section{Experiments}
\label{sec:exp}
\begin{table}[t]
    \centering
    \caption{Performance comparison with \textbf{GUI-Owl-1.5} series. \textbf{Avg.\,(\%)} denotes performance relative to the upper bound. Best and second-best results are shown in \textbf{bold} and \underline{underlined}, respectively.}
    \label{tab:main-combined}
    \resizebox{\textwidth}{!}{%
        \setlength{\tabcolsep}{6pt}%
        \renewcommand{\arraystretch}{0.92}%
        \begin{tabular}{l ccc ccc c}
            \toprule
            \multirow{2}{*}{\textbf{Method}} & \multicolumn{3}{c}{\textbf{Single-step (Accuracy)}} & \multicolumn{3}{c}{\textbf{Multi-step (Step~SR)}} & \multirow{2}{*}{\textbf{Avg.\,(\%)}} \\
            \cmidrule(lr){2-4}\cmidrule(lr){5-7}
            & SS-v2 & SS-Pro & MMBench-GUI & OmniGUI & Mind2Web & AndroidControl &  \\
            \midrule
            \rowcolor{headergray}\multicolumn{8}{c}{\textsc{GUI-Owl-1.5-8B: Upper Bound (100\% Tokens)}} \\
            GUI-Owl-1.5-8B & 93.79 & 70.34 & 82.64 & 52.45 & 52.90 & 60.41 & 100.0\% \\
            \rowcolor{headergray}\multicolumn{8}{c}{\textsc{Mild Budget (Single-step: $\keepr{=}10\%$ \,$||$\, Multi-step: $\keepc{=}50\%$, $\keeph{=}10\%$)}} \\
            Random & 33.81 & 6.14 & 18.53 & 43.16 & 35.13 & 58.82 & 52.2\% \\
            Uniform & 43.79 & 10.69 & 28.27 & 46.11 & 39.65 & 59.08 & 59.5\% \\
            DivPrune\pub{(CVPR25)\cite{alvar2025divprune}} & 51.97 & 13.98 & 31.61 & 45.76 & 40.70 & 58.83 & 62.5\% \\
            CDPruner\pub{(NeurIPS25)\cite{zhang2025cdpruner}} & 33.18 & 11.26 & 12.21 & 41.45 & 27.88 & 54.57 & 48.0\% \\
            VisPruner\pub{(ICCV25)\cite{zhang2025vispruner}} & 51.73 & 34.41 & \underline{42.35} & 46.11 & 44.45 & 59.25 & 70.9\% \\
            PruMerge+\pub{(ICCV25)\cite{shang2025prumerge}} & 40.49 & 19.80 & 30.47 & 46.50 & 41.53 & 59.28 & 62.2\% \\
            TRIM\pub{(COLING25)\cite{song2025trim}} & 36.24 & 3.29 & 25.24 & 45.84 & 41.20 & 56.86 & 55.5\% \\
            FastV\pub{(ECCV24)\cite{chen2024fastv}} & 52.04 & 28.53 & 40.21 & 46.89 & 43.12 & 59.11 & 68.9\% \\
            VisionTrim\pub{(ICLR26)\cite{yu2026visiontrim}} & 58.65 & \textbf{38.52} & 40.71 & 45.72 & 43.91 & 59.10 & \underline{72.4\%} \\
            ZOO-Prune\pub{(CVPR26)\cite{kim2026zooprune}} & 56.29 & 9.55 & 35.98 & \underline{48.17} & \underline{44.98} & 59.30 & 65.4\% \\
            PruneSID\pub{(ICLR26)\cite{fang2026prunesid}} & \underline{58.96} & 15.69 & 40.65 & 47.40 & 43.28 & \underline{59.52} & 67.5\% \\
            \rowcolor{rowblue}\textbf{\method{}} & \textbf{73.90} & \underline{37.63} & \textbf{50.47} & \textbf{48.91} & \textbf{45.52} & \textbf{60.20} & \textbf{78.7\%} \\
            \rowcolor{headergray}\multicolumn{8}{c}{\textsc{Tight Budget (Single-step: $\keepr{=}5\%$ \,$||$\, Multi-step: $\keepc{=}25\%$, $\keeph{=}5\%$)}} \\
            Random & 15.49 & 1.27 & 9.29 & 34.25 & 18.43 & 52.07 & 36.0\% \\
            Uniform & 21.38 & 3.04 & 12.24 & 38.96 & 20.61 & \underline{56.13} & 41.3\% \\
            DivPrune\pub{(CVPR25)\cite{alvar2025divprune}} & 25.71 & 3.86 & 14.11 & 37.17 & 24.29 & 54.75 & 42.9\% \\
            CDPruner\pub{(NeurIPS25)\cite{zhang2025cdpruner}} & 15.88 & 3.48 & 4.76 & 27.76 & 10.86 & 40.90 & 28.1\% \\
            VisPruner\pub{(ICCV25)\cite{zhang2025vispruner}} & 20.68 & 12.21 & 16.75 & 39.54 & \underline{30.67} & 54.93 & 47.3\% \\
            PruMerge+\pub{(ICCV25)\cite{shang2025prumerge}} & 17.85 & 5.88 & 12.19 & 36.98 & 23.45 & 54.01 & 41.1\% \\
            TRIM\pub{(COLING25)\cite{song2025trim}} & 15.88 & 0.44 & 9.85 & 34.37 & 23.01 & 47.17 & 36.1\% \\
            FastV\pub{(ECCV24)\cite{chen2024fastv}} & 28.38 & 11.83 & \underline{20.51} & 39.39 & 29.33 & 54.61 & 48.8\% \\
            VisionTrim\pub{(ICLR26)\cite{yu2026visiontrim}} & 25.86 & \underline{18.22} & 16.86 & 38.26 & 30.20 & 55.16 & \underline{49.2\%} \\
            ZOO-Prune\pub{(CVPR26)\cite{kim2026zooprune}} & 28.46 & 3.48 & 15.33 & 40.01 & 27.90 & 55.76 & 45.9\% \\
            PruneSID\pub{(ICLR26)\cite{fang2026prunesid}} & \underline{34.20} & 4.55 & 19.48 & \underline{40.09} & 27.57 & 55.46 & 47.8\% \\
            \rowcolor{rowblue}\textbf{\method{}} & \textbf{55.19} & \textbf{20.30} & \textbf{30.88} & \textbf{43.08} & \textbf{34.20} & \textbf{57.06} & \textbf{61.1\%} \\
            \rowcolor{headergray}\multicolumn{8}{c}{\textsc{GUI-Owl-1.5-2B: Upper Bound (100\% Tokens)}} \\
            GUI-Owl-1.5-2B & 90.49 & 57.94 & 71.84 & 39.70 & 44.23 & 57.93 & 100.0\% \\
            \rowcolor{headergray}\multicolumn{8}{c}{\textsc{Tight Budget (Single-step: $\keepr{=}5\%$ \,$||$\, Multi-step: $\keepc{=}25\%$, $\keeph{=}5\%$)}} \\
            Random & 14.54 & 1.71 & 8.35 & 24.77 & 16.21 & 47.43 & 35.3\% \\
            Uniform & 13.52 & 2.59 & 8.07 & 25.66 & 15.92 & 52.38 & 36.9\% \\
            DivPrune\pub{(CVPR25)\cite{alvar2025divprune}} & 18.00 & 2.28 & 8.57 & 26.52 & 20.52 & 50.30 & 39.3\% \\
            CDPruner\pub{(NeurIPS25)\cite{zhang2025cdpruner}} & \underline{22.17} & 2.66 & \underline{13.05} & 27.10 & 18.20 & 50.19 & 40.6\% \\
            VisPruner\pub{(ICCV25)\cite{zhang2025vispruner}} & 12.81 & 2.34 & 6.68 & 22.55 & 18.85 & 48.59 & 35.1\% \\
            PruMerge+\pub{(ICCV25)\cite{shang2025prumerge}} & 12.03 & 2.34 & 6.98 & 23.17 & 14.68 & 47.35 & 33.4\% \\
            TRIM\pub{(COLING25)\cite{song2025trim}} & 17.37 & 1.58 & 9.63 & 25.35 & 23.58 & 51.28 & 40.2\% \\
            FastV\pub{(ECCV24)\cite{chen2024fastv}} & 10.46 & \underline{6.58} & 7.57 & 25.51 & \textbf{27.62} & 51.71 & 41.6\% \\
            VisionTrim\pub{(ICLR26)\cite{yu2026visiontrim}} & 17.30 & 4.43 & 9.43 & 23.72 & 17.80 & 49.43 & 37.5\% \\
            ZOO-Prune\pub{(CVPR26)\cite{kim2026zooprune}} & 19.50 & 1.58 & 10.35 & 26.17 & 22.78 & 53.06 & 41.3\% \\
            PruneSID\pub{(ICLR26)\cite{fang2026prunesid}} & 19.97 & 1.83 & 10.66 & \underline{27.53} & 25.68 & \textbf{54.13} & \underline{43.5\%} \\
            \rowcolor{rowblue}\textbf{\method{}} & \textbf{38.29} & \textbf{6.83} & \textbf{21.93} & \textbf{31.88} & \underline{26.52} & \underline{53.72} & \textbf{52.9\%} \\
            \bottomrule
        \end{tabular}%
    }
\end{table}

\subsection{Experimental Setup}
\label{sec:exp-setup}
\noindent\textbf{Models and Evaluation Scope.}
We evaluate on native GUI agent models that ground natural-language instructions on raw screenshots and emit executable actions.
The \textit{GUI-Owl-1.5} series~\citep{guiowl2025} serves as the primary backbone for grounding and multi-step control, and \textit{UI-TARS-1.5-7B}~\citep{qin2025uitars} tests generalization under a different model family and action grammar.
\noindent\textbf{Benchmarks and Metrics.}
We evaluate under both single-step and multi-step settings across six public benchmarks.
The single-step setting measures grounding and general GUI understanding on ScreenSpot-v2 (SS-v2)~\citep{wu2024osatlas}, ScreenSpot-Pro (SS-Pro)~\citep{li2025screenspotpro}, and MMBench-GUI L2~\citep{wang2025mmbenchgui}, using their official accuracy metrics.
The multi-step setting evaluates action prediction on OmniGUI~\citep{henry2026omnigui}, Mind2Web~\citep{deng2023mind2web}, and AndroidControl~\citep{li2024androidcontrol}, and reports Step~SR, which credits a step only when both the action type and its target or argument are correct.
All methods are compared at matched visual-token budgets. The single-step setting retains a fraction $\keepr$ of the visual tokens, and the multi-step setting independently controls the current and historical budgets with $\keepc$ and $\keeph$.
More details are in the appendix.
\begin{table}[t]
    \centering
    \begin{minipage}[t]{0.545\textwidth}
        \centering
        \caption{Module ablation with \textbf{GUI-Owl-1.5-8B}. Rows without \nwe{} select by top-$k$ over the prior.}
        \label{tab:ablation-main}
        \adjustbox{max width=\linewidth}{%
            \footnotesize
            \setlength{\tabcolsep}{3.5pt}
            \renewcommand{\arraystretch}{1.0}
            \begin{tabular}{ccc cccc}
                \toprule
                \lmp{} & \nwe{} & \awc{} & \textbf{SS-v2} & \textbf{SS-Pro} & \textbf{OmniGUI} & \textbf{Mind2Web} \\
                \midrule
                \rowcolor{headergray}\multicolumn{7}{c}{\textsc{Mild budgets ($50\%/10\%$, SS-v2 $\keepr{=}10\%$, SS-Pro $\keepr{=}25\%$)}} \\
                \cmark &  &  & 32.63 & 41.94 & 40.94 & 41.75 \\
                 & \cmark &  & 36.48 & 42.19 & 45.88 & 34.31 \\
                \cmark & \cmark &  & \underline{56.45} & \underline{56.29} & \underline{47.43} & \underline{43.92} \\
                \rowcolor{rowblue}\cmark & \cmark & \cmark & \textbf{73.90} & \textbf{58.25} & \textbf{48.91} & \textbf{45.52} \\
                \midrule
                \rowcolor{headergray}\multicolumn{7}{c}{\textsc{Tight budgets ($25\%/5\%$, SS-v2 $\keepr{=}5\%$, SS-Pro $\keepr{=}10\%$)}} \\
                \cmark &  &  & 17.06 & 14.55 & 30.33 & 25.43 \\
                 & \cmark &  & 18.32 & 13.73 & 36.39 & 14.31 \\
                \cmark & \cmark &  & \underline{38.36} & \underline{28.15} & \underline{39.46} & \underline{27.79} \\
                \rowcolor{rowblue}\cmark & \cmark & \cmark & \textbf{55.19} & \textbf{37.63} & \textbf{43.08} & \textbf{34.20} \\
                \bottomrule
            \end{tabular}
        }
    \end{minipage}\hfill
    \begin{minipage}[t]{0.425\textwidth}
        \centering
        \caption{Factor ablation inside \nwe{}.}
        \label{tab:ablation-neo}
        \adjustbox{max width=\linewidth}{%
            \footnotesize
            \setlength{\tabcolsep}{4pt}
            \renewcommand{\arraystretch}{1.3}
            \begin{tabular}{l cccc}
                \toprule
                \textbf{Removed} &\textbf{SS-v2} & \textbf{SS-Pro} & \textbf{OmniGUI} & \textbf{Mind2Web} \\
                \midrule
                $-$ prior & $-23.74$ & $-16.50$ & $-5.64$ & $-15.50$ \\
                $-$ diversity & $-5.35$ & $-4.74$ & $-2.64$ & $-1.98$ \\
                $-$ instruction & $-15.57$ & $-14.16$ & $-4.12$ & $-1.39$ \\
                \bottomrule
            \end{tabular}
        }
        \par\vspace{8pt}
        \caption{Ordering ablation inside \nwe{}.}
        \label{tab:ablation-neoform}
        \adjustbox{max width=\linewidth}{%
            \footnotesize
            \setlength{\tabcolsep}{6pt}
            \renewcommand{\arraystretch}{1.1}
            \begin{tabular}{l cc}
                \toprule
                \textbf{Substituted rule} & \textbf{SS-v2} & \textbf{SS-Pro} \\
                \midrule
                $2\log a_j+\log d_j^2$ & $-10.53$ & $-3.61$ \\
                log-mean-exp pooling & $-8.49$ & $-6.45$ \\
                mean query row only & $-8.81$ & $-5.63$ \\
                \bottomrule
            \end{tabular}
        }
    \end{minipage}
    \vspace{-2mm}
\end{table}

\begin{table}[t]
    \centering
    \begin{minipage}[t]{0.535\textwidth}
        \centering
        \caption{Evaluation on \textbf{UI-TARS-1.5-7B}.}
        \label{tab:uitars-transfer}
        \adjustbox{max width=\linewidth}{%
            \footnotesize
            \setlength{\tabcolsep}{4pt}
            \renewcommand{\arraystretch}{0.95}
            \begin{tabular}{l cc cc}
                \toprule
                \textbf{Method} & \textbf{SS-v2} & \textbf{SS-Pro} & \textbf{OmniGUI} & \textbf{Mind2Web} \\
                \midrule
                UI-TARS-1.5-7B & 89.86 & 42.19 & 49.77 & 42.94 \\
                \rowcolor{headergray}\multicolumn{5}{c}{\textsc{Mild ($\keepr{=}25\%$, $\keepc{=}50\%$, $\keeph{=}10\%$)}} \\
                Random & 50.08 & 7.15 & 38.00 & 27.69 \\
                Uniform & 48.74 & 4.49 & 40.24 & 25.98 \\
                DivPrune & \underline{69.89} & \underline{14.86} & \underline{42.47} & \underline{32.33} \\
                CDPruner & 59.43 & 12.90 & 35.83 & 27.16 \\
                PruneSID & 68.87 & 13.41 & 41.55 & 31.21 \\
                \rowcolor{rowblue}\textbf{\method{}} & \textbf{70.28} & \textbf{16.76} & \textbf{43.52} & \textbf{35.13} \\
                \rowcolor{headergray}\multicolumn{5}{c}{\textsc{Tight ($\keepr{=}10\%$, $\keepc{=}25\%$, $\keeph{=}5\%$)}} \\
                Random & 16.59 & 1.83 & 21.83 & 8.55 \\
                Uniform & 18.00 & 1.71 & 23.14 & 6.51 \\
                DivPrune & \underline{36.79} & \underline{3.67} & \underline{28.60} & \underline{13.00} \\
                CDPruner & 30.03 & 3.35 & 23.80 & 12.13 \\
                PruneSID & 33.88 & 2.78 & 24.65 & 10.88 \\
                \rowcolor{rowblue}\textbf{\method{}} & \textbf{37.58} & \textbf{6.20} & \textbf{29.78} & \textbf{17.92} \\
                \bottomrule
            \end{tabular}
        }
    \end{minipage}\hfill
    \begin{minipage}[t]{0.435\textwidth}
        \centering
        \caption{Attribute ablation inside \lmp{}.}
        \label{tab:ablation-lip}
        \adjustbox{max width=\linewidth}{%
            \footnotesize
            \setlength{\tabcolsep}{10pt}
            \renewcommand{\arraystretch}{1.15}
            \begin{tabular}{l cc}
                \toprule
                \textbf{Removed} & \textbf{SS-v2} & \textbf{SS-Pro} \\
                \midrule
                $-$ containment & $-2.91$ & $-1.45$ \\
                $-$ resonance & $-2.28$ & $-1.01$ \\
                $-$ contrast & $-1.89$ & $-1.20$ \\
                $-$ entropy & $-0.08$ & $-0.57$ \\
                \bottomrule
            \end{tabular}
        }
        \par\vspace{8pt}
        \caption{Repair ablation inside \awc{}.}
        \label{tab:ablation-ncr}
        \adjustbox{max width=\linewidth}{%
            \footnotesize
            \setlength{\tabcolsep}{4pt}
            \renewcommand{\arraystretch}{1.15}
            \begin{tabular}{l cc cc}
                \toprule
                \multirow{2}{*}{\textbf{Rule}} & \multicolumn{2}{c}{Current frame} & \multicolumn{2}{c}{History frame} \\
                \cmidrule(lr){2-3}\cmidrule(lr){4-5}
                & \textbf{SS-v2} & \textbf{SS-Pro} & \textbf{OmniGUI} & \textbf{Mind2Web} \\
                \midrule
                Medoid & 58.65 & 30.11 & \textbf{43.08} & \textbf{34.20} \\
                Stride & \textbf{67.92} & \textbf{37.63} & 41.68 & 34.18 \\
                \bottomrule
            \end{tabular}
        }
    \end{minipage}
    \vspace{-0mm}
\end{table}

\\
\subsection{Main Results}
\label{sec:exp-single}
\label{sec:exp-multi}
\begin{figure}[t]
        \centering
        \includegraphics[width=0.95\linewidth, trim=0 0 0 426, clip]{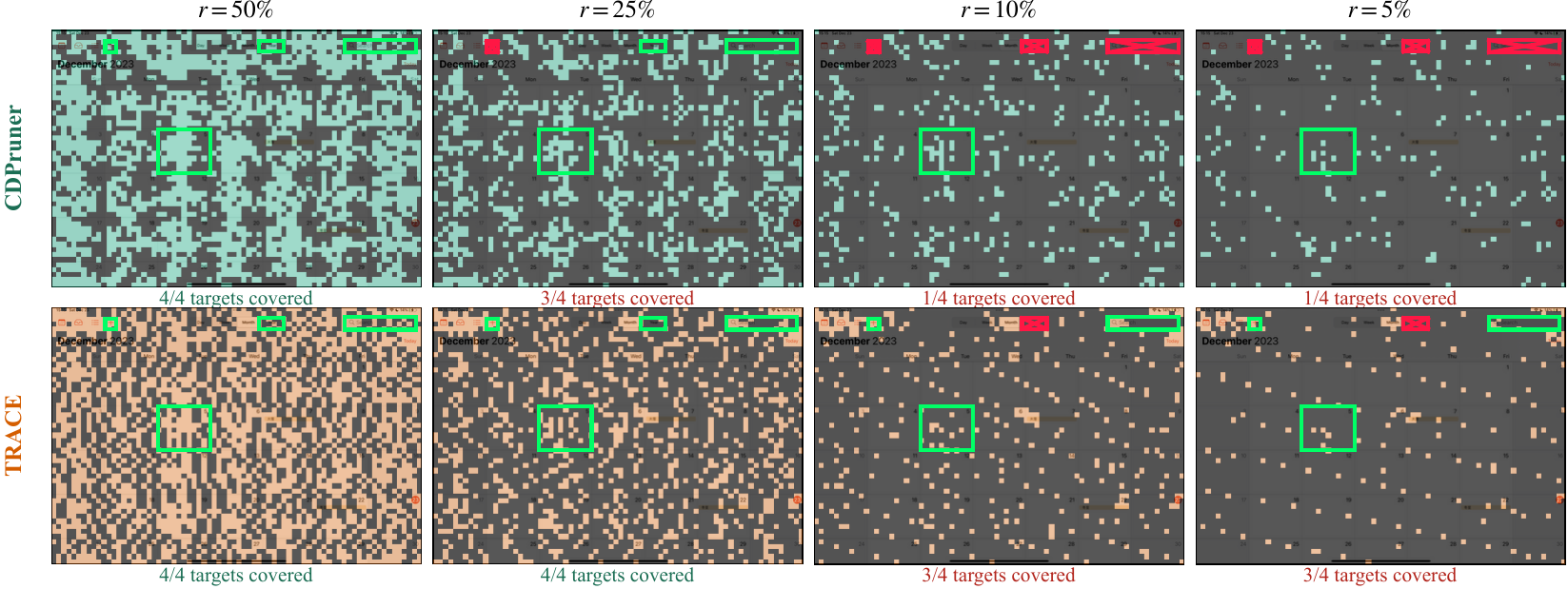}
        \vspace{-3.5mm}
        \caption{Keep maps on one SS-v2 screen carrying four instruction targets with \textbf{GUI-Owl-1.5-8B}.}
        \label{fig:keep-maps}
        \vspace{-2mm}
    \end{figure}

As shown in Tab.~\ref{tab:main-combined}, \method{} has the best performance among existing methods, retaining $78.7\%$ and $61.1\%$ of dense performance on GUI-Owl-1.5-8B at the mild and tight budgets.
\noindent\textbf{Single-step Evaluation.}
Specifically, at the mild budget, \method{} leads PruneSID by \textbf{14.94\%} on SS-v2 and VisPruner by \textbf{8.12\%} on MMBench-GUI.
Meanwhile, VisionTrim stays ahead by $0.89\%$ on the 4K icon-dense SS-Pro.
At $\keepr=5\%$, \method{} leads on all three benchmarks. In particular, its margin over PruneSID on SS-v2 grows to \textbf{20.99\%}, and the comparison with VisionTrim on SS-Pro turns to $+2.08\%$.
The advantage therefore widens as the budget tightens. 
As shown in Fig.~\ref{fig:keep-maps}, in this regime the instruction-conditioned keep contracts onto its highest-scoring region and falls from four covered targets to one, whereas the repaired keep still retains three of four at $\keepr{=}5\%$.
\noindent\textbf{Multi-step Evaluation.}
Meanwhile, \method{} achieves impressive performance on multi-step benchmarks.
Margins over the best existing method grow from $0.54\%$--$0.74\%$ at the mild budget to $1.30\%$--$3.53\%$ at the tight budget.
The largest gap appears on Mind2Web, where \method{} reaches $34.20\%$ and VisPruner reaches $30.67\%$.
\noindent\textbf{Cross-scale Evaluation.}
We further evaluate \method{} on GUI-Owl-1.5-2B.
Under the tight budget, \method{} retains $52.9\%$ of dense performance on average, while the best existing method, PruneSID, retains only $43.5\%$.
Moreover, \method{} ranks first on four of the six benchmarks and second on Mind2Web and AndroidControl, behind FastV and PruneSID.
Since the prior and coverage signals are computed before prefill and do not depend on the checkpoint, the advantages survive the scale change.
These results fully demonstrate the effectiveness of our proposed \method{}.
\begin{figure}[t]
    \centering
    \includegraphics[width=0.95\linewidth]{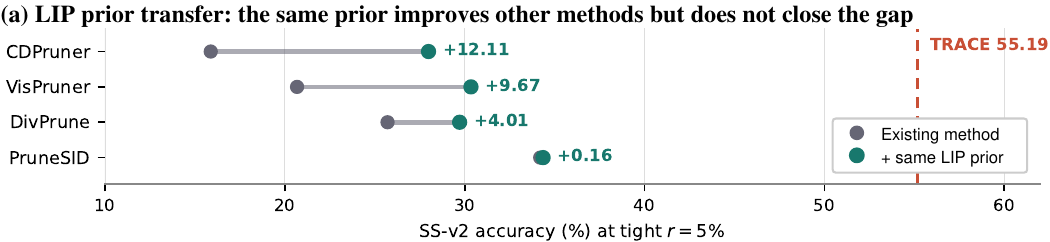}\\[1pt]
    \vspace{-1mm}
    \includegraphics[width=0.95\linewidth, trim=0 0 0 176, clip]{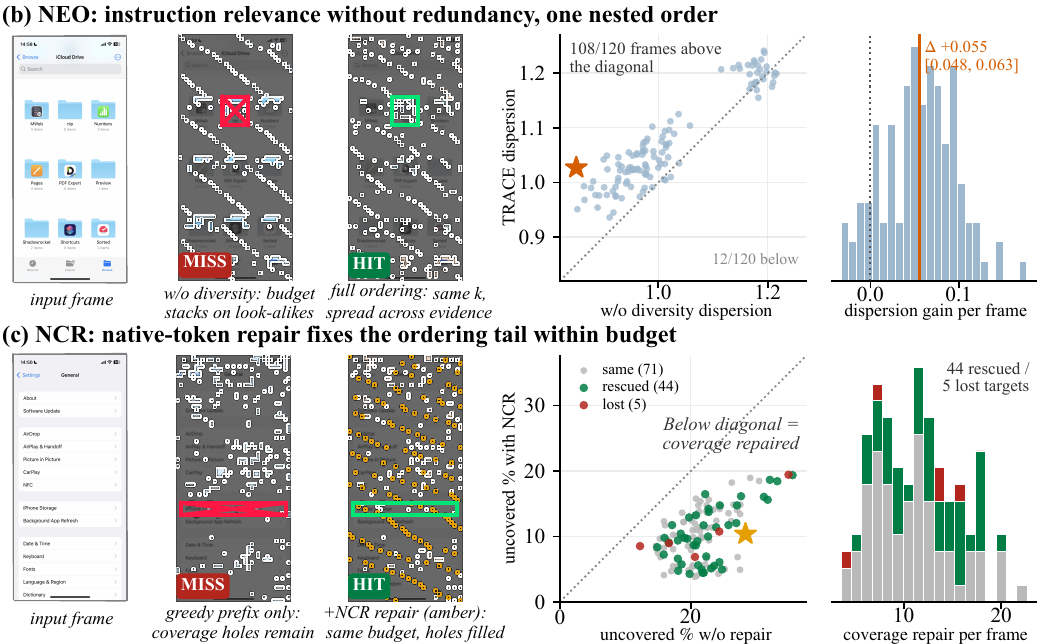}
    \vspace{-3mm}
    \caption{Mechanism validation with \textbf{GUI-Owl-1.5-8B} on SS-v2. 
    (a)~Analysis of the \lmp{} prior. 
    (b)~Analysis of the \nwe{} diversity term. 
    (c)~Analysis of the \awc{} coverage and target retention.}
    \label{fig:modules}
    \vspace{-3.5mm}
    \end{figure}
\subsection{Ablation Studies}
\label{sec:exp-ablation}
\label{sec:exp-modules}
\label{sec:exp-mech}
\label{sec:exp-analysis}
\label{sec:exp-transfer}
\label{sec:exp-eff}
\noindent\textbf{Analysis of Key Modules.}
As shown in Tab.~\ref{tab:ablation-main}, we analyze the contribution of each module in \method{}.
On SS-v2 at $\keepr{=}10\%$, the prior scored by top-$k$ reaches $32.63\%$ and \nwe{} alone reaches $36.48\%$.
In contrast, combining the two stages reaches $56.45\%$, which is \textbf{19.97\%} higher than either stage alone.
Adding \awc{} then raises SS-v2 accuracy from $56.45\%$ to $73.90\%$ and lifts tight Mind2Web from $27.79\%$ to $34.20\%$.
The prior thus locates operable regions, and the residual ordering keeps redundant tokens from consuming the budget.
Finally, \awc{} restores the spatial coverage lost to importance concentration.
These results fully verify the effectiveness of each module.
\\
\noindent\textbf{Module Design Choices.}
We further evaluate the design choices inside each module.
\noindent\textit{(1) Scoring factors inside \nwe{}.}
As shown in Tab.~\ref{tab:ablation-neo}, removing any one scoring factor degrades accuracy on all four benchmarks. 
Specifically, removing the prior costs the most, up to $-23.74\%$ on SS-v2. 
These results verify the complementarity of the three factors.
\noindent\textit{(2) Ordering form inside \nwe{}.}
As shown in Tab.~\ref{tab:ablation-neoform}, doubling the likelihood weight, pooling the query by log-mean-exp or collapsing it to the mean row costs $3.61\%$--$10.53\%$ on the two grounding benchmarks. 
These results verify the importance of each choice in the ordering rule.
\noindent\textit{(3) Energy attributes inside \lmp{}.}
As shown in Tab.~\ref{tab:ablation-lip}, removing any single energy attribute lowers accuracy on both benchmarks. 
Among the attributes, containment and resonance contribute the most, while contrast and entropy contribute less.
\noindent\textit{(4) Repair rules inside \awc{}.}
As shown in Tab.~\ref{tab:ablation-ncr}, stride tokens outperform feature medoids on the current frame, reaching $67.92\%$ on SS-v2 and $37.63\%$ on SS-Pro while the medoids reach $58.65\%$ and $30.11\%$.
The preference reverses on the history frame, where stride tokens cost $1.40\%$ on OmniGUI.
Thus, we adopt the stride rule on the current frame and the medoid rule on the history frame.
\begin{figure}[t]
    \centering
    \includegraphics[width=1\linewidth]{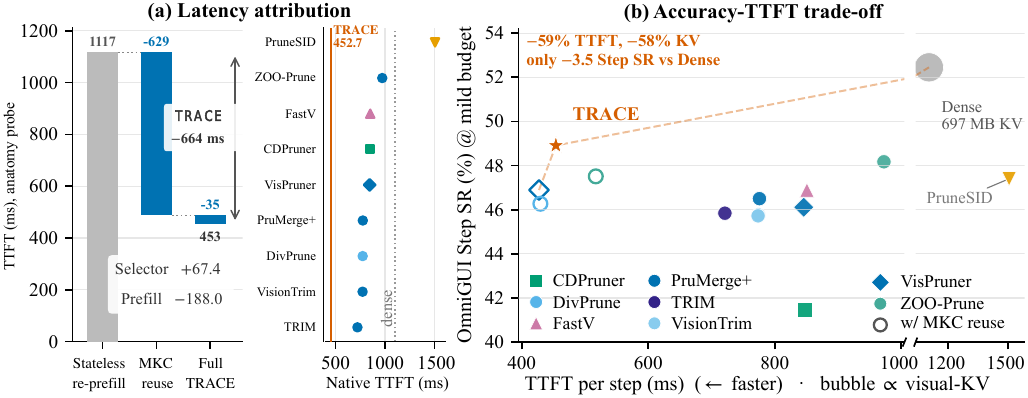}
    \vspace{-4mm}
    \caption{Serving efficiency analysis with \textbf{GUI-Owl-1.5-8B} on OmniGUI. 
    (a)~Latency attribution of the TTFT budget.
    (b)~Accuracy--TTFT trade-off, where hollow markers denote existing methods ported onto the \kvr{} path and bubble area encodes the visual KV cache.}
    \label{fig:pareto}
    \vspace{-2mm}
\end{figure}
\\
\noindent\textbf{Mechanism Validation.}
\noindent\textit{(1) Prior transfer across methods.}
We graft the same interaction prior onto four existing methods at the tight budget. 
As shown in Fig.~\ref{fig:modules} (a), every method improves, by $+0.16\%$ to $+12.11\%$. However, the best reaches only $34.36\%$, still far below \method{}'s $55.19\%$. Moreover, the prior selected alone scores just $17.06\%$.
The margin therefore comes from converting the layout signal into a nested and coverage-aware order rather than from the detector itself.
\noindent\textit{(2) Dispersion under the diversity term.}
As shown in Fig.~\ref{fig:modules} (b), without the \nwe{} diversity term, the order spends its budget on repeated glyphs. By comparison, the full ordering spreads the keep more widely on GUI screenshots.
\noindent\textit{(3) Target rescue under coverage repair.}
As shown in Fig.~\ref{fig:modules} (c), \awc{} reduces the uncovered targets on most frames at the same budget. In turn, this repair rescues the target on $44$ frames while losing it on only $5$.
These results verify the mechanism of each module.
\\
\noindent\textbf{Generalization across Backbones.}
We evaluate \method{} on UI-TARS-1.5-7B for generalization across model families.
As shown in Tab.~\ref{tab:uitars-transfer}, \method{} achieves the best performance at both budgets. 
Moreover, its margins over DivPrune grow from $0.39\%$--$2.80\%$ at the mild budget to $0.79\%$--$4.92\%$ at the tight budget.
The gap is again widest on tight Mind2Web, where \method{} reaches $17.92\%$ and DivPrune reaches $13.00\%$.
These gains fully demonstrate the generalization ability of \method{}.
\\
\noindent\textbf{Efficiency Analysis.}
\noindent\textit{(1) End-to-end serving result.}
We analyze the serving efficiency of \method{} with one CUDA-synchronized GPU.
As shown in Fig.~\ref{fig:pareto} (a), \method{} reduces TTFT from $1116.8$\,ms to $452.7$\,ms and the visual KV cache from $697$\,MB to $292$\,MB.
The pixel-input detector runs concurrently with the vision encode, so this measurement already charges the full critical-path selection cost.
Overall, \method{} is \textbf{2.4$\times$} faster than dense re-prefill and attains the highest Step~SR among existing pruning methods.
\noindent\textit{(2) Attribution of the gain.}
Fig.~\ref{fig:pareto} (a) further attributes this gain, where applying \kvr{} alone lowers TTFT from $1116.8$\,ms to $487.4$\,ms at unchanged accuracy.
Adding the complete selector then lowers TTFT further, to $452.7$\,ms at the mild budget and $397.8$\,ms at the tight budget.
Specifically, pruning cuts the remaining LLM prefill from $328.6$ to $140.6$ and $101.9$\,ms, which repays the $67.4$/$47.0$\,ms selection cost several times over.
As shown in Fig.~\ref{fig:pareto} (b), porting DivPrune, VisPruner and ZOO-Prune onto the same \kvr{} path improves their TTFT by $1.8$--$2.0\times$.
Even after this acceleration, all three methods still lag behind \method{}.
The \kvr{} path therefore supplies the largest saving, whereas the admitted order adds a second saving and the accuracy margin.
\noindent\textit{(3) Window truncation as an alternative.}
Keeping fewer dense frames is the non-selective alternative.
However, three recent frames still spend $90\%$ of the dense budget, and one frame still costs $1.66\times$ ours.
Thus, window truncation still leaves latency and memory usage high without accuracy gain.

\section{Conclusion}
\label{sec:conclusion}
In this paper, we study visual token pruning for efficient GUI agents under a reusable lifecycle.
We show that cache reuse turns pruning into an irreversible write-time commitment, so a selector must admit a nested order that remains useful before future grounding demands are known and that preserves spatial coverage under tight budgets.
To this end, we propose \textbf{\method{}}, a training-free framework for efficient GUI agents.
Specifically, we leverage Layout-derived Interaction Prior (\lmp{}) to derive a query-independent interaction prior from the interface layout.
Then, we employ Nested Evidence Ordering (\nwe{}) to couple that prior with instruction and feature novelty into one nested order.
Building upon this, we adopt Native-token Coverage Repair (\awc{}) to restore spatial coverage with native tokens while preserving nesting.
Finally, we propose Monotone KV Contraction (\kvr{}), which contracts retired frames into reusable session state without re-encoding.
Extensive experiments across diverse settings demonstrate the effectiveness and efficiency of \method{}.

\bibliography{refs}

@inproceedings{chen2024fastv,
  title     = {An Image is Worth 1/2 Tokens After Layer 2: Plug-and-Play Inference Acceleration for Large Vision-Language Models},
  author    = {Chen, Liang and Zhao, Haozhe and Liu, Tianyu and Bai, Shuai and Lin, Junyang and Zhou, Chang and Chang, Baobao},
  booktitle = {ECCV},
  year      = {2024}
}

@inproceedings{alvar2025divprune,
  title     = {DivPrune: Diversity-based Visual Token Pruning for Large Multimodal Models},
  author    = {Alvar, Saeed Ranjbar and Singh, Gursimran and Akbari, Mohammad and Zhang, Yong},
  booktitle = {CVPR},
  year      = {2025}
}

@inproceedings{zhang2025cdpruner,
  title     = {Beyond Attention or Similarity: Maximizing Conditional Diversity for Token Pruning in MLLMs},
  author    = {Zhang, Qizhe and Liu, Mengzhen and Li, Lichen and Lu, Ming and Zhang, Yuan and Pan, Junwen and She, Qi and Zhang, Shanghang},
  booktitle = {NeurIPS},
  year      = {2025}
}

@inproceedings{yang2025visionzip,
  title     = {VisionZip: Longer is Better but Not Necessary in Vision Language Models},
  author    = {Yang, Senqiao and Chen, Yukang and Tian, Zhuotao and Wang, Chengyao and Li, Jingyao and Yu, Bei and Jia, Jiaya},
  booktitle = {CVPR},
  year      = {2025}
}

@inproceedings{zhang2025vispruner,
  title     = {Beyond Text-Visual Attention: Exploiting Visual Cues for Effective Token Pruning in VLMs},
  author    = {Zhang, Qizhe and Cheng, Aosong and Lu, Ming and Zhuo, Zhiyong and Wang, Minqi and Cao, Jiajun and Guo, Shaobo and She, Qi and Zhang, Shanghang},
  booktitle = {ICCV},
  year      = {2025}
}

@inproceedings{shang2025prumerge,
  title     = {LLaVA-PruMerge: Adaptive Token Reduction for Efficient Large Multimodal Models},
  author    = {Shang, Yuzhang and Cai, Mu and Xu, Bingxin and Lee, Yong Jae and Yan, Yan},
  booktitle = {ICCV},
  year      = {2025}
}

@inproceedings{song2025trim,
  title     = {Less is More: A Simple yet Effective Token Reduction Method for Efficient Multi-modal LLMs},
  author    = {Song, Dingjie and Wang, Wenjun and Chen, Shunian and Wang, Xidong and Guan, Michael and Wang, Benyou},
  booktitle = {COLING},
  year      = {2025}
}

@inproceedings{xing2025pdrop,
  title     = {PyramidDrop: Accelerating Your Large Vision-Language Models via Pyramid Visual Redundancy Reduction},
  author    = {Xing, Long and Huang, Qidong and Dong, Xiaoyi and Lu, Jiajie and Zhang, Pan and Zang, Yuhang and Cao, Yuhang and He, Conghui and Wang, Jiaqi and Wu, Feng and Lin, Dahua},
  booktitle = {CVPR},
  year      = {2025}
}

@inproceedings{yu2026visiontrim,
  title     = {VisionTrim: Unified Vision Token Compression for Training-Free {MLLM} Acceleration},
  author    = {Yu, Hanxun and Li, Wentong and Qu, Xuan and Wang, Song and Chen, Junbo and Zhu, Jianke},
  booktitle = {ICLR},
  year      = {2026}
}

@inproceedings{kim2026zooprune,
  title     = {{ZOO}-Prune: Training-Free Token Pruning via Zeroth-Order Gradient Estimation in Vision-Language Models},
  author    = {Kim, Youngeun and Zhang, Youjia and Liu, Huiling and Jung, Aecheon and Lee, Sunwoo and Hong, Sungeun},
  booktitle = {CVPR},
  year      = {2026},
  pages     = {39572--39582}
}

@inproceedings{fang2026prunesid,
  title     = {Prune Redundancy, Preserve Essence: Vision Token Compression in {VLM}s via Synergistic Importance-Diversity},
  author    = {Fang, Zhengyao and Lyu, Pengyuan and Zhang, Chengquan and Lu, Guangming and Yu, Jun and Pei, Wenjie},
  booktitle = {ICLR},
  year      = {2026}
}

@inproceedings{sun2026ivcprune,
  title     = {{IVC}-Prune: Revealing the Implicit Visual Coordinates in {LVLM}s for Vision Token Pruning},
  author    = {Sun, Zhichao and Ma, Yidong and Liu, Gang and Chen, Yibo and Tang, Xu and Hu, Yao and Xu, Yongchao},
  booktitle = {ICLR},
  year      = {2026}
}

@article{qin2025uitars,
  title     = {UI-TARS: Pioneering Automated GUI Interaction with Native Agents},
  author    = {Qin, Yujia and Ye, Yining and Fang, Junjie and Wang, Haoming and Liang, Shihao and Tian, Shizuo and Zhang, Junda and Li, Jiahao and Li, Yunxin and Huang, Shijue and Zhong, Wanjun and Li, Kuanye and Yang, Jiale and Miao, Yu and Lin, Woyu and Liu, Longxiang and Jiang, Xu and Ma, Qianli and Li, Jingyu and Xiao, Xiaojun and Cai, Kai and Li, Chuang and Zheng, Yaowei and Jin, Chaolin and Li, Chen and Zhou, Xiao and Wang, Minchao and Chen, Haoli and Li, Zhaojian and Yang, Haihua and Liu, Haifeng and Lin, Feng and Peng, Tao and Liu, Xin and Shi, Guang},
  journal   = {arXiv preprint arXiv:2501.12326},
  year      = {2025}
}

@article{guiowl2025,
  title     = {Mobile-Agent-v3.5: Multi-platform Fundamental GUI Agents},
  author    = {Xu, Haiyang and Zhang, Xi and Liu, Haowei and Wang, Junyang and Zhu, Zhaoqing and Zhou, Shengjie and Hu, Xuhao and Gao, Feiyu and Cao, Junjie and Wang, Zihua and Chen, Zhiyuan and Liao, Jitong and Zheng, Qi and Zeng, Jiahui and Xu, Ze and Bai, Shuai and Lin, Junyang and Zhou, Jingren and Yan, Ming},
  journal   = {arXiv preprint arXiv:2602.16855},
  year      = {2026}
}

@article{histprune2026,
  title     = {Rethinking Token Pruning for Historical Screenshots in GUI Visual Agents: Semantic, Spatial, and Temporal Perspectives},
  author    = {Li, Daiqiang and Pan, Zihao and Zhang, Zeyu and Chen, Ronghao and Wang, Huacan and Chen, Honggang and Jiang, Haiyun},
  journal   = {arXiv preprint arXiv:2603.26041},
  year      = {2026}
}

@article{huang2025guikv,
  title     = {GUI-KV: Efficient GUI Agents via KV Cache with Spatio-Temporal Awareness},
  author    = {Huang, Kung-Hsiang and Qiu, Haoyi and Dai, Yutong and Xiong, Caiming and Wu, Chien-Sheng},
  journal   = {arXiv preprint arXiv:2510.00536},
  year      = {2025}
}

@article{xu2026guipruner,
  title     = {Spatio-Temporal Token Pruning for Efficient High-Resolution GUI Agents},
  author    = {Xu, Zhou and Zhou, Bowen and Wang, Qi and Feng, Shuwen and Xiao, Jingyu},
  journal   = {arXiv preprint arXiv:2602.23235},
  year      = {2026}
}

@article{zhou2026stlite,
  title     = {Efficient Long-Horizon GUI Agents via Training-Free KV Cache Compression},
  author    = {Zhou, Bowen and Xu, Zhou and Li, Wanli and Xiao, Jingyu and Wang, Haoqian},
  journal   = {arXiv preprint arXiv:2603.00188},
  year      = {2026}
}

@article{li2026aquaui,
  title     = {AQuaUI: Visual Token Reduction for GUI Agents with Adaptive Quadtrees},
  author    = {Li, Yuankai and Zhu, Tinghui and Son, Ha Min and Zhao, Zhe and Liu, Xin and Chen, Muhao},
  journal   = {arXiv preprint arXiv:2605.19260},
  year      = {2026}
}

@article{han2026starkv,
  title     = {STaR-KV: Spatio-Temporal Adaptive Re-weighting for KV Cache Compression in GUI Vision-Language Models},
  author    = {Han, Yuhang and Yang, Wenzheng and Chen, Yujie and Jin, Xiangqi and Zhang, Yaojie and Huang, Siteng and Zhang, Linfeng},
  journal   = {arXiv preprint arXiv:2606.01790},
  year      = {2026}
}

@inproceedings{cheng2024seeclick,
  title     = {SeeClick: Harnessing GUI Grounding for Advanced Visual GUI Agents},
  author    = {Cheng, Kanzhi and Sun, Qiushi and Chu, Yougang and Xu, Fangzhi and Li, Yantao and Zhang, Jianbing and Wu, Zhiyong},
  booktitle = {ACL},
  year      = {2024}
}

@inproceedings{deng2023mind2web,
  title     = {Mind2Web: Towards a Generalist Agent for the Web},
  author    = {Deng, Xiang and Gu, Yu and Zheng, Boyuan and Chen, Shijie and Stevens, Samuel and Wang, Boshi and Sun, Huan and Su, Yu},
  booktitle = {NeurIPS},
  year      = {2023}
}

@article{henry2026omnigui,
  title     = {OmniGUI: Benchmarking GUI Agents in Omni-Modal Smartphone Environments},
  author    = {Henry, Felix and Lin, Xiaochen and Zhu, Jiangyou and Zhang, Bingqian and Chen, Min and Huang, Shiyu and others},
  journal   = {arXiv preprint arXiv:2605.18758},
  year      = {2026}
}

@article{li2024androidcontrol,
  title     = {On the Effects of Data Scale on UI Control Agents},
  author    = {Li, Wei and Bishop, William and Li, Alice and Rawles, Christopher and Campbell-Ajala, Folawiyo and Tyamagundlu, Divya and Riva, Oriana},
  journal   = {NeurIPS},
  year      = {2024}
}

@article{wu2024osatlas,
  title     = {OS-ATLAS: A Foundation Action Model for Generalist GUI Agents},
  author    = {Wu, Zhiyong and Wu, Zhenyu and Xu, Fangzhi and Wang, Yian and Sun, Qiushi and Jia, Chengyou and Cheng, Kanzhi and Ding, Zichen and Chen, Liheng and Liang, Paul Pu and Qiao, Yu},
  journal   = {arXiv preprint arXiv:2410.23218},
  year      = {2024}
}

@inproceedings{li2025screenspotpro,
  title     = {ScreenSpot-Pro: GUI Grounding for Professional High-Resolution Computer Use},
  author    = {Li, Kaixin and Meng, Ziyang and Lin, Hongzhan and Luo, Ziyang and Tian, Yuchen and Ma, Jing and Huang, Zhiyong and Chua, Tat-Seng},
  booktitle = {Workshop on Reasoning and Planning for Large Language Models},
  year      = {2025}
}

@article{wang2025mmbenchgui,
  title     = {MMBench-GUI: Hierarchical Multi-Platform Evaluation Framework for GUI Agents},
  author    = {Wang, Xuehui and Wu, Zhenyu and Xie, JingJing and Ding, Zichen and Yang, Bowen and Li, Zehao and Liu, Zhaoyang and Li, Qingyun and Dong, Xuan and Chen, Zhe and Wang, Weiyun and Zhao, Xiangyu and Chen, Jixuan and Duan, Haodong and Xie, Tianbao and Yang, Chenyu and Su, Shiqian and Yu, Yue and Zhang, Yanting and Yue, Xiangyu and Su, Weijie and Zhu, Xizhou and Shen, Wei and Dai, Jifeng and Wang, Wenhai},
  journal   = {arXiv preprint arXiv:2507.19478},
  year      = {2025}
}

@article{lu2024omniparser,
  title     = {OmniParser for Pure Vision Based GUI Agent},
  author    = {Lu, Yadong and Yang, Jianwei and Shen, Yelong and Awadallah, Ahmed},
  journal   = {arXiv preprint arXiv:2408.00203},
  year      = {2024}
}

@article{hong2024cogagent,
  title     = {CogAgent: A Visual Language Model for GUI Agents},
  author    = {Hong, Wenyi and Wang, Weihan and Lv, Qingsong and Xu, Jiazheng and Yu, Wenmeng and Ji, Junhui and Wang, Yan and Wang, Zihan and Dong, Yuxiao and Ding, Ming and Tang, Jie},
  journal   = {CVPR},
  year      = {2024}
}

@inproceedings{khaki2025sparsevila,
  title     = {{SparseVILA}: Decoupling Visual Sparsity for Efficient {VLM} Inference},
  author    = {Khaki, Samir and Guo, Junxian and Tang, Jiaming and Yang, Shang and Chen, Yukang and Plataniotis, Konstantinos N. and Lu, Yao and Han, Song and Liu, Zhijian},
  booktitle = {ICCV},
  year      = {2025}
}

@inproceedings{yang2025topv,
  title     = {{TopV}: Compatible Token Pruning with Inference Time Optimization for Fast and Low-Memory Multimodal Vision Language Model},
  author    = {Yang, Cheng and Sui, Yang and Xiao, Jinqi and Huang, Lingyi and Gong, Yu and Li, Chendi and Yan, Jinghua and Bai, Yu and Sadayappan, Ponnuswamy and Hu, Xia and Yuan, Bo},
  booktitle = {CVPR},
  year      = {2025}
}

@inproceedings{deng2025scope,
  title     = {{SCOPE}: Saliency-Coverage Oriented Token Pruning for Efficient Multimodal {LLMs}},
  author    = {Deng, Jinhong and Li, Wen and Zhou, Joey Tianyi and He, Yang},
  booktitle = {NeurIPS},
  year      = {2025},
  volume    = {38},
  pages     = {161527--161552}
}

@inproceedings{xu2026score,
  title     = {{SCoRe}: Salience-Coverage Reduction for Vision Token Pruning in Vision-Language Models},
  author    = {Xu, Tong and Shi, Hailong and Gao, Xingyu},
  booktitle = {CVPR},
  year      = {2026}
}

@inproceedings{endo2025feather,
  title     = {Feather the Throttle: Revisiting Visual Token Pruning for Vision-Language Model Acceleration},
  author    = {Endo, Mark and Wang, Xiaohan and Yeung-Levy, Serena},
  booktitle = {ICCV},
  year      = {2025}
}

@inproceedings{cai2025matryoshka,
  title     = {Matryoshka Multimodal Models},
  author    = {Cai, Mu and Yang, Jianwei and Gao, Jianfeng and Lee, Yong Jae},
  booktitle = {ICLR},
  year      = {2025}
}

@article{liu2025palui,
  title     = {{PAL-UI}: Planning with Active Look-back for Vision-Based {GUI} Agents},
  author    = {Liu, Zikang and Li, Junyi and Zhao, Wayne Xin and Gao, Dawei and Li, Yaliang and Wen, Ji-Rong},
  journal   = {arXiv preprint arXiv:2510.00413},
  year      = {2025}
}

@article{abaskohi2026revision,
  title     = {{ReVision}: Scaling Computer-Use Agents via Temporal Visual Redundancy Reduction},
  author    = {Abaskohi, Amirhossein and He, Yuhang and West, Peter and Carenini, Giuseppe and Chawla, Pranit and Vineet, Vibhav},
  journal   = {arXiv preprint arXiv:2605.11212},
  year      = {2026}
}

@article{xie2024osworld,
  title={Osworld: Benchmarking multimodal agents for open-ended tasks in real computer environments},
  author={Xie, Tianbao and Zhang, Danyang and Chen, Jixuan and Li, Xiaochuan and Zhao, Siheng and Cao, Ruisheng and Hua, Toh J and Cheng, Zhoujun and Shin, Dongchan and Lei, Fangyu and others},
  journal={NeurIPS},
  volume={37},
  pages={52040--52094},
  year={2024}
}

@inproceedings{tian2025mmina,
  title={Mmina: Benchmarking multihop multimodal internet agents},
  author={Tian, Shulin and Zhang, Ziniu and Chen, Liang-Yu and Liu, Ziwei},
  booktitle={ACL Findings},
  pages={13682--13697},
  year={2025}
}

@inproceedings{zheng2024sglang,
  title     = {{SGLang}: Efficient Execution of Structured Language Model Programs},
  author    = {Zheng, Lianmin and Yin, Liangsheng and Xie, Zhiqiang and Sun, Chuyue and Huang, Jeff and Yu, Cody Hao and Cao, Shiyi and Kozyrakis, Christos and Stoica, Ion and Gonzalez, Joseph E. and others},
  booktitle = {NeurIPS},
  year      = {2024}
}
\bibliographystyle{arxiv_preprint}
\clearpage
\appendix
\vspace{1ex}
\centerline{\LARGE\sc Appendix}
\vspace{1.5ex}
\label{app:appendix}
This appendix presents implementation details, the evaluation protocol, and additional experimental evidence that complements the main paper.
Section~\ref{app:method-details} specifies the lifecycle contract, the method implementation, the comparison protocol, and the configuration choices.
Section~\ref{app:evidence} presents additional experiments, covering lifecycle behavior, benchmark breakdowns, ablations, and serving efficiency.
Section~\ref{app:transfer} examines transfer to other backbones and states the scope and limitations of the evidence.
Together, these analyses provide a comprehensive understanding of our proposed \method{}.
\renewcommand{\topfraction}{0.85}
\renewcommand{\bottomfraction}{0.75}
\renewcommand{\textfraction}{0.12}
\renewcommand{\floatpagefraction}{0.5}
\setcounter{topnumber}{2}
\setcounter{bottomnumber}{2}
\setcounter{totalnumber}{4}
\setlength{\textfloatsep}{8pt plus 2pt minus 3pt}
\setlength{\intextsep}{6pt plus 2pt minus 2pt}
\setlength{\floatsep}{8pt plus 2pt minus 2pt}
\raggedbottom
\makeatletter
\let\trace@origtable\table
\def\table{\@ifnextchar[\trace@tableopt{\trace@origtable[ht]}}
\def\trace@tableopt[#1]{\trace@origtable[ht]}
\makeatother
\section{Method, Protocol, and Configuration}
\label{app:method-details}
\subsection{Lifecycle Contract and Method Space}
Tab.~\ref{tab:contract} compares the nine existing methods of \S\ref{sec:exp} against the three constraints of Eq.~\ref{eq:lifecycle}.
We score each method from its official selection rule.
Since the constraints describe our deletion-only serving path, a mark records compatibility with this path rather than a universal judgment on the method.
Under condition (ii), \cmark{} denotes a rank or greedy-prefix rule that nests by construction, \pmark{} denotes a quota, merge, or de-duplication rule that nests only after a specified modification, and \xmark{} denotes per-step re-scoring, which cannot nest.
Condition (i) requires that the keep be decided from information available before the first LLM prefill.
The last column lists the tensor that the official rule must observe to produce that keep.
When this tensor is the LLM self-attention of the live prompt, as in FastV, it is available only during prefill, so condition (i) fails.
A superscript asterisk marks methods that consult the episode instruction, which precedes prefill and is therefore admitted by condition (i).
For detailed compliance, FastV fails conditions (i) and (ii), because its score is the LLM self-attention of the live prompt.
Merge-family methods (PruMerge+, TRIM, and VisionTrim) fail condition (iii), because synthetic tokens cannot later be dropped as cache rows on this path.
VisPruner and PruneSID receive \pmark{} under condition (ii), because quota or de-duplication nests only after a specified modification.
In contrast, DivPrune, CDPruner, ZOO-Prune, and \method{} satisfy all three conditions under their official rules, which enables a fair comparison on the same lifecycle.
\begin{table}[t]
\centering
\caption{Existing methods evaluated against rules of lifecycle-aware visual pruning.}
\label{tab:contract}
\vspace{2pt}
\small
\setlength{\tabcolsep}{5pt}
\begin{adjustbox}{max width=\linewidth}
\begin{tabular}{lcccl}
\toprule
Method & (i) pre-prefill & (ii) nested & (iii) index-level & Query-dependent tensor (layer) \\
\midrule
FastV & \xmark & \xmark & \cmark & LLM self-attention of the live prompt (layer 2) \\
PruMerge+ & \cmark & \pmark & \xmark & none (synthetic merged embeddings) \\
TRIM & \cmark & \pmark & \xmark & none$^{\ast}$ (discarded set fused into one token) \\
VisionTrim & \cmark & \pmark & \xmark & none$^{\ast}$ (synthetic TGVC merge rounds) \\
VisPruner & \cmark & \pmark & \cmark & none (ViT final-block attention) \\
PruneSID & \cmark & \pmark & \cmark & none (encoder-side PCA grouping) \\
DivPrune & \cmark & \cmark & \cmark & none (feature geometry only) \\
CDPruner & \cmark & \cmark & \cmark & none$^{\ast}$ (instruction-conditioned DPP) \\
ZOO-Prune & \cmark & \cmark & \cmark & none (projector-input sensitivity) \\
\midrule
\method{} (ours) & \cmark & \cmark & \cmark & none$^{\ast}$ (prior + repair anchored) \\
\bottomrule
\end{tabular}
\end{adjustbox}
\end{table}

\noindent\textbf{Feature Matrix of Prior Families.}
\label{app:feature-matrix}
Tab.~\ref{tab:feature-matrix} compares representative prior families along the six axes of our setting.
The columns record six properties. They cover multi-step reuse of visual state, GUI-screen design, decisions made before language-model processing, native token rows at their original positions, a nested order that serves two budgets, and no re-encoding of historical frames.
TopV's no-re-encoding mark ($^{a}$) holds within a single generation rather than across frames.
The multi-step mark of VisionZip and PruMerge ($^{b}$) denotes text-agnostic compression for multi-turn use.
SparseVILA's pre-admission mark ($^{c}$) is conservative prefill pruning followed by decode-time retrieval.
Matryoshka's nested-budget mark ($^{d}$) encodes nested granularity in the representation rather than ordering native rows at admission.
Existing visual-token pruners cover pre-admission or native rows, but not GUI multi-step reuse.
In contrast, GUI input-side methods cover multi-step screens and a decision before the language model, yet they do not keep native rows or a nested order.
GUI cache-side methods keep native rows without re-encoding, but they re-score at every step, so they miss pre-admission and nested budgets.
Different from the prior families, \method{} covers all six axes with our proposed lifecycle contract, providing a unified framework for GUI agents.
\begin{table}[t]
    \centering
    \caption{Comparison of representative prior families over the six axes of our setting.}
    \label{tab:feature-matrix}
    \adjustbox{max width=\textwidth,center}{%
        \footnotesize
        \setlength{\tabcolsep}{4.5pt}
        \renewcommand{\arraystretch}{1.02}
        \begin{tabular}{l cccccc}
            \toprule
            \textbf{Method family} & \textbf{Multi-step} & \textbf{GUI} & \textbf{Pre-admission} & \textbf{Native rows} & \textbf{Nested budgets} & \textbf{No re-encoding} \\
            \midrule
            FastV / PyramidDrop / IVC-Prune~\citep{chen2024fastv,xing2025pdrop,sun2026ivcprune} & -- & -- & -- & \cmark & -- & -- \\
            TopV~\citep{yang2025topv} & -- & -- & \cmark & \cmark & -- & \cmark$^{a}$ \\
            DivPrune / CDPruner / VisPruner~\citep{alvar2025divprune,zhang2025cdpruner,zhang2025vispruner} & -- & -- & \cmark & \cmark & -- & -- \\
            VisionZip / PruMerge~\citep{yang2025visionzip,shang2025prumerge} & \cmark$^{b}$ & -- & \cmark & -- & -- & -- \\
            SparseVILA~\citep{khaki2025sparsevila} & \cmark & -- & \cmark$^{c}$ & \cmark & -- & \cmark \\
            SCOPE / SCoRe / FEATHER~\citep{deng2025scope,xu2026score,endo2025feather} & -- & -- & \cmark & \cmark & -- & -- \\
            Matryoshka (M$^3$)~\citep{cai2025matryoshka} & \cmark & -- & \cmark & -- & \cmark$^{d}$ & -- \\
            GUIPruner / AQuaUI / ReVision~\citep{xu2026guipruner,li2026aquaui,abaskohi2026revision} & \cmark & \cmark & \cmark & -- & -- & -- \\
            GUI-KV / ST-Lite / STaR-KV~\citep{huang2025guikv,zhou2026stlite,han2026starkv} & \cmark & \cmark & -- & \cmark & -- & \cmark \\
            HistPrune-GUI~\citep{histprune2026} & \cmark & \cmark & -- & \cmark & -- & -- \\
            PAL-UI~\citep{liu2025palui} & \cmark & \cmark & -- & -- & -- & -- \\
            \midrule
            \rowcolor{rowblue}\textbf{\method{} (ours)} & \cmark & \cmark & \cmark & \cmark & \cmark & \cmark \\
            \bottomrule
        \end{tabular}
    }
\end{table}

\subsection{Method Specification}
This subsection formalizes the complete decision path during inference, from detector-derived interaction priors and evidence ordering to native-token repair and monotone cache contraction.
We present these components in pipeline order for easy understanding.
The details are as follows.
\\
\noindent\textbf{Layout-derived Interaction Prior.}
A generic detector that was not trained on GUI screens readily misses dense operable widgets.
Thus, we use the \texttt{icon\_detect} branch of OmniParser-v2~\citep{lu2024omniparser} which is fine-tuned on interactable web and desktop elements.
We load this branch alone to get axis-aligned boxes.
Specifically, we detect widgets at a confidence of $0.05$ with non-maximum suppression at an intersection-over-union of $0.1$.
Besides, we resize the longer image edge to $768$ pixels before snapping to the network stride.
The detector reads screenshot pixels rather than encoder features, so it can run at admission.
After admission, retirement operates solely on the shrinking order, so historical frames require neither bounding boxes nor rerunning the detector.
\\
\noindent\cnum{1}~\textit{Energy Definitions.}
For each detection $b$, we compute entropy~$H_b$, contrast~$C_b$, containment~$G_b$ and resonance~$R_b$.
Then we generate the interaction energy with the following equation:
\[
E_b=H_b+C_b+G_b+R_b.
\]
All four attributes lie in $[0,1]$ under our definition.
Specifically, $H_b$ is the base-2 entropy of a 256-bin Sobel-magnitude histogram on a $32\times32$ grayscale crop, divided by eight and clipped to the unit interval.
$C_b$ is the ascending rank of the euclidean difference between the mean CIELAB values on the three-pixel inner and outer boundary rings.
$G_b=1/(1+n_b)$, where $n_b$ counts strictly smaller boxes contained by $b$.
$R_b$ is the min to max normalized resonance score. It is the larger of the row and column peer counts within half the median box height or width.
Each attribute is computed from the box alone, so the prior is query-independent and concentrates on operable regions.
\\
\noindent\cnum{2}~\textit{Energy Attributes.}
Tab.~\ref{tab:lip-energy} shows that removing any single attribute from $E_b$ lowers accuracy on both grounding benchmarks.
Containment $G_b$ is the costliest removal ($-2.91\%$ and $-1.45\%$), and resonance $R_b$ follows it on ScreenSpot-v2 ($-2.28\%$).
Contrast and entropy contribute less but are still significant.
These results fully demonstrate that the prior captures operable regions.
\begin{table}[ht]
    \centering
    \caption{Energy attributes inside \lmp{}. $\Delta$ is accuracy minus the full energy.}
    \label{tab:lip-energy}
    \adjustbox{max width=0.86\textwidth,center}{%
        \footnotesize
        \setlength{\tabcolsep}{8pt}
        \renewcommand{\arraystretch}{0.98}
        \begin{tabular}{ll cc cc}
            \toprule
            & & \multicolumn{2}{c}{\textbf{ScreenSpot-v2} ($\keepr{=}5\%$)} & \multicolumn{2}{c}{\textbf{ScreenSpot-Pro} ($\keepr{=}10\%$)} \\
            \cmidrule(lr){3-4}\cmidrule(lr){5-6}
            \textbf{Energy} & \textbf{Attribute} & Acc & $\Delta$ & Acc & $\Delta$ \\
            \midrule
            \rowcolor{rowblue}$H{+}C{+}G{+}R$ & retained & \textbf{55.19} & --- & \textbf{37.63} & --- \\
            \midrule
            $-\,G_b$ & containment & 52.28 & $-2.91$ & 36.18 & $-1.45$ \\
            $-\,R_b$ & resonance & 52.91 & $-2.28$ & 36.62 & $-1.01$ \\
            $-\,C_b$ & boundary contrast & 53.30 & $-1.89$ & 36.43 & $-1.20$ \\
            $-\,H_b$ & texture entropy & 55.11 & $-0.08$ & 37.07 & $-0.57$ \\
            \bottomrule
        \end{tabular}
    }
\end{table}

\\
\noindent\textbf{Nested Evidence Ordering.}
With the interaction prior defined, Nested Evidence Ordering (NEO) determines how visual evidence is admitted and how the resulting order can be reused under different budgets.
It combines within-frame instruction relevance with prior-weighted novelty and residual coverage to produce a single nested sequence of native visual tokens.
In this subsection, we clarify three points before presenting the final ordering.
They concern degenerate cosine scores, the role of $m_j$ in the residual, and the exact scoring and pooling form.
These points correspond to the following three implementation details and lead directly to the admission procedure summarized in Alg.~\ref{alg:nested}.
\\
\noindent\cnum{1}~\textit{Degenerate Relevance.}
The z-score in Eq.~\ref{eq:likelihood} is taken over the $N$ visual tokens of the current frame, so $a_j$ is a within-frame ranking rather than an absolute cosine.
If that variance is zero, $a_j$ falls back to the uniform distribution.
Then $\log a_j$ becomes a constant in Eq.~\ref{eq:qd}.
Under this degenerate case, the order reduces to prior-weighted residual selection rather than failing, ensuring robust token pruning.
\\
\noindent\cnum{2}~\textit{Equivalent Residual.}
The mass $m_j$ enters Eq.~\ref{eq:qd} through $\boldsymbol\psi_j=\sqrt{m_j}\,\mathbf z_j$.
Because $m_j=1+\alpha Np_j\geq 1$, the selected features $\{\boldsymbol\psi_i\}_{i\in S}$ and $\{\mathbf z_i\}_{i\in S}$ span the same subspace.
The projector $\Pi_S$ is therefore the euclidean projector onto $\mathrm{span}(\mathbf z_S)$.
We compute the residual $d_j^2(S)$ as follows.
\begin{equation}
d_j^2(S)
=
m_j
\left\lVert
\mathbf z_j-\Pi_{\mathrm{span}(\mathbf z_S)}\mathbf z_j
\right\rVert_2^2 .
\label{eq:mass-residual}
\end{equation}
Thus, $m_j$ scales only the candidate's novelty, and it does not reweight a direction already in $S$.
\\
\noindent\textbf{\textit{Ordering Form.}}
Eq.~\ref{eq:likelihood} and Eq.~\ref{eq:qd} determine several design choices that might otherwise appear arbitrary.
To be specific, the coefficient on $\log a_j$ is fixed to one.
The query rows are combined by max-pooling.
The query matrix $\mathbf U$ includes the instruction-token rows.
Tab.~\ref{tab:neo-form} reports the ablation results at $\keepr{=}10\%$.
Doubling the coefficient of $\log a_j$ is the costliest substitution on ScreenSpot-v2 ($-10.53\%$).
In contrast, replacing max-pooling with log-mean-exp is the costliest on ScreenSpot-Pro ($-6.45\%$).
Collapsing $\mathbf U$ to its mean row costs $8.81\%$ and $5.63\%$ on the two suites.
These results show that the adopted scoring form is not interchangeable.
Alg.~\ref{alg:nested} then commits the resulting prefixes before prefill.
The next stage repairs spatial coverage while preserving the nested sets.
\begin{table}[t]
    \centering
    \caption{Scoring-form ablation inside \nwe{} at $\keepr{=}10\%$. $\Delta$ is accuracy minus the adopted form.}
    \label{tab:neo-form}
    \adjustbox{max width=0.86\textwidth,center}{%
        \footnotesize
        \setlength{\tabcolsep}{8pt}
        \renewcommand{\arraystretch}{0.98}
        \begin{tabular}{ll cc cc}
            \toprule
            & & \multicolumn{2}{c}{\textbf{ScreenSpot-v2}} & \multicolumn{2}{c}{\textbf{ScreenSpot-Pro}} \\
            \cmidrule(lr){3-4}\cmidrule(lr){5-6}
            \textbf{Substitution} & \textbf{Replaces} & Acc & $\Delta$ & Acc & $\Delta$ \\
            \midrule
            \rowcolor{rowblue}adopted rule & --- & \textbf{73.90} & --- & \textbf{37.63} & --- \\
            \midrule
            $2\log a_j+\log d_j^2$ & unit likelihood coefficient & 63.36 & $-10.53$ & 34.03 & $-3.61$ \\
            log-mean-exp pooling & max over query rows & 65.41 & $-8.49$ & 31.18 & $-6.45$ \\
            mean row only & instruction-token rows in $\mathbf U$ & 65.09 & $-8.81$ & 32.01 & $-5.63$ \\
            \bottomrule
        \end{tabular}
    }
\end{table}

\begin{algorithm}[tpb]
\caption{Admission-time Nested Selection for One Frame}
\label{alg:nested}
\begin{algorithmic}[1]
\STATE \textbf{Input:} tokens $\mathbf{E}\in\mathbb{R}^{N\times D}$, instruction matrix $\mathbf U$, detections, budgets $k_h\leq k_c$, doses $\rhocur$ and $\rhohist$
\STATE \textbf{Output:} admitted sets $S_t^{\mathrm{hist}}\subseteq S_t^{\mathrm{cur}}$
\STATE Rasterize detections into $\mathbf{p}$, compute $\alpha_t^\star$, and set $m_j\leftarrow 1+\alpha_t^\star Np_j$ \hfill (Eq.~\ref{eq:alpha-star})
\STATE Normalize features to $\mathbf{z}_j$, compute $a_j$, and set $\boldsymbol\psi_j\leftarrow\sqrt{m_j}\mathbf{z}_j$ \hfill (Eq.~\ref{eq:likelihood})
\STATE Apply greedy matching pursuit to obtain a length-$k_c$ sequence $\pi$ \hfill (Eq.~\ref{eq:qd})
\STATE Set $g_c\leftarrow\lceil\rhocur k_c\rceil$ and replace the $g_c$-tail of $\pi$ with stride representatives
\STATE Set $g_h\leftarrow\lceil\rhohist k_h\rceil$ and protect $\pi[1{:}k_h-g_h]$
\STATE Partition the remaining full-frame complement and select medoids $\{\mu_b\}_{b=1}^{g_h}$ \hfill (Eq.~\ref{eq:medoid})
\STATE Remove medoids from $\pi[k_h-g_h{:}]$ and insert them after the protected prefix
\STATE Evict the lowest-ranked eligible tail tokens needed to keep length $k_c$
\STATE \textbf{return} $S_t^{\mathrm{cur}}=\mathrm{set}(\pi[1{:}k_c])$ and $S_t^{\mathrm{hist}}=\mathrm{set}(\pi[1{:}k_h])$
\end{algorithmic}
\end{algorithm}

\noindent\textbf{Native-token Coverage Repair.}
Condition (iii) of Eq.~\ref{eq:lifecycle} keeps every repaired token as an original cache row.
Condition (ii) keeps the history set nested in the current set.
The current-frame stride on $U_c$ is specified in \S\ref{sec:method-c}.
We next turn to the history pass, explaining the detailed partition process.
\noindent\cnum{1}~\textit{History Representative.}
When a frame is retired, we keep only $k_h$ native rows instead of $k_c$, where $k_h<k_c$.
Each retained token should therefore represent a local region that may be relevant to future queries.
A centroid of that neighborhood would be synthetic and would violate condition (iii).
Thus, we choose the native token that minimizes within-region squared feature distance as the medoid.
Specifically, \awc{}-H protects a prefix of length $k_h-g_h$ and partitions the raster-ordered complement into $g_h$ contiguous regions.
From each region $R_b$ it keeps the medoid as defined below.
\begin{equation}
    \mu_b
    =
    \arg\min_{j\in R_b}
    \sum_{l\in R_b}
    \bigl\lVert\mathbf h_j-\mathbf h_l\bigr\rVert_2^2
    \label{eq:medoid}
\end{equation}
where $\mathbf h_j$ is a renormalized stride slice of at most $256$ channels of $\mathbf z_j$.
Inserting these medoids after the protected prefix yields the nested history subset with the following equation:
\begin{equation}
    S_t^{\mathrm{hist}}
    =
    \{\pi_1,\ldots,\pi_{k_h-g_h}\}
    \cup
    \{\mu_1,\ldots,\mu_{g_h}\}
    \subseteq S_t^{\mathrm{cur}},
    \label{eq:admit}
\end{equation}
so every element of $S_t^{\mathrm{hist}}$ already corresponds to an existing row in $S_t^{\mathrm{cur}}$.
The subset relation makes retirement a deletion-only operation that removes $S_t^{\mathrm{cur}}\setminus S_t^{\mathrm{hist}}$.
This preserves the exact encoder features and avoids synthetic features or new KV rows that need to be reconstructed from scratch.
\\
\noindent\cnum{2}~\textit{History Partition.}
The regions are not an arbitrary split of the complement.
Restricting region sizes to $\{\lfloor L/g_h\rfloor,\lceil L/g_h\rceil\}$ over an $L$-token complement turns boundary placement into choosing which $r=L\bmod g_h$ regions receive the extra token.
A dynamic program places those boundaries to minimize the maximum within-region sum of squared distances to the region mean.
It has $\mathcal{O}(g_h r)$ states and transitions, each evaluated in $\mathcal{O}(1)$ from prefix sums of $\mathbf{h}_j$ and $\lVert\mathbf{h}_j\rVert^2$.
Preparing the sums and extracting medoids costs $\mathcal{O}(L\tilde{d})$ with $\tilde{d}\leq 256$.
Alg.~\ref{alg:medoid} specifies this partition in detail.
\\
\noindent\cnum{3}~\textit{Repair Rules.}
The current-frame repair could use the same medoid rule, and the history repair could use stride instead.
Tab.~\ref{tab:ablation-ncr} in the main text compares the representative rules at each pass.
On the current frame, stride reaches $67.92\%$ on ScreenSpot-v2 and $37.63\%$ on ScreenSpot-Pro, while medoids reach $58.65\%$ and $30.11\%$.
In contrast, the preference reverses on the history frame.
Medoids reach $24.04\%$ on OmniGUI against $23.20\%$ for stride, and Mind2Web differs by $0.02\%$.
The current budget is large enough that the binding risk is a residual spatial gap, and equally spaced stride positions close such gaps at the lowest cost per token.
The history budget is far tighter, so each surviving representative must summarize an entire region.
In that case, the native token closest to its region in feature space is a better single prototype than a position fixed by geometry alone.
Thus, the larger current budget is better spent on spatial stride, whereas the tighter history budget is better spent on native region prototypes.
These results show that the two repairs should use different representative rules.
Alg.~\ref{alg:nested} then commits both repaired prefixes for effective spatial coverage.
\begin{algorithm}[tpb]
\caption{History Medoid Partition (\awc{}-H)}
\label{alg:medoid}
\begin{algorithmic}[1]
\STATE \textbf{Input:} normalized frame features $\{\mathbf z_j\}_{j=1}^{N}$, protected prefix $P_h=\pi[1{:}k_h-g_h]$, region count $g_h$
\STATE \textbf{Output:} real-token medoids $\{\mu_b\}_{b=1}^{g_h}$
\STATE Form the raster-ordered complement $L=[N]\setminus P_h$
\STATE For each $j\in L$, take a deterministic stride slice of at most $256$ channels of $\mathbf z_j$ and renormalize it to $\mathbf h_j$
\STATE Precompute prefix sums of $\mathbf h_j$ and $\lVert\mathbf h_j\rVert^2$ along $L$
\STATE The dynamic program uses contiguous boundaries with region sizes in $\{\lfloor|L|/g_h\rfloor,\lceil|L|/g_h\rceil\}$. It minimizes the maximum within-region sum of squared distances to the region mean
\STATE \textbf{return} the medoid $\mu_b$ of each region $B_b$ under Eq.~\ref{eq:medoid}
\end{algorithmic}
\end{algorithm}
\\
\noindent\textbf{Monotone KV Contraction.}
\S\ref{sec:method-kv} already specifies how a retiring frame is cropped to the first $k_h$ positions of $\pi$ and how those native rows are replayed with the incoming step.
We next explain how contraction preserves the original positions of native tokens and reduces the serving cost.
\\
\noindent\cnum{1}~\textit{Cache Contraction.}
Each frame occupies one contiguous visual span in the cache, and the admitted order fixes $S_t^{\mathrm{hist}}$ as row indices within that span.
Specifically, text and action rows are untouched.
Positional indices are never re-derived from the compacted sequence.
Before the first LLM prefill, \kvr{} assigns multimodal RoPE indices on the unpruned sequence and records the original-position of every admitted row.
Since contraction only removes rows, the remaining rows keep their original positions.
Besides, rotary offsets for subsequent text are recomputed from the surviving position.
\\
\noindent\cnum{2}~\textit{Serving Cost.}
The serving cost has three parts, namely visual encoding, LLM prefill, and persistent visual-cache storage.
After initialization, each transition encodes one new screenshot and retires the previous current frame.
Retirement only removes rows from the stored order of that frame.
The merged LLM forward therefore receives the non-visual prefix, followed by the $k_h$ retained historical tokens and the $k_c$ tokens admitted from the current screenshot.
The resulting lengths are as follows:
\begin{equation}
L_t^{\mathrm{enc}}=k_c,\qquad
L_t^{\mathrm{KV}}=T_t+k_h+k_c,\qquad
|\mathrm{KV}_{\mathrm{visual}}|=\mathcal{O}(k_c+Hk_h).
\label{eq:accounting}
\end{equation}
Here, $T_t$ is the prefix length at step $t$, and $H$ is the number of retained historical frames.
The first equality counts the visual tokens supplied by the new screenshot.
The second equality counts the tokens processed by the merged LLM forward.
The last equality counts only visual KV rows and excludes the non-visual history.
The replayed $k_h$ rows reuse features encoded when their frame first arrived.
They add only the retained $k_h$ visual tokens to the LLM prefill.
Because prefill processes these tokens in parallel rather than autoregressively, their additional latency is small, while reusing the original encoded features preserves feature fidelity.
This replaces the repeated $\mathcal{O}(HN)$ visual encoding of dense historical frames.
Each retained historical frame occupies $k_h$ visual rows instead of $N$, while the visual cache grows as $\mathcal{O}(k_c+Hk_h)$.
Under our multi-step evaluation protocol on OmniGUI, one contraction costs $6.4$\ ms per transition, compared with $90.8$\ ms for encoding one screenshot.
Thus, restoring a retired frame is much cheaper than encoding it again.
Meanwhile, the replayed set is the history prefix returned by Alg.~\ref{alg:nested}, which satisfies condition (ii) of Eq.~\ref{eq:lifecycle}.
However, the above feature reuse guarantees nesting but not downstream context consistency after contraction.
\\
\noindent\cnum{3}~\textit{Context-consistent replay.}
The central issue is the interaction between visual-token deletion and the downstream text and action states already stored in the cache.
In a deletion-only update, these downstream rows may have been computed while the retiring frame still contained visual rows that will later be removed.
Keeping those rows after deletion creates a mixed context.
New queries then attend to visual rows from the shortened context while also attending to text or action rows computed from the original context.
Such stale cross-modal states can make the evidence inconsistent and may contribute to confusion or hallucination in later steps.
Our contraction removes this mismatch by truncating the cache at the retiring frame boundary before appending the next step.
It keeps only the selected $k_h$ visual rows, so the merged forward adds only a small number of visual tokens.
These rows are processed in parallel with the downstream text and action rows during prefill.
The extra computation and latency therefore remain small, while the downstream states are aligned with the contracted context.
Every downstream state is therefore computed against the same visual context that will remain available during subsequent decoding.
At the same time, the retained visual features are reused exactly, and no synthetic visual representation is introduced.
The resulting design provides \emph{context-consistent replay} and \emph{exact feature reuse}.
It does not claim exact equivalence to a dense recomputation after visual rows are removed, because the dense computation includes those removed rows in the context.
At the $100\%$ budget, no visual rows are removed.
With this condition, the byte-exact no-op test in Appendix~\ref{app:protocol} verifies equivalence with the dense computation.
At pruned budgets, we evaluate the lifecycle and report its accuracy consequences in Appendix~\ref{app:evidence}.

\subsection{Evaluation Protocol and Baseline Reproduction}
\label{app:protocol}
\S\ref{sec:exp-setup} already names the backbones, the six evaluated benchmarks, and the matched visual-token budgets.
This subsection specifies the scoring rules and method adaptations that are reproducible.
\\
\noindent\textbf{Baseline Setting.}
We evaluate nine leading training-free methods against random and uniform pruning.
These methods span diversity, relevance, saliency, merging, posterior attention, sensitivity, and semantic grouping.
Every method follows its official selection settings and target budget.
\\
\noindent\textbf{Scoring Protocol.}
\S\ref{sec:exp-setup} counts a step as correct only when both the action type and its target or argument are correct.
We adopt the reference protocol's click threshold for every reported result.
For click and long-press actions, the predicted point must lie within a normalized euclidean distance of $0.04$ from the gold coordinate after both points are normalized by the screen dimensions.
Text and scroll arguments follow the reference protocol's string and direction rules.
\\
\noindent\textbf{Implementation Fidelity.}
Each existing method is re-implemented from its official repository.
VisPruner, PruMerge+, and the global term of VisionTrim use true final-block ViT attention.
FastV is implemented inline in the LLM according to its official protocol.
Merge methods use the declared real-token representative rule, and every method is held to an exact keep budget.
Five checks establish mechanical correctness.
\noindent\textit{(1) No-op test.}
A byte-exact no-op test verifies that every pruning run at a $100\%$ budget reproduces the unpruned responses across probe documents.
\noindent\textit{(2) Unit tests.}
Unit tests cover selector invariants, index and RoPE consistency, and per-method oracle validation against the official selection blocks.
\noindent\textit{(3) Exact-budget check.}
An exact-budget check verifies $|kept-\lceil rN\rceil|\leq1$ on every probe document.
\noindent\textit{(4) Position consistency check.}
A position consistency check confirms that all methods preserve the same original-position convention.
\noindent\textit{(5) Multi-step ledger check.}
A multi-step ledger check reconstructs each realized $(\keepc,\keeph)$ from its per-step kept-token ledger and compares it with the nominal budget.
Every experiment in the main table passes the above checks.
\\
\noindent\textbf{Excluded Methods.}
PyramidDrop~\citep{xing2025pdrop} was evaluated but excluded from the comparison tables.
Its pyramid schedule leaves shallow layers unpruned, so its effective pruning is approximately zero at our nominal budgets.
The latency of PyramidDrop is nearly identical to the unpruned model.
This is not a comparable setting.
IVC-Prune~\citep{sun2026ivcprune} is excluded for the same reason.
It prunes at layer $22$ of the LLM, so the vision encoder, all shallow layers, and most prefill FLOPs and KV rows still operate at dense length.
Because these nominal ratios are not comparable with other methods' budgets, we exclude both methods for fair comparisons.
\\
\noindent\textbf{Baseline Sensitivity.}
Official hyperparameters of the compared methods may not be optimal for GUI screens.
Thus, we separately test one hyperparameter for four representative baselines on ScreenSpot-v2 at $\keepr\!=\!10\%$.
Specifically, VisPruner improves from $46.07\%$ to $58.81\%$ as its importance ratio increases over $\{0.25,0.50 (\text{ Official }),0.75,0.90\}$.
The official equal importance and diversity split is therefore not optimal on this GUI suite.
This supports the main-text observation that generic diversity is weaker than attention importance on these screens.
VisionTrim also improves as its DVTS share increases from $0.5$ to the official $0.7$ and then to $0.9$, reaching $44.26\% \to 58.65\% \to 61.24\%$.
PruneSID gives $61.01\% \to 58.96\% \to 58.49\%$ as its NMS threshold scale changes over $\{0.7,1.0 ( \text{ Official }),1.3\}$.
ZOO-Prune remains within a narrow range of $57.47\% \to 56.29\% \to 55.90\%$ as its direction count changes over $\{16,64 (\text{ Official }),256\}$.
The lowest direction count gives the best score among these settings, while larger counts mildly reduce accuracy.
Among the tuned baselines, VisionTrim at a DVTS share of $0.9$ is strongest, reaching $61.24\%$.
Despite this improvement, VisPruner remains $12.66$\% below \method{} at $73.90\%$.
These results further demonstrate the robustness of our proposed \method{} under diverse comparisons.
\subsection{Configuration Validation}
\label{app:hyper}
This section first defines the configuration variables and then evaluates their effect across tasks and budgets.
We begin with the prior rule and selected scalar grids.
The following subsections test whether these choices remain useful when the task or budget changes.
\noindent\textit{Prior strength and support calibration.}
\label{app:alpha}
For each benchmark and budget, we select the prior cap $\bar{\alpha}$ from the public grid $\{1,2,4,6\}$.
An adopted run either uses this cap or applies the effective-support rule in Eq.~\ref{eq:alpha-star}.
Tab.~\ref{tab:alpha-sens} reports the results across this grid.
The setting with $\alpha=0$ appears in the row without the prior in Tab.~\ref{tab:ablation-neo}.
For a normalized prior $\mathbf p$ over the $N$ frame tokens, define
$m_j(\alpha)=1+\alpha Np_j$ and
$N_{\mathrm{eff}}(\alpha)=\bigl(\sum_j m_j(\alpha)\bigr)^2\!/\sum_j m_j^2(\alpha)$.
The quantity $N_{\mathrm{eff}}$ acts as an effective token count.
If $r$ tokens have equal mass $c$ and all other tokens have zero mass, then
the numerator is $(rc)^2$ and the denominator is $rc^2$.
The ratio is therefore $r$.
The squared sum in the numerator makes the quantity independent of the overall mass scale.
The squared masses in the denominator increase when the mass concentrates on a few tokens.
Uniform mass over $N$ tokens gives $N_{\mathrm{eff}}=N$.
Concentration on one token gives $N_{\mathrm{eff}}\approx1$.
The unit term in $m_j(\alpha)$ keeps every token eligible.
The term $\alpha Np_j$ increases the mass of tokens in regions indicated by the interaction prior.
This bias favors operable regions.
An excessive value of $\alpha$ can concentrate the order too narrowly and remove context that later actions may require.
For a given cap, $\alpha_t^\star$ is the largest strength defined as follows:
\begin{equation}
\alpha_t^\star
=
\max\left\{
0\leq\alpha\leq\bar{\alpha}
\ \middle|\
\frac{\left(\sum_j m_j(\alpha)\right)^2}{\sum_j m_j^2(\alpha)}
\geq \zeta k
\right\},
\qquad \zeta=1 .
\label{eq:alpha-star}
\end{equation}

The applied strength is $\min(\bar{\alpha},\alpha^{\star})$.
Frames without detector support use no prior.
\noindent\textit{Selected scalars.}
With the prior rule fixed, each result of \method{} in the main tables uses one configuration for its dataset and budget.
The declared grid selects the prior cap, the calibration mode, and the current dose.
Multi-step runs also select one history dose.
The current frame pass uses $\rhocur$ for both single-step and multi-step tasks.
The history pass uses the independent dose $\rhohist$ only when $k_h<k_c$.
The doses are selected from fixed public grids $\rhohist\in\{0.05,0.10,0.20\}$ and $\rhocur\in\{0,0.2,0.3,0.5\}$.

\subsubsection{Fixed-Configuration Check}
\label{app:calibration}
We first evaluate whether each benchmark and budget needs its own configuration.
Specifically, we evaluate one global configuration with $\alpha=2$ and the default applicable dose at $16$ benchmark-budget combinations across all six benchmarks.
The six single-step results at the mild budget in Tab.~\ref{tab:global-fixed} use the corresponding $\alpha\!=\!2$ sensitivity estimates.
The gap between the two configurations stays within $1.9\%$.
The selected configuration improves the median result by $+1.0\%$ and improves 14 of the 16 combinations by less than $+2\%$.
The fixed configuration is ahead by $0.3\%$ on MMBench at $\keepr=5\%$.
The small aggregate gap shows that one fixed setting is often sufficient, while the larger differences identify where benchmark-specific selection matters.
The Mind2Web Cross-Website split gains $3.0\%$ at $\keepc=25\%$, where the strongest prior is most useful.
AndroidControl gains $4.1\%$ at the tight budget, where the heavier current dose supplies the main improvement.
The fixed setting omits the heavy current dose required by this sparse-UI domain.
Its $\alpha$ response is flat in Tab.~\ref{tab:alpha-sens}.
The result is therefore dose-specific.
Besides this dose-specific result, we rely on Tab.~\ref{tab:global-fixed} for the broader comparison with existing methods.
Against the strongest existing methods, the fixed setting remains close on five of the six benchmarks.
Separate selection improves the result on most benchmarks, while AndroidControl remains sensitive to the current dose.
The comparison therefore supports per-benchmark configuration without implying that one prior strength is best for every task.
\begin{table}[t]
    \centering
    \caption{One global configuration compared with the per-benchmark-budget configuration. Fixed uses $\alpha{=}2$ and the default current dose. Selected uses the adopted cap, calibration mode and doses.}
    \label{tab:global-fixed}
    \adjustbox{max width=0.86\textwidth,center}{%
        \footnotesize
        \setlength{\tabcolsep}{7pt}
        \renewcommand{\arraystretch}{1.2}
        \begin{tabular}{llccc}
            \toprule
            \textbf{Benchmark (metric)} & \textbf{Budget} & \textbf{Fixed} & \textbf{Selected} & $\Delta$ \\
            \midrule
            \rowcolor{headergray}\multicolumn{5}{c}{\textsc{Single-step (grounding / choice accuracy)}} \\
            ScreenSpot-v2 (Acc) & $\keepr{=}50\%$ & 92.85 & 93.08 & $+0.23$ \\
            ScreenSpot-v2 (Acc) & $\keepr{=}10\%$ & 73.66 & 73.90 & $+0.24$ \\
            ScreenSpot-Pro (Acc) & $\keepr{=}10\%$ & 35.80 & 37.63 & $+1.83$ \\
            MMBench-GUI L2 (Acc) & $\keepr{=}25\%$ & 71.09 & 72.45 & $+1.36$ \\
            MMBench-GUI L2 (Acc) & $\keepr{=}10\%$ & 48.86 & 50.47 & $+1.61$ \\
            MMBench-GUI L2 (Acc) & $\keepr{=}5\%$ & 31.19 & 30.88 & $-0.31$ \\
            \midrule
            \rowcolor{headergray}\multicolumn{5}{c}{\textsc{Multi-step (Step~SR)}} \\
            OmniGUI (Step~SR) & $\keepc{=}50\%$ & 48.25 & 48.91 & $+0.66$ \\
            OmniGUI (Step~SR) & $\keepc{=}25\%$ & 42.65 & 43.08 & $+0.43$ \\
            Mind2Web Cross-Task (Step SR) & $\keepc{=}50\%$ & 45.72 & 46.86 & $+1.14$ \\
            Mind2Web Cross-Task (Step~SR) & $\keepc{=}25\%$ & 32.22 & 33.96 & $+1.74$ \\
            Mind2Web Cross-Domain (Step~SR) & $\keepc{=}50\%$ & 44.28 & 45.22 & $+0.94$ \\
            Mind2Web Cross-Domain (Step~SR) & $\keepc{=}25\%$ & 32.86 & 34.59 & $+1.73$ \\
            Mind2Web Cross-Website (Step~SR) & $\keepc{=}50\%$ & 43.93 & 44.75 & $+0.82$ \\
            Mind2Web Cross-Website (Step~SR) & $\keepc{=}25\%$ & 30.01 & 32.99 & $+2.98$ \\
            AndroidControl (Step~SR) & $\keepc{=}50\%$ & 59.60 & 60.20 & $+0.60$ \\
            AndroidControl (Step~SR) & $\keepc{=}25\%$ & 52.97 & 57.06 & $+4.09$ \\
            \bottomrule
        \end{tabular}
    }
\end{table}

\subsubsection{Benchmark and Budget Dependence}
\noindent\cnum{1}~\textit{Task-level pattern.}
We investigate the task-level pattern of the preferred prior strength.
To be specific, candidate-based selection chooses from a finite set of candidate elements or answers.
Coordinate grounding predicts a location on the screenshot, while structured action prediction emits an action and its required arguments.
In this comparison, Mind2Web Task uses candidate-based selection. 
MMBench-GUI L2, ScreenSpot-v2, and ScreenSpot-Pro form the coordinate-grounding family. 
OmniGUI and AndroidControl form the structured-action family.
The preferred prior cap follows the task type and then varies with the budget.
Candidate-based selection favors strong caps in $\{4,6\}$, while coordinate grounding and structured action prediction favor milder caps because location and action prediction need surrounding context.
Further analyses are presented below.
\begin{table}[t]
    \centering
    \caption{Prior strength sensitivity with \textbf{GUI-Owl-1.5-8B}. Rows are grouped by task family. Candidate-based selection and coordinate grounding use the public grid $\bar{\alpha}\in\{1,2,4,6\}$ where available, while structured action prediction uses $\bar{\alpha}\in\{1,2,4\}$. Bold denotes the adopted settings.}
    \label{tab:alpha-sens}
    \footnotesize
    \setlength{\tabcolsep}{7pt}
    \renewcommand{\arraystretch}{0.98}
    \adjustbox{max width=\textwidth,center}{%
        \begin{tabular}{llcccc}
            \toprule
            \textbf{Benchmark (metric)} & \textbf{Budget} & $\bar{\alpha}{=}1$ & $\bar{\alpha}{=}2$ & $\bar{\alpha}{=}4$ & $\bar{\alpha}{=}6$ \\
            \midrule
            \rowcolor{headergray}\multicolumn{6}{c}{\textsc{Candidate-based selection}} \\
            Mind2Web Task (Step~SR) & $\keepc{=}50\%$ & 43.68 & 45.22 & 46.22 & \textbf{46.86} \\
            Mind2Web Task (Step~SR) & $\keepc{=}25\%$ & 31.18 & 32.22 & 33.81 & \textbf{33.96} \\
            \bottomrule
        \end{tabular}
    }
    \par\vspace{4pt}
    \adjustbox{max width=\textwidth,center}{%
        \begin{tabular}{llcccc}
            \toprule
            \textbf{Benchmark (metric)} & \textbf{Budget} & $\bar{\alpha}{=}1$ & $\bar{\alpha}{=}2$ & $\bar{\alpha}{=}4$ & $\bar{\alpha}{=}6$ \\
            \midrule
            \rowcolor{headergray}\multicolumn{6}{c}{\textsc{Coordinate grounding}} \\
            MMBench-GUI L2 (Acc) & $\keepr{=}50\%$ & 80.02 & 80.66 & \textbf{81.39} & 79.97 \\
            MMBench-GUI L2 (Acc) & $\keepr{=}25\%$ & 71.06 & 72.15 & 72.34 & \textbf{72.45} \\
            ScreenSpot-v2 (Acc) & $\keepr{=}50\%$ & 93.00 & \textbf{93.08} & 92.77 & \text{N/A} \\
            ScreenSpot-v2 (Acc) & $\keepr{=}25\%$ & 89.15 & \textbf{90.33} & 89.86 & \text{N/A} \\
            ScreenSpot-v2 (Acc) & $\keepr{=}10\%$ & \textbf{73.90} & 73.35 & 72.80 & \text{N/A} \\
            ScreenSpot-v2 (Acc) & $\keepr{=}5\%$ & 52.52 & \textbf{55.19} & 54.17 & \text{N/A} \\
            ScreenSpot-Pro (Acc) & $\keepr{=}50\%$ & 66.79 & 65.40 & \textbf{67.30} & \text{N/A} \\
            ScreenSpot-Pro (Acc) & $\keepr{=}25\%$ & 57.18 & 57.50 & \textbf{57.50} & \text{N/A} \\
            ScreenSpot-Pro (Acc) & $\keepr{=}10\%$ & 36.37 & 36.18 & \textbf{37.63} & \text{N/A} \\
            ScreenSpot-Pro (Acc) & $\keepr{=}5\%$ & \textbf{20.30} & 19.29 & 17.39 & \text{N/A} \\
            \bottomrule
        \end{tabular}
    }
    \par\vspace{4pt}
    \adjustbox{max width=\textwidth,center}{%
        \begin{tabular}{llccc}
            \toprule
            \textbf{Benchmark (metric)} & \textbf{Budget} & $\bar{\alpha}{=}1$ & $\bar{\alpha}{=}2$ & $\bar{\alpha}{=}4$ \\
            \midrule
            \rowcolor{headergray}\multicolumn{5}{c}{\textsc{Structured action prediction}} \\
            OmniGUI (Step~SR) & $\keepc{=}50\%$ & \textbf{48.91} & 48.44 & 47.71 \\
            OmniGUI (Step~SR) & $\keepc{=}25\%$ & 42.11 & \textbf{42.65} & 42.38 \\
            AndroidControl (Step~SR) & $\keepc{=}50\%$ & \textbf{60.20} & 60.00 & 59.89 \\
            AndroidControl (Step~SR) & $\keepc{=}25\%$ & \textbf{56.58} & 56.62 & 56.23 \\
            \bottomrule
        \end{tabular}
    }
\end{table}

\begin{table}[t]
    \centering
    \caption{Interaction of prior strength $\bar{\alpha}$ and current dose $\rhocur$ on Mind2Web Cross-Task Step~SR.}
    \label{tab:alpha-dose-2d}
    \adjustbox{max width=0.86\textwidth,center}{%
        \footnotesize
        \setlength{\tabcolsep}{8pt}
        \renewcommand{\arraystretch}{0.98}
        \begin{tabular}{lcccccc}
            \toprule
            & \multicolumn{3}{c}{$\keepc{=}50\%$} & \multicolumn{3}{c}{$\keepc{=}25\%$} \\
            \cmidrule(lr){2-4}\cmidrule(lr){5-7}
            & $\rhocur{=}0$ & $\rhocur{=}0.3$ & $\rhocur{=}0.5$ & $\rhocur{=}0$ & $\rhocur{=}0.3$ & $\rhocur{=}0.5$ \\
            \midrule
            $\bar{\alpha}{=}1$ & 43.23 & 43.43 & 43.68 & 28.04 & 29.23 & 31.18 \\
            $\bar{\alpha}{=}2$ & 45.37 & 45.72 & 45.22 & 32.27 & 32.22 & 30.18 \\
            $\bar{\alpha}{=}4$ & 45.42 & 45.32 & \textbf{46.22} & \textbf{33.81} & 32.27 & 32.57 \\
            \bottomrule
        \end{tabular}
    }
\end{table}

\\
\noindent\cnum{2}~\textit{Configured-strength analysis.}
Tab.~\ref{tab:alpha-sens} shows that the useful cap depends on both the task and the budget.
On candidate-based selection, a stronger prior cap improves accuracy.
For example, Mind2Web Cross-Task Step~SR rises from $43.68\%$ to $46.86\%$ across the public grid, so the strongest cap is selected there.
MMBench-GUI instead peaks at an intermediate cap at $\keepr=50\%$ and declines at the strongest setting, showing that stronger concentration is not uniformly better even within candidate-based selection.
Within coordinate grounding, ScreenSpot-v2 favors mild caps and gains up to $+2.7\%$ at the tightest budget.
Meanwhile, ScreenSpot-Pro tolerates a stronger cap only when enough budget remains for context.
Within structured action prediction, OmniGUI changes little across caps.
Meanwhile, AndroidControl stays within $\pm0.4\%$ under the official protocol.
The selected configurations therefore follow task structure rather than a single global prior strength.
\begin{figure}[t]
    \centering
    \includegraphics[width=0.99\linewidth]{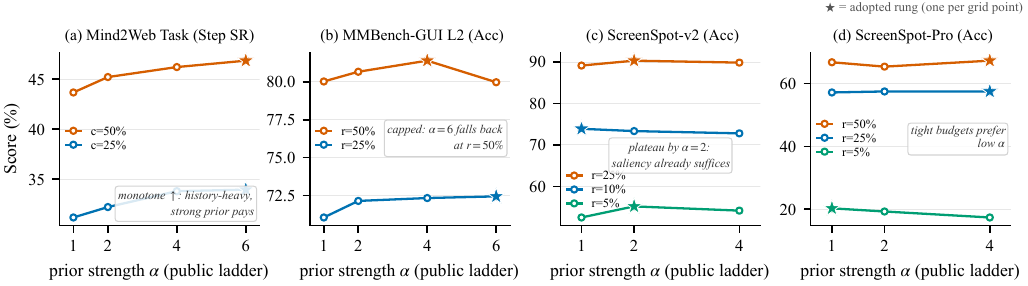}
    \caption{Prior strength analysis with \textbf{GUI-Owl-1.5-8B}. The filled marker is the adopted $\bar{\alpha}$.}
    \label{fig:alpha-grid}
    \end{figure}
    \begin{figure}[t]
    \centering
    \includegraphics[width=0.50\linewidth]{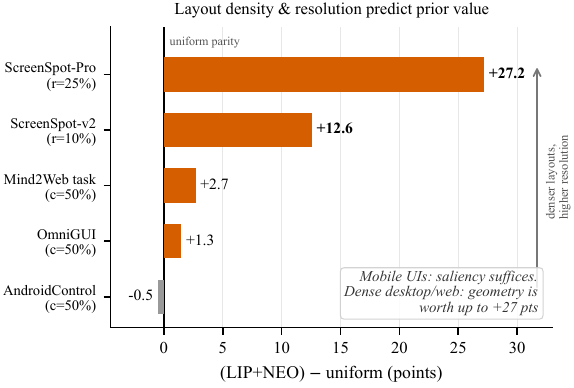}\hfill
    \includegraphics[width=0.48\linewidth]{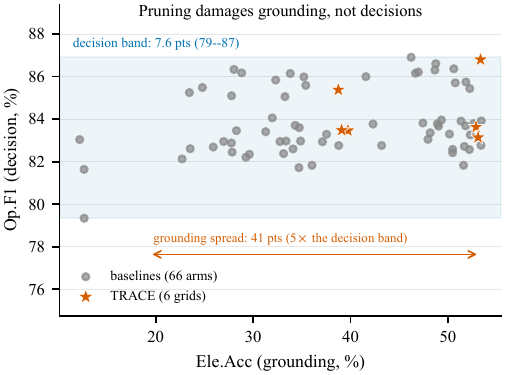}
    \caption{Prior value and damage localization with \textbf{GUI-Owl-1.5-8B}. The left panel shows the gain over uniform. The right panel shows Mind2Web Op.F1 versus Element-Accuracy.}
    \label{fig:spectrum-damage}
    \end{figure}
    \begin{figure}[t]
        \centering
        \includegraphics[width=0.48\linewidth, trim=0 115 0 0, clip]{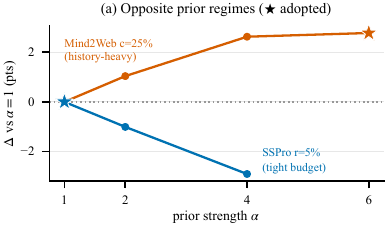}\hfill
        \includegraphics[width=0.48\linewidth, trim=0 0 0 115, clip]{figures/fig6_sens.pdf}
        \caption{Sensitivity analysis with \textbf{GUI-Owl-1.5-8B}. (a)~Prior strength. (b)~Current dose.}
        \label{fig:hyperparams}
        \end{figure}
\subsubsection{Prior and Coverage Interaction}
The prior choice remains stable when the coverage dose changes.
Tab.~\ref{tab:alpha-dose-2d} shows that strong caps are usually preferred, but their interaction with coverage depends on the budget.
At tight budgets, the prior can already cover the interactive surface and extra repair competes for the same tokens.
At milder budgets, the selected configuration combines a strong cap with a larger current dose because additional tokens remain available for coverage repair.
The prior identifies likely interactive regions, while repair protects regions that the concentrated order leaves exposed.
Details are as follows.
\\
\noindent\cnum{1}~\textit{Prior value and damage.}
Figs.~\ref{fig:alpha-grid} and~\ref{fig:spectrum-damage} show that the prior improves spatial grounding, but its benefit depends on the task and the available context.
The effective-support rule limits the applied strength when concentration would remove too much context.
The interaction prior contributes little over uniform on AndroidControl but produces a much larger gain on ScreenSpot-Pro.
On Mind2Web, the remaining variation appears mainly in element grounding.
Thus, the prior mainly helps the model locate the target. 
When later actions need more context, a milder prior is preferred.
\\
\noindent\cnum{2}~\textit{Sensitivity curves.}
Fig.~\ref{fig:hyperparams} summarizes the two parameters over their tested ranges.
The curves show that the current dose repairs coverage gaps left by greedy selection, but excessive repair approaches uniform sampling.
The prior and dose therefore address different failure modes.
The prior changes where evidence is concentrated, while the dose restores regions that concentration would otherwise discard.
These complementary effects demonstrate the robustness of our \method{}.
\\
\noindent\cnum{3}~\textit{Calibration diagnostics.}
In the $1{,}000$-frame OmniGUI dump, Eq.~\ref{eq:alpha-star} is active mainly at the mild budget and reaches the configured cap on $18.4\%$ of frames, compared with $70.3\%$ at the tight budget.
AndroidControl shows the same shift, with cap saturation increasing from $11.7\%$ to $70.6\%$.
The support rule therefore permits stronger priors only when the retained context remains sufficient.
\begin{figure}[t]
    \centering
    \includegraphics[width=0.90\linewidth]{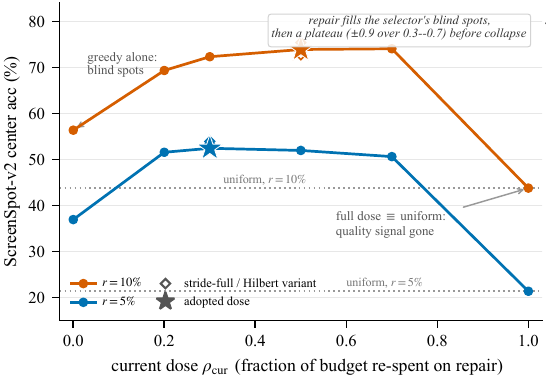}
    \caption{Dose response analysis on ScreenSpot-v2 with \textbf{GUI-Owl-1.5-8B}.}
    \label{fig:dose}
    \end{figure}
    \begin{figure}[t]
        \centering
        \includegraphics[width=1\linewidth]{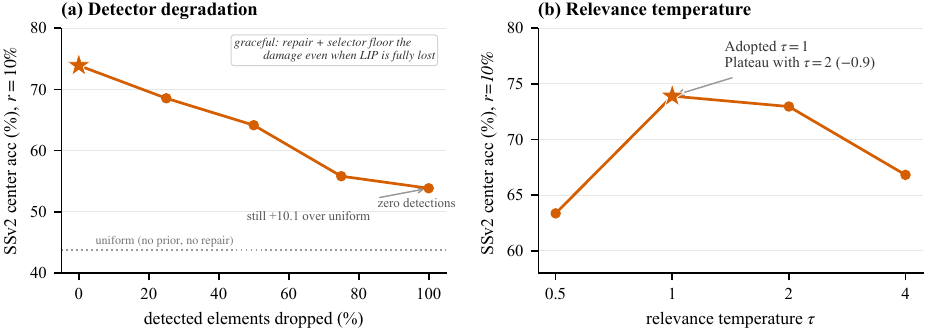}
        \caption{Robustness analysis of selector components with \textbf{GUI-Owl-1.5-8B} on ScreenSpot-v2 at $\keepr=10\%$. (a)~Detector degradation. (b)~Instruction-likelihood temperature.}
        \label{fig:robustness}
        \end{figure}
\subsubsection{Budget and Dose Sensitivity}
We next turn from prior strength to the split between current and history evidence.
On OmniGUI we hold the dose fixed and compare the two budgets in a $2{\times}2$ test.
Halving the current budget costs $5.29\%$, while halving the history budget costs $0.74\%$.
The current frame therefore carries most of the useful evidence, and the adopted setting keeps a larger current budget with a smaller history budget.
After this split is fixed, the dose spends part of each budget on covering tokens that greedy selection would drop.
Most multi-step benchmarks use a light dose.
Sparse AndroidControl is the exception and needs heavier repair on the current frame.
On the adopted OmniGUI tight setting, the history-dose grid changes Step~SR only slightly.
In contrast, the AndroidControl current-dose curve moves across the uniform baseline.
History retirement can therefore stay light, while the current dose must follow the domain.
Fig.~\ref{fig:dose} then shows how large that current dose should be on ScreenSpot-v2.
A moderate number of stride representatives improves the greedy keep.
A full dose removes the importance order and reduces the selector to uniform sampling.
Alternative scan rules stay within $1.5\%$ of the adopted dose.
The gain therefore comes from coverage repair itself rather than from a particular scan.
The adopted setting uses a larger current budget, a light history dose, and a current dose that repairs coverage gaps without collapsing to uniform sampling.
\subsubsection{Calibration and Robustness}
The previous paragraphs fix the prior, the budgets, and the doses.
We now test whether those choices remain usable when two inputs are imperfect.
Both tests use ScreenSpot-v2 at $\keepr=10\%$.
\\
\noindent\cnum{1}~\textit{Detector degradation.}
The first test removes detections at random.
Fig.~\ref{fig:robustness} (a) drops $25\%$, $50\%$, $75\%$, and $100\%$ of the boxes.
Accuracy falls from $73.90\%$ to $68.55\%$, $64.15\%$, $55.82\%$, and $53.85\%$.
The decrease is gradual rather than sudden.
When every detection is removed, the prior becomes flat and selection uses only instruction likelihood, feature diversity, and coverage repair.
That endpoint stays $10.1\%$ above uniform sampling.
The remaining channels therefore bound the damage, and the method does not collapse when the detector fails.
\noindent\cnum{2}~\textit{Instruction-likelihood temperature.}
The second test changes the temperature $\tau$ that scales the instruction-likelihood scores.
As shown in Fig.~\ref{fig:robustness} (b), the adopted $\tau=1$ lies on a plateau with $\tau=2$ and changes accuracy by only $0.9\%$.
Sharpening to $\tau=0.5$ concentrates relevance on too few tokens and loses $10.5\%$.
Flattening to $\tau=4$ dilutes the instruction signal and loses $7.1\%$.
We therefore keep the default $\tau=1$ in every experiment and do not tune it.
The selector remains usable when detections are incomplete and when the relevance scale is left at its default.
These analyses show that our \method{} is robust to imperfect inputs.
\section{Additional Experimental Evidence}
\label{app:evidence}
The previous section fixes the method, comparison protocol, and operating points. We now examine what is preserved when a visual frame is selected once and reused at a smaller history budget.
We follow four evidence chains. Lifecycle traces test later-target retention and nested retirement. Benchmark tables locate gains across budgets, metrics, categories, and model scales. Ablations connect those gains to selector components. Serving measurements separate cache reuse from selection cost.
Section~\ref{app:premise} examines admission and retirement, while Section~\ref{app:supp-results} expands the benchmark results.
Section~\ref{app:supp-ablation} combines component ablations with keep visualizations, and Section~\ref{app:efficiency} separates cache reuse from token-selection cost.
Each diagnostic answers a different question. Keep visualizations show spatial allocation on the displayed frame. Paired ablations measure score changes. Serving measurements report the evaluated execution path.
Detailed descriptions are as follows.
\subsection{Lifecycle Behavior}
\label{app:premise}
\begin{figure}[t]
\centering
\includegraphics[width=\linewidth]{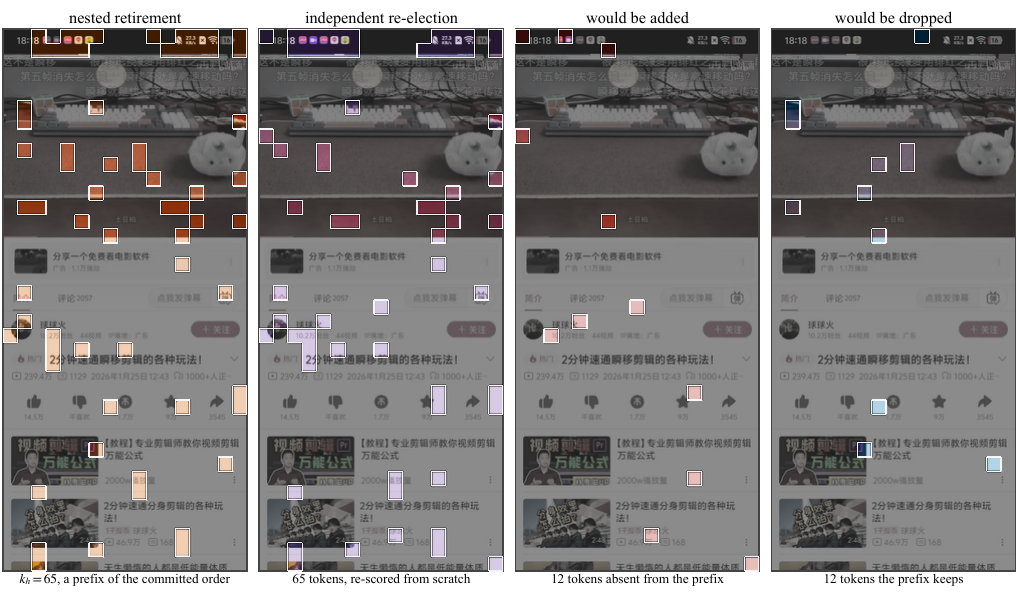}
\caption{Comparison of nested retirement against an unconstrained re-selection with \textbf{GUI-Owl-1.5-8B}. A fresh selection of $65$ tokens would add $12$ and drop $12$ that the committed prefix keeps. Over $1000$ steps the same re-selection moves $23.5\%$ of the surviving set.}
\label{fig:g-nested}
\end{figure}

\begin{figure}[t]
\centering
\includegraphics[width=1\linewidth]{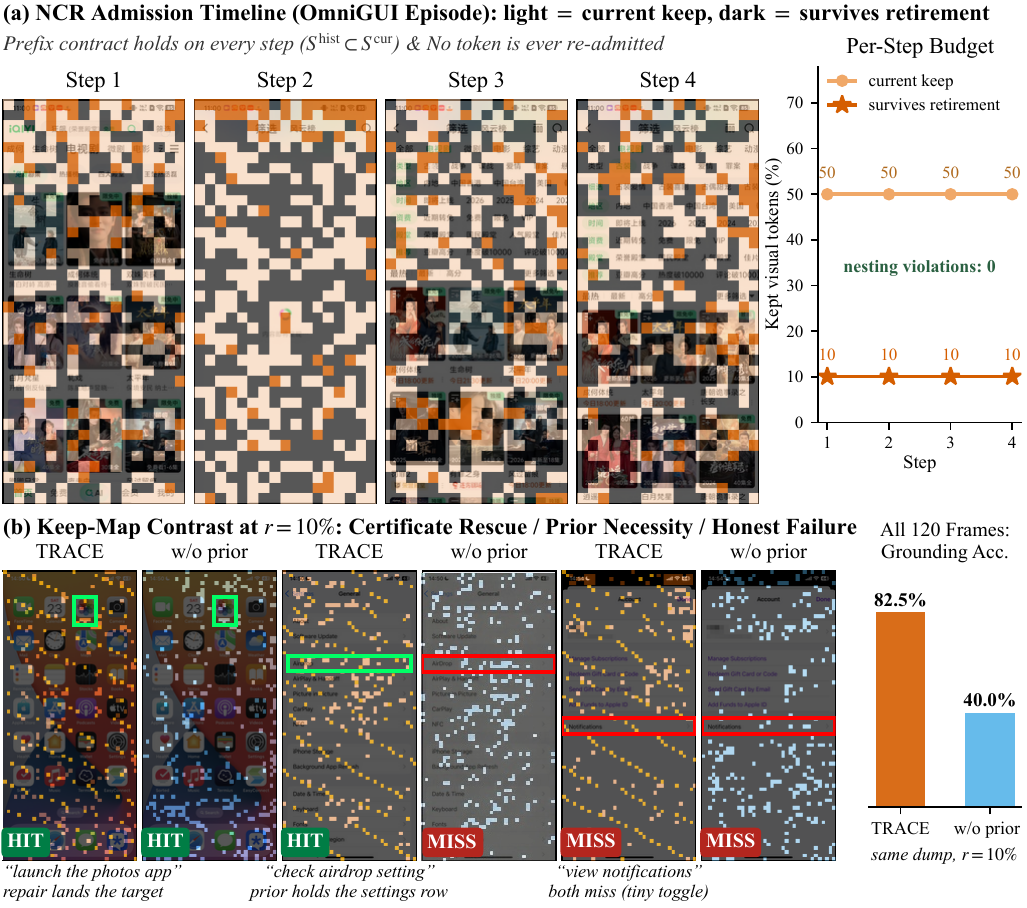}
\caption{Committed lifecycle with \textbf{GUI-Owl-1.5-8B}. (a)~An OmniGUI episode admits at about $50\%$ and retires to a $10\%$ prefix with zero nesting violations. (b)~On ScreenSpot-v2 at $\keepr=10\%$, the keep of \method{} grounds $82.5\%$ of frames against $40.0\%$ without the prior.}
\label{fig:admit-timeline}
\end{figure}

\begin{figure}[h]
\centering
\includegraphics[width=0.815\linewidth]{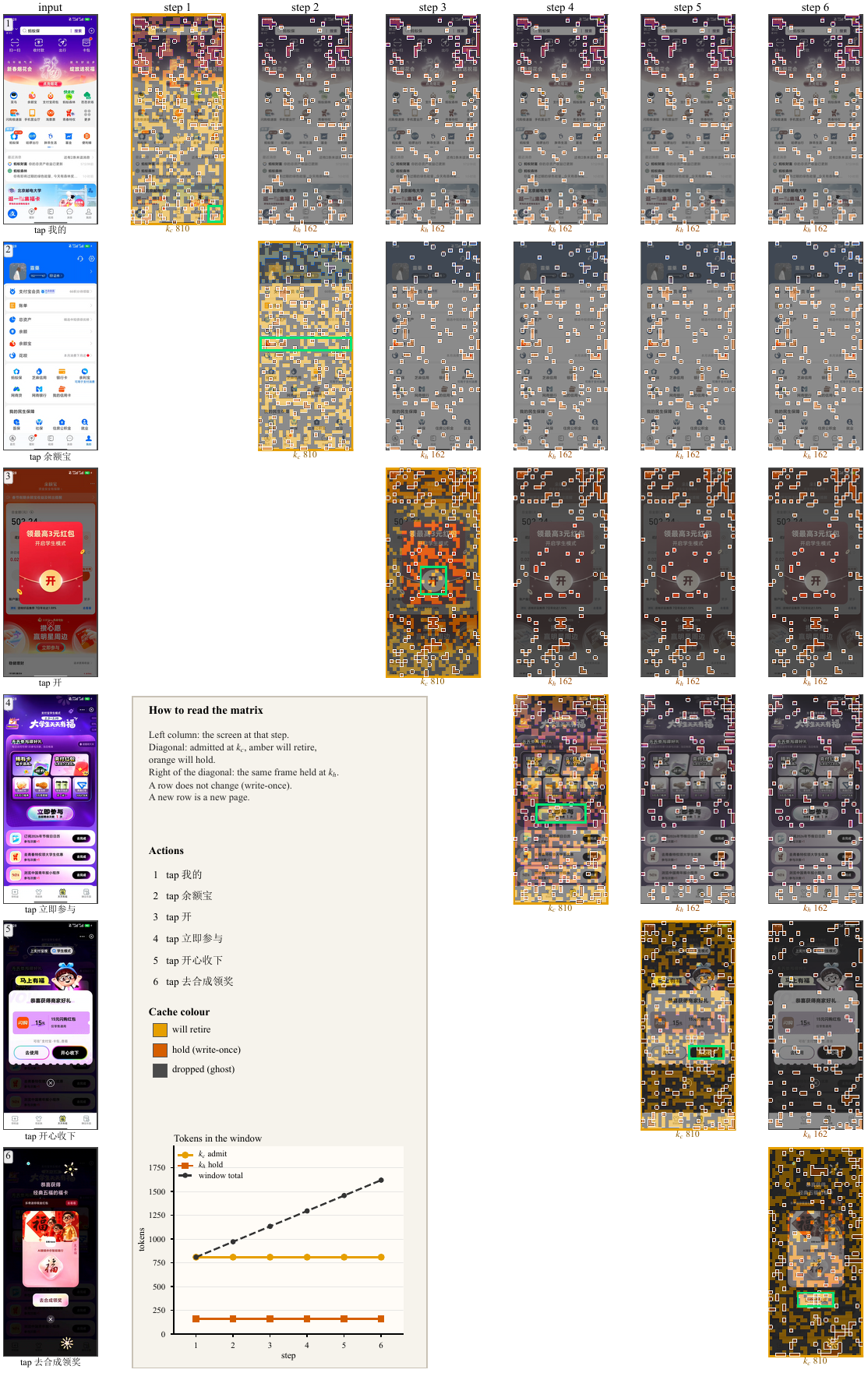}
\caption{Cache state of OmniGUI Alipay episode T0039 with \textbf{GUI-Owl-1.5-8B} at $(\keepc,\keeph)=(50\%,10\%)$. The left column is the input. The upper triangle is that frame's cache. The diagonal is admission. The same frame to the right of the diagonal is the contracted history.}
\label{fig:g-matrix}
\end{figure}

\begin{figure}[t]
\centering
\includegraphics[width=1\linewidth]{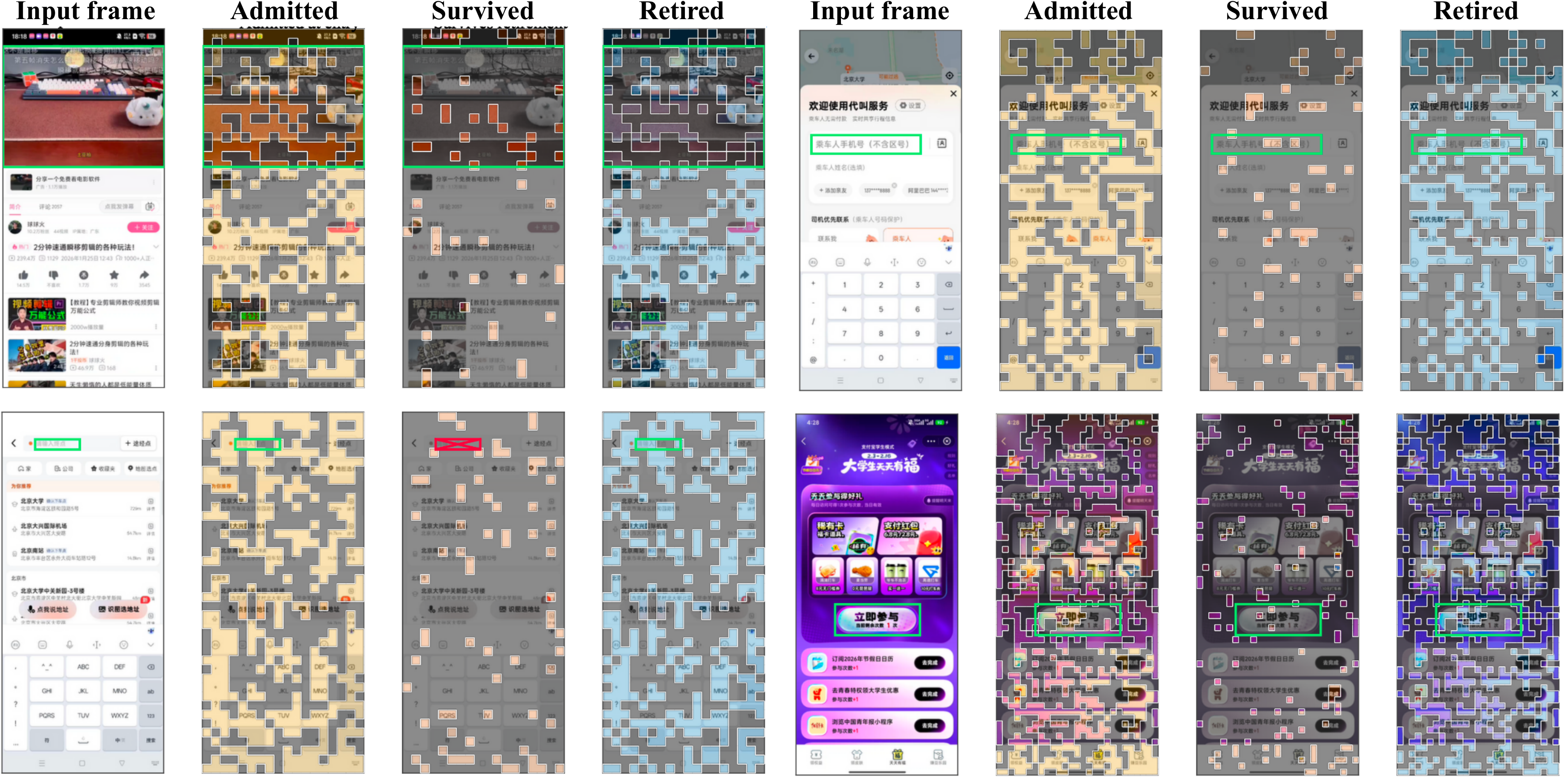}
\caption{Retirement dissected on four frames with \textbf{GUI-Owl-1.5-8B}. Each row is the input, the admitted keep, the surviving prefix, and the tokens retirement frees.}
\label{fig:g-retire}
\end{figure}

This subsection tests the two premises of Fig.~\ref{fig:evidence} on instrumented trajectories.
The first is that a keep written at first admission should contain later targets.
The second is that the history keep should be a prefix of the current keep to avoid re-encoding.
We first measure later-target retention. 
Then, we compare nested retirement with independent re-selection. Finally, we inspect the cache states.
\\
\noindent\cnum{1}~\textit{Admission Retention.}
Same-screen recurrences are identified by an exact content hash or a $64$-bit dhash within Hamming distance $4$ at the same resolution.
At the mild and tight budgets, \method{} retains $96.09\%$ and $91.41\%$ of the later targets, while existing methods perform near chance.
The relevant quantity is later-target retention rather than success on the instruction available at admission. A keep can serve the present action yet omit another widget on the same screen.
The layout prior reserves evidence beyond the action used at admission on these recurrences. It does not guarantee recovery of every later action.
Retirement poses the same question after the keep is contracted.
Admission error is the rate at which the contracted history loses the patch needed by a later action.
It is $12.37\%$ and $20.23\%$ for \method{} at the two budgets, against about $72\%$ for uniform.
The comparison uses the same retirement budgets, so token count alone does not explain the difference.
The retained prefix contains the later target more often, demonstrating the effectiveness of the layout prior.
\\
\noindent\cnum{2}~\textit{Nested Reuse.}
Independent re-selection then asks whether that prefix is forced by serving.
On OmniGUI it requests tokens outside the admitted cache on $99.5\%$ and $99.1\%$ of frames.
Those requests can be satisfied only by encoding the retired frame again.
Forcing the independent choice onto the surviving rows changes Step~SR by only $-0.7\%$ and $-0.4\%$.
Fig.~\ref{fig:g-nested} shows that restriction on one frame.
The first panel is the nested prefix of $65$ tokens.
The second panel is an independent re-selection of $65$ tokens scored from scratch.
The last two panels show the two sides of the symmetric difference: tokens selected only by the fresh selection and tokens selected only by the committed prefix.
Independent selection would add $12$ tokens that the prefix does not contain and would drop $12$ tokens that the prefix keeps.
Over the same $1000$ steps, fresh re-selection changes $23.5\%$ of the surviving set.
The nested order is therefore a serving constraint rather than a relabeling of independent selection.
In particular, equal token counts do not imply equal cache contents. The independent selection changes token identities, whereas contraction must use the order already committed.
\\
\noindent\cnum{3}~\textit{Committed Cache.}
The remaining figures follow one path.
They show the committed prefix, what one episode keeps after retirement, and how a single frame is contracted.
Selected patches keep the original pixels.
Discarded patches are desaturated so the dropped content stays visible.
Specifically, Fig.~\ref{fig:admit-timeline} (a) records one OmniGUI episode.
The light keep is current at about $50\%$.
The dark keep is the $10\%$ prefix that survives retirement.
The bars stay at those two levels on every step, and the figure records zero nesting violations.
No token is re-admitted after the frame is written, which is the prefix cut of Eq.~\ref{eq:admit}.
Fig.~\ref{fig:admit-timeline} (b) compares \method{} with the no-prior keep on ScreenSpot-v2 at $\keepr=10\%$.
When the instruction is to launch the photos app, both keeps hit.
When the instruction is to check the AirDrop setting, the prior holds the settings row and \method{} hits, while the keep without the prior misses.
When the instruction is to view notifications, both keeps miss a tiny toggle.
Over the $120$ frames, \method{} reaches $82.5\%$ grounding accuracy against $40.0\%$ without the prior.
The two keeps differ by the prior rather than by a second admission. 
Besides the above visual evidence, we read OmniGUI Alipay episode T0039 frame by step as shown in Fig.~\ref{fig:g-matrix}.
The left column is the input and the action on that screen.
The upper triangle is the cache.
On the diagonal the frame is admitted at $\keepc=50\%$ and split into tokens that will survive and tokens that will be evicted.
Every frame to the right of the diagonal is held after contraction to $\keeph=10\%$.
A row does not change after admission because history is write-once.
The episode opens Yu'ebao from Me and then completes the red-packet flow.
The six actions are tap Me, tap Yu'ebao, tap Open, tap Participate, tap Accept, and tap Claim reward.
The acted-on widget on each current-frame keep remains in the contracted keep of that row.
The cache keeps the acted-on evidence of each page rather than only the latest screenshot.
Fig.~\ref{fig:g-retire} then dissects four frames.
Each row shows the input, the admitted keep, the survivors, and the complement that retirement frees.
The surviving prefix is a subset of the admitted keep, while the released tokens form the complement of that prefix within the admitted set.
Thus, the survivor and released panels are disjoint, and together they reconstruct the admitted keep. Only the survivor is nested within the original selection.
The marked target remains in the survivor panel on the displayed frames. This illustrates what contraction preserves and frees without implying that every removed patch is irrelevant to all possible future actions.
The committed cache retains the displayed later-usable evidence after retirement. The nested prefix is not rebuilt.
These visualizations provide additional evidence to demonstrate the robustness of our proposed \method{}.
\subsection{Benchmark Breakdowns}
\label{app:supp-results}
\label{app:full}
\begin{figure}[t]
\centering
        \includegraphics[width=0.86\linewidth,height=0.40\textheight,keepaspectratio]{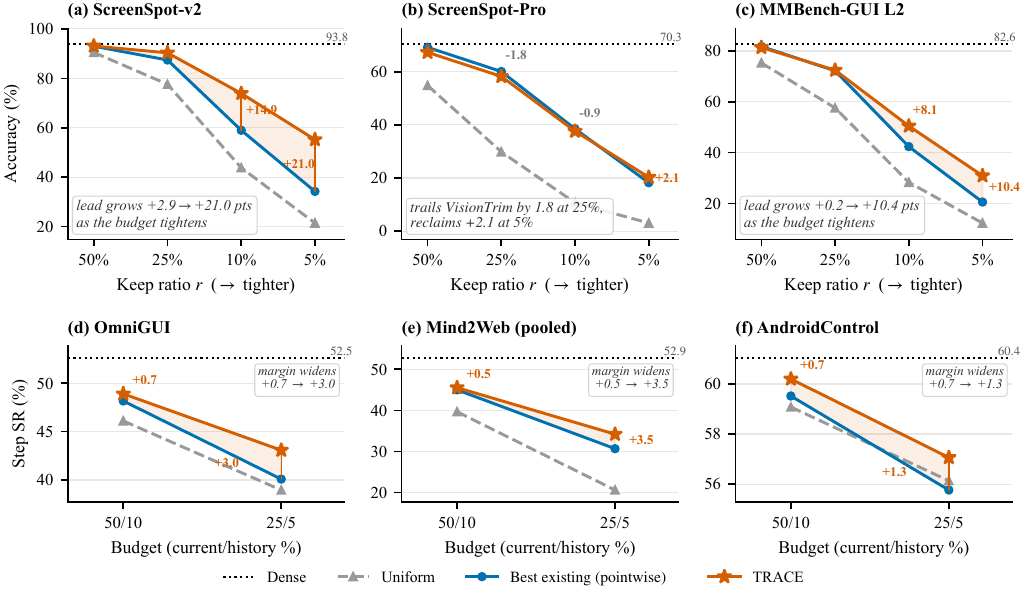}
        \caption{Accuracy versus keep ratio with \textbf{GUI-Owl-1.5-8B}. The top row shows single-step grounding. The bottom row shows multi-step Step~SR. The blue curve uses the best published baseline at each budget, while uniform is shown separately for reference.}
        \label{fig:budget-curves}
\end{figure}

We unpack Tab.~\ref{tab:main-combined} along four axes. The tables separate target categories from action metrics. The budget curves show where margins change. The 2B tables test the same comparisons at another scale.
Throughout this subsection, benchmark averages report accuracy, while the Avg.\,(\%) column reports average retained performance relative to the dense model.
Details are as follows.
\\
\noindent\textbf{Single-step Breakdown.}
\noindent\textit{(1) Tight-budget grounding.}
As shown in Tab.~\ref{tab:main-single}, \method{} reaches $73.90\%$, $37.63\%$, and $50.47\%$ on ScreenSpot-v2, ScreenSpot-Pro, and MMBench-GUI at $\keepr=10\%$.
These scores retain $64.5\%$ of dense performance on average.
Specifically, \method{} leads PruneSID by $14.94\%$ on ScreenSpot-v2 and VisPruner by $8.12\%$ on MMBench-GUI, while VisionTrim remains ahead by $0.89\%$ on ScreenSpot-Pro.
At $\keepr=5\%$, the corresponding scores fall to $55.19\%$, $20.30\%$, and $30.88\%$, but \method{} ranks first on all three benchmark averages.
Its margin over PruneSID on ScreenSpot-v2 grows to $20.99$ points, and the comparison with VisionTrim on ScreenSpot-Pro turns to $+2.08$ points.
Thus, the larger relative advantage at the tight budget does not mean that pruning preserves absolute accuracy. It means that \method{} loses less task performance than the compared selectors at the same budget.
\noindent\textit{(2) Higher-retention budgets.}
Tab.~\ref{tab:supp-single} extends the comparison to $\keepr=50\%$ and $25\%$.
On the 8B model at $\keepr=50\%$, \method{} reaches $93.08\%$ on ScreenSpot-v2 and $81.39\%$ on MMBench-GUI, compared with dense scores of $93.79\%$ and $82.64\%$ in Tab.~\ref{tab:main-single}.
However, its $67.30\%$ on ScreenSpot-Pro trails VisPruner's $69.20\%$.
At $\keepr=25\%$, \method{} reaches $90.33\%$, $58.25\%$, and $72.45\%$ on the three benchmarks, while VisionTrim leads ScreenSpot-Pro at $60.09\%$.
The upper panels of Fig.~\ref{fig:budget-curves} summarize this distinction. The margins widen on ScreenSpot-v2 and MMBench-GUI as the budget tightens, whereas ScreenSpot-Pro shows a crossover rather than a consistent lead.
The blue curve selects the best published baseline separately at each budget, so it need not represent the same method throughout the sweep.
\\
\noindent\textbf{Multi-step Breakdown.}
\noindent\textit{(1) Joint action success.}
Tab.~\ref{tab:main-multi} separates Step~SR from action type, element accuracy, operation F1, and grounding.
Step~SR requires the action type and its target or argument to be correct together, so a strong auxiliary score does not by itself imply a successful step.
At $(\keepc,\keeph)=(50\%,10\%)$, \method{} reaches $48.91\%$, $45.52\%$, and $60.20\%$ Step~SR on OmniGUI, Mind2Web, and AndroidControl.
At $(\keepc,\keeph)=(25\%,5\%)$, these scores become $43.08\%$, $34.20\%$, and $57.06\%$.
The average retained performance decreases from $93.0\%$ to $80.4\%$, while the lead over the best published baseline grows from $0.74$ to $2.99$ points on OmniGUI and from $0.54$ to $3.53$ points on Mind2Web.
The lower panels of Fig.~\ref{fig:budget-curves} show these budget-dependent comparisons alongside the dense and uniform references.
On tight-budget AndroidControl, the distinction between references matters. \method{} leads the best published baseline, ZOO-Prune, by $1.30$ points, but its margin over the stronger uniform control is $0.93$ points.
\begin{table}[t]
    \centering
    \caption{Per-metric breakdown of single-step grounding with \textbf{GUI-Owl-1.5-8B} at $\keepr{=}10\%$ and $5\%$. \textbf{Avg.\,(\%)} denotes retained performance relative to the full-token upper bound.}
    \label{tab:main-single}
    \adjustbox{max width=\textwidth,max totalheight=0.70\textheight,center}{%
        \scriptsize
        \setlength{\tabcolsep}{4.5pt}
        \renewcommand{\arraystretch}{0.96}
        \begin{tabular}{l ccc ccc ccc c}
            \toprule
            & \multicolumn{3}{c}{\textbf{ScreenSpot-v2}} & \multicolumn{3}{c}{\textbf{ScreenSpot-Pro}} & \multicolumn{3}{c}{\textbf{MMBench-GUI L2}} & \\
            \cmidrule(lr){2-4}\cmidrule(lr){5-7}\cmidrule(lr){8-10}
            \textbf{Method} & Text & Icon & \textbf{Avg.} & Text & Icon & \textbf{Avg.} & Basic & Adv. & \textbf{Avg.} & \textbf{Avg.\,(\%)} \\
            \midrule
            \rowcolor{headergray}\multicolumn{11}{c}{\textsc{GUI-Owl-1.5-8B: Upper Bound (100\% Tokens)}} \\
            GUI-Owl-1.5-8B & 97.91 & 88.45 & 93.79 & 81.68 & 51.99 & 70.34 & 90.54 & 74.82 & 82.64 & 100.0\% \\
            \rowcolor{headergray}\multicolumn{11}{c}{\textsc{Retain 10\% Tokens ($\downarrow$\,90\%)}} \\
            Random & 35.38 & 31.77 & 33.81 & 8.19 & 2.81 & 6.14 & 21.88 & 15.22 & 18.53 & 22.4\% \\
            Uniform & 44.29 & 43.14 & 43.79 & 13.10 & 6.79 & 10.69 & 35.70 & 20.92 & 28.27 & 32.0\% \\
            DivPrune\pub{(CVPR25)\cite{alvar2025divprune}} & 51.81 & 52.17 & 51.97 & 17.30 & 8.61 & 13.98 & 37.05 & 26.23 & 31.61 & 37.8\% \\
            CDPruner\pub{(NeurIPS25)\cite{zhang2025cdpruner}} & 38.72 & 25.99 & 33.18 & 14.53 & 5.96 & 11.26 & 17.07 & 7.42 & 12.21 & 22.1\% \\
            VisPruner\pub{(ICCV25)\cite{zhang2025vispruner}} & 46.38 & 58.66 & 51.73 & 39.30 & \underline{26.49} & 34.41 & \underline{50.08} & \underline{34.70} & \underline{42.35} & 51.8\% \\
            PruMerge+\pub{(ICCV25)\cite{shang2025prumerge}} & 36.35 & 45.85 & 40.49 & 21.80 & 16.56 & 19.80 & 37.83 & 23.19 & 30.47 & 36.1\% \\
            TRIM\pub{(COLING25)\cite{song2025trim}} & 29.39 & 45.13 & 36.24 & 4.20 & 1.82 & 3.29 & 32.12 & 18.43 & 25.24 & 24.6\% \\
            FastV\pub{(ECCV24)\cite{chen2024fastv}} & 50.00 & 54.69 & 52.04 & 31.93 & 23.01 & 28.53 & 47.23 & 33.26 & 40.21 & 48.2\% \\
            VisionTrim\pub{(ICLR26)\cite{yu2026visiontrim}} & \underline{58.50} & 58.84 & 58.65 & \underline{44.11} & \textbf{29.47} & \textbf{38.52} & 48.29 & 33.20 & 40.71 & \underline{55.5\%} \\
            ZOO-Prune\pub{(CVPR26)\cite{kim2026zooprune}} & 54.18 & 59.03 & 56.29 & 11.77 & 5.96 & 9.55 & 43.54 & 28.50 & 35.98 & 39.0\% \\
            PruneSID\pub{(ICLR26)\cite{fang2026prunesid}} & 57.66 & \underline{60.65} & \underline{58.96} & 19.24 & 9.93 & 15.69 & 47.12 & 34.26 & 40.65 & 44.8\% \\
            \rowcolor{rowblue}\textbf{\method{}} & \textbf{78.27} & \textbf{68.23} & \textbf{73.90} & \textbf{45.14} & 25.50 & \underline{37.63} & \textbf{60.49} & \textbf{40.56} & \textbf{50.47} & \textbf{64.5\%} \\
            \rowcolor{headergray}\multicolumn{11}{c}{\textsc{Retain 5\% Tokens ($\downarrow$\,95\%)}} \\
            Random & 14.76 & 16.43 & 15.49 & 1.33 & 1.16 & 1.27 & 11.30 & 7.30 & 9.29 & 9.9\% \\
            Uniform & 20.61 & 22.38 & 21.38 & 3.48 & 2.32 & 3.04 & 15.50 & 9.02 & 12.24 & 14.0\% \\
            DivPrune\pub{(CVPR25)\cite{alvar2025divprune}} & 23.40 & 28.70 & 25.71 & 4.30 & 3.15 & 3.86 & 17.63 & 10.63 & 14.11 & 16.7\% \\
            CDPruner\pub{(NeurIPS25)\cite{zhang2025cdpruner}} & 18.38 & 12.64 & 15.88 & 4.50 & 1.82 & 3.48 & 7.39 & 2.16 & 4.76 & 9.2\% \\
            VisPruner\pub{(ICCV25)\cite{zhang2025vispruner}} & 15.04 & 27.98 & 20.68 & 13.20 & 10.60 & 12.21 & 20.09 & 13.45 & 16.75 & 19.9\% \\
            PruMerge+\pub{(ICCV25)\cite{shang2025prumerge}} & 14.62 & 22.02 & 17.85 & 6.96 & 4.14 & 5.88 & 14.77 & 9.63 & 12.19 & 14.0\% \\
            TRIM\pub{(COLING25)\cite{song2025trim}} & 12.12 & 20.76 & 15.88 & 0.41 & 0.50 & 0.44 & 13.43 & 6.31 & 9.85 & 9.8\% \\
            FastV\pub{(ECCV24)\cite{chen2024fastv}} & 26.74 & 30.51 & 28.38 & 12.69 & 10.43 & 11.83 & \underline{25.46} & \underline{15.61} & \underline{20.51} & 24.0\% \\
            VisionTrim\pub{(ICLR26)\cite{yu2026visiontrim}} & 22.14 & 30.69 & 25.86 & \underline{20.98} & \textbf{13.74} & \underline{18.22} & 20.15 & 13.61 & 16.86 & \underline{24.6\%} \\
            ZOO-Prune\pub{(CVPR26)\cite{kim2026zooprune}} & 24.23 & 33.94 & 28.46 & 3.89 & 2.81 & 3.48 & 19.14 & 11.57 & 15.33 & 17.9\% \\
            PruneSID\pub{(ICLR26)\cite{fang2026prunesid}} & \underline{29.67} & \underline{40.07} & \underline{34.20} & 5.02 & 3.81 & 4.55 & 23.89 & 15.11 & 19.48 & 22.2\% \\
            \rowcolor{rowblue}\textbf{\method{}} & \textbf{58.22} & \textbf{51.26} & \textbf{55.19} & \textbf{25.90} & \underline{11.26} & \textbf{20.30} & \textbf{40.51} & \textbf{21.36} & \textbf{30.88} & \textbf{41.7\%} \\
            \bottomrule
        \end{tabular}
    }
\end{table}

    \label{app:m2w}
\noindent\textit{(2) Auxiliary metrics.}
The AndroidControl columns also show why the native metrics should be read separately.
At the tight budget, VisionTrim reaches $67.76\%$ grounding accuracy, above \method{}'s $65.50\%$, yet its Step~SR is lower at $55.16\%$ versus $57.06\%$.
Meanwhile, \method{} reaches $82.06\%$ action-type accuracy, compared with $81.26\%$ for VisionTrim.
These results demonstrate the advantage of our proposed \method{} in different metrics.
\begin{table}[t]
    \centering
    \caption{Per-metric breakdown of multi-step agents with \textbf{GUI-Owl-1.5-8B} under the mild ($\keepc{=}50\%$, $\keeph{=}10\%$) and tight ($\keepc{=}25\%$, $\keeph{=}5\%$) budgets. The dataset-specific columns retain each benchmark's native auxiliary metrics, including action type, element accuracy, operation F1, and grounding.}
    \label{tab:main-multi}
    \adjustbox{max width=\textwidth,max totalheight=0.70\textheight,center}{%
        \scriptsize
        \setlength{\tabcolsep}{4.5pt}
        \renewcommand{\arraystretch}{0.96}
        \begin{tabular}{l cc ccc ccc c}
            \toprule
            & \multicolumn{2}{c}{\textbf{OmniGUI}} & \multicolumn{3}{c}{\textbf{Mind2Web}} & \multicolumn{3}{c}{\textbf{AndroidControl}} & \\
            \cmidrule(lr){2-3}\cmidrule(lr){4-6}\cmidrule(lr){7-9}
            \textbf{Method} & Type & \textbf{Step~SR} & Ele.Acc & Op.F1 & \textbf{Step~SR} & Type & Ground. & \textbf{Step~SR} & \textbf{Avg.\,(\%)} \\
            \midrule
            \rowcolor{headergray}\multicolumn{10}{c}{\textsc{GUI-Owl-1.5-8B: Upper Bound (100\% Tokens)}} \\
            GUI-Owl-1.5-8B & 70.65 & 52.45 & 61.80 & 83.82 & 52.90 & 83.57 & 74.41 & 60.41 & 100.0\% \\
            \rowcolor{headergray}\multicolumn{10}{c}{\textsc{Retain 50\% Current $+$ 10\% History Tokens}} \\
            Random & 66.37 & 43.16 & 42.28 & 84.13 & 35.13 & 82.65 & 70.02 & 58.82 & 82.0\% \\
            Uniform & 66.87 & 46.11 & 47.48 & 84.35 & 39.65 & 82.82 & 69.50 & 59.08 & 86.9\% \\
            DivPrune\pub{(CVPR25)\cite{alvar2025divprune}} & 66.91 & 45.76 & 48.44 & 84.54 & 40.70 & 82.40 & 70.80 & 58.83 & 87.2\% \\
            CDPruner\pub{(NeurIPS25)\cite{zhang2025cdpruner}} & 63.96 & 41.45 & 33.02 & 83.13 & 27.88 & 80.65 & 66.42 & 54.57 & 74.0\% \\
            VisPruner\pub{(ICCV25)\cite{zhang2025vispruner}} & 66.45 & 46.11 & 52.58 & 84.09 & 44.45 & 83.04 & 73.06 & 59.25 & 90.0\% \\
            PruMerge+\pub{(ICCV25)\cite{shang2025prumerge}} & 66.17 & 46.50 & 49.48 & 83.95 & 41.53 & 83.18 & \textbf{73.22} & 59.28 & 88.4\% \\
            TRIM\pub{(COLING25)\cite{song2025trim}} & 65.40 & 45.84 & 48.78 & 84.06 & 41.20 & 82.18 & 70.18 & 56.86 & 86.5\% \\
            FastV\pub{(ECCV24)\cite{chen2024fastv}} & 66.91 & 46.89 & 50.62 & 84.32 & 43.12 & 83.10 & \underline{73.13} & 59.11 & 89.6\% \\
            VisionTrim\pub{(ICLR26)\cite{yu2026visiontrim}} & 65.47 & 45.72 & 51.83 & 83.60 & 43.91 & \textbf{83.50} & 72.45 & 59.10 & 89.3\% \\
            ZOO-Prune\pub{(CVPR26)\cite{kim2026zooprune}} & \underline{67.34} & \underline{48.17} & \underline{52.96} & 84.06 & \underline{44.98} & 83.12 & 72.72 & 59.30 & \underline{91.7\%} \\
            PruneSID\pub{(ICLR26)\cite{fang2026prunesid}} & \underline{67.34} & 47.40 & 51.51 & 84.15 & 43.28 & \underline{83.47} & 72.61 & \underline{59.52} & 90.2\% \\
            \rowcolor{rowblue}\textbf{\method{}} & \textbf{68.62} & \textbf{48.91} & \textbf{53.02} & 84.28 & \textbf{45.52} & 83.44 & 72.74 & \textbf{60.20} & \textbf{93.0\%} \\
            \rowcolor{headergray}\multicolumn{10}{c}{\textsc{Retain 25\% Current $+$ 5\% History Tokens}} \\
            Random & 62.44 & 34.25 & 23.39 & 83.14 & 18.43 & 79.93 & 58.86 & 52.07 & 62.1\% \\
            Uniform & \underline{64.66} & 38.96 & 26.32 & 83.49 & 20.61 & 81.38 & 63.77 & \underline{56.13} & 68.7\% \\
            DivPrune\pub{(CVPR25)\cite{alvar2025divprune}} & 63.72 & 37.17 & 29.95 & 83.72 & 24.29 & 80.92 & 63.53 & 54.75 & 69.1\% \\
            CDPruner\pub{(NeurIPS25)\cite{zhang2025cdpruner}} & 57.50 & 27.76 & 12.55 & 81.61 & 10.86 & 72.42 & 41.98 & 40.90 & 47.1\% \\
            VisPruner\pub{(ICCV25)\cite{zhang2025vispruner}} & 62.99 & 39.54 & \underline{37.20} & 83.83 & \underline{30.67} & 81.55 & \underline{67.36} & 54.93 & \underline{74.8\%} \\
            PruMerge+\pub{(ICCV25)\cite{shang2025prumerge}} & 62.52 & 36.98 & 29.38 & 83.20 & 23.45 & 81.44 & 66.21 & 54.01 & 68.1\% \\
            TRIM\pub{(COLING25)\cite{song2025trim}} & 60.61 & 34.37 & 28.17 & 83.96 & 23.01 & 78.67 & 53.74 & 47.17 & 62.4\% \\
            FastV\pub{(ECCV24)\cite{chen2024fastv}} & 61.47 & 39.39 & 34.48 & 83.26 & 29.33 & \underline{81.58} & 66.81 & 54.61 & 73.6\% \\
            VisionTrim\pub{(ICLR26)\cite{yu2026visiontrim}} & 61.78 & 38.26 & 36.56 & 83.38 & 30.20 & 81.26 & \textbf{67.76} & 55.16 & 73.8\% \\
            ZOO-Prune\pub{(CVPR26)\cite{kim2026zooprune}} & 64.19 & 40.01 & 34.17 & 84.10 & 27.90 & 81.34 & 67.05 & 55.76 & 73.8\% \\
            PruneSID\pub{(ICLR26)\cite{fang2026prunesid}} & 64.27 & \underline{40.09} & 33.88 & 84.01 & 27.57 & 81.44 & 66.45 & 55.46 & 73.5\% \\
            \rowcolor{rowblue}\textbf{\method{}} & \textbf{66.52} & \textbf{43.08} & \textbf{39.41} & 83.90 & \textbf{34.20} & \textbf{82.06} & 65.50 & \textbf{57.06} & \textbf{80.4\%} \\
            \bottomrule
        \end{tabular}
    }
\end{table}

\begin{table}[t]
    \centering
    \caption{Single-step grounding evaluation with \textbf{GUI-Owl-1.5-8B} and \textbf{GUI-Owl-1.5-2B}.}
    \label{tab:supp-single}
    \adjustbox{max width=\textwidth,max totalheight=0.70\textheight,center}{%
        \scriptsize
        \setlength{\tabcolsep}{4.5pt}
        \renewcommand{\arraystretch}{0.96}
        \begin{tabular}{l ccc ccc ccc c}
            \toprule
            & \multicolumn{3}{c}{\textbf{ScreenSpot-v2}} & \multicolumn{3}{c}{\textbf{ScreenSpot-Pro}} & \multicolumn{3}{c}{\textbf{MMBench-GUI L2}} & \\
            \cmidrule(lr){2-4}\cmidrule(lr){5-7}\cmidrule(lr){8-10}
            \textbf{Method} & Text & Icon & \textbf{Avg.} & Text & Icon & \textbf{Avg.} & Basic & Adv. & \textbf{Avg.} & \textbf{Avg.\,(\%)} \\
            \midrule
            \rowcolor{headergray}\multicolumn{11}{c}{\textsc{GUI-Owl-1.5-2B: Upper Bound (100\% Tokens)}} \\
            GUI-Owl-1.5-2B & 93.31 & 86.82 & 90.49 & 67.66 & 42.22 & 57.94 & 83.10 & 60.71 & 71.84 & 100.0\% \\
            \rowcolor{headergray}\multicolumn{11}{c}{\textsc{Retain 50\% Tokens ($\downarrow$\,50\%)}} \\
            Random & 84.68 & 78.88 & 82.15 & 44.93 & 26.49 & 37.89 & 71.91 & 50.08 & 60.93 & 80.3\% \\
            Uniform & 87.19 & 80.32 & 84.20 & 46.57 & 32.62 & 41.24 & 75.66 & 51.96 & 63.75 & 84.3\% \\
            DivPrune\pub{(CVPR25)\cite{alvar2025divprune}} & 87.74 & 82.31 & 85.38 & 51.38 & 27.81 & 42.38 & 75.55 & 54.29 & 64.86 & 85.9\% \\
            CDPruner\pub{(NeurIPS25)\cite{zhang2025cdpruner}} & 87.05 & 83.57 & 85.53 & 45.55 & 24.50 & 37.51 & 77.34 & 52.63 & 64.91 & 83.2\% \\
            VisPruner\pub{(ICCV25)\cite{zhang2025vispruner}} & 86.07 & 83.57 & 84.98 & 61.00 & 37.09 & 51.87 & 78.90 & 55.51 & 67.14 & 92.3\% \\
            PruMerge+\pub{(ICCV25)\cite{shang2025prumerge}} & 84.40 & 79.42 & 82.23 & 54.15 & 35.10 & 46.87 & 76.27 & 50.42 & 63.27 & 86.6\% \\
            TRIM\pub{(COLING25)\cite{song2025trim}} & 87.60 & 81.41 & 84.91 & 44.63 & 21.69 & 35.86 & 73.53 & 49.53 & 61.46 & 80.4\% \\
            FastV\pub{(ECCV24)\cite{chen2024fastv}} & 89.69 & 83.39 & 86.95 & \underline{61.51} & \underline{40.56} & \underline{53.51} & 80.19 & 55.95 & 68.00 & \textbf{94.4\%} \\
            VisionTrim\pub{(ICLR26)\cite{yu2026visiontrim}} & 88.30 & 83.94 & 86.40 & \textbf{62.85} & \textbf{40.73} & \textbf{54.40} & 79.80 & 55.23 & 67.45 & \textbf{94.4\%} \\
            ZOO-Prune\pub{(CVPR26)\cite{kim2026zooprune}} & \underline{90.25} & 83.57 & 87.34 & 52.20 & 27.48 & 42.76 & 78.18 & 56.23 & 67.14 & 87.9\% \\
            PruneSID\pub{(ICLR26)\cite{fang2026prunesid}} & 90.11 & \underline{84.12} & \underline{87.50} & 56.50 & 36.92 & 49.02 & \underline{80.30} & \underline{56.34} & \underline{68.25} & 92.1\% \\
            \rowcolor{rowblue}\textbf{\method{}} & \textbf{90.95} & \textbf{86.10} & \textbf{88.84} & 59.47 & 38.41 & 51.42 & \textbf{80.92} & \textbf{56.45} & \textbf{68.61} & \underline{94.1\%} \\
            \rowcolor{headergray}\multicolumn{11}{c}{\textsc{Retain 25\% Tokens ($\downarrow$\,75\%)}} \\
            Random & 63.23 & 59.39 & 61.56 & 22.42 & 13.74 & 19.10 & 49.69 & 30.55 & 40.07 & 52.3\% \\
            Uniform & 64.62 & 62.82 & 63.84 & 25.38 & 17.55 & 22.39 & 52.77 & 33.43 & 43.04 & 56.4\% \\
            DivPrune\pub{(CVPR25)\cite{alvar2025divprune}} & 69.50 & 71.48 & 70.36 & 27.23 & 16.72 & 23.21 & 57.53 & 39.51 & 48.47 & 61.8\% \\
            CDPruner\pub{(NeurIPS25)\cite{zhang2025cdpruner}} & 65.88 & 71.48 & 68.32 & 28.25 & 16.56 & 23.78 & 61.89 & 37.41 & 49.58 & 61.9\% \\
            VisPruner\pub{(ICCV25)\cite{zhang2025vispruner}} & 57.10 & 68.41 & 62.03 & 42.58 & 23.84 & 35.42 & 56.13 & 33.87 & 44.94 & 64.1\% \\
            PruMerge+\pub{(ICCV25)\cite{shang2025prumerge}} & 59.75 & 61.01 & 60.30 & 27.64 & 18.87 & 24.29 & 50.76 & 29.44 & 40.04 & 54.8\% \\
            TRIM\pub{(COLING25)\cite{song2025trim}} & 62.26 & 66.61 & 64.15 & 18.32 & 9.93 & 15.12 & 51.76 & 30.66 & 41.15 & 51.4\% \\
            FastV\pub{(ECCV24)\cite{chen2024fastv}} & \underline{77.58} & 65.52 & 72.33 & \underline{46.57} & \textbf{31.62} & \underline{40.86} & \underline{64.19} & \underline{42.39} & \underline{53.23} & \underline{74.8\%} \\
            VisionTrim\pub{(ICLR26)\cite{yu2026visiontrim}} & 63.65 & 70.76 & 66.75 & \textbf{49.23} & \underline{29.80} & \textbf{41.81} & 62.79 & 38.35 & 50.50 & 72.1\% \\
            ZOO-Prune\pub{(CVPR26)\cite{kim2026zooprune}} & 73.12 & 72.92 & 73.03 & 24.26 & 11.92 & 19.54 & 57.19 & 39.18 & 48.14 & 60.5\% \\
            PruneSID\pub{(ICLR26)\cite{fang2026prunesid}} & 74.51 & \underline{75.09} & \underline{74.76} & 31.12 & 22.85 & 27.96 & 63.23 & 41.56 & 52.34 & 67.9\% \\
            \rowcolor{rowblue}\textbf{\method{}} & \textbf{80.36} & \textbf{80.87} & \textbf{80.58} & 45.55 & 28.64 & 39.09 & \textbf{72.92} & \textbf{46.82} & \textbf{59.79} & \textbf{79.9\%} \\
            \midrule
            \rowcolor{headergray}\multicolumn{11}{c}{\textsc{GUI-Owl-1.5-8B: Upper Bound (100\% Tokens)}} \\
            GUI-Owl-1.5-8B & 97.91 & 88.45 & 93.79 & 81.68 & 51.99 & 70.34 & 90.54 & 74.82 & 82.64 & 100.0\% \\
            \rowcolor{headergray}\multicolumn{11}{c}{\textsc{Retain 50\% Tokens ($\downarrow$\,50\%)}} \\
            Random & 91.36 & 80.51 & 86.64 & 57.52 & 33.11 & 48.20 & 78.96 & 61.15 & 70.01 & 81.9\% \\
            Uniform & 96.10 & 83.21 & 90.49 & 67.14 & 34.93 & 54.84 & 84.00 & 66.74 & 75.32 & 88.5\% \\
            DivPrune\pub{(CVPR25)\cite{alvar2025divprune}} & 93.87 & 84.84 & 89.94 & 74.51 & 41.89 & 62.05 & 84.78 & 68.29 & 76.49 & 92.2\% \\
            CDPruner\pub{(NeurIPS25)\cite{zhang2025cdpruner}} & 91.78 & 80.32 & 86.79 & 70.42 & 40.89 & 59.14 & 75.27 & 57.78 & 66.47 & 85.7\% \\
            VisPruner\pub{(ICCV25)\cite{zhang2025vispruner}} & 95.82 & 86.28 & 91.67 & \textbf{80.66} & \underline{50.66} & \textbf{69.20} & \textbf{90.15} & 73.49 & \textbf{81.78} & \textbf{98.4\%} \\
            PruMerge+\pub{(ICCV25)\cite{shang2025prumerge}} & 96.10 & 86.28 & 91.82 & 74.21 & 47.35 & 63.95 & 87.63 & 70.95 & 79.24 & 94.9\% \\
            TRIM\pub{(COLING25)\cite{song2025trim}} & 93.73 & 86.46 & 90.57 & 76.46 & 43.21 & 63.76 & 88.75 & 69.89 & 79.27 & 94.4\% \\
            FastV\pub{(ECCV24)\cite{chen2024fastv}} & 95.68 & 87.73 & 92.22 & 77.99 & \textbf{51.82} & \underline{67.99} & 89.03 & \underline{73.88} & \underline{81.41} & 97.8\% \\
            VisionTrim\pub{(ICLR26)\cite{yu2026visiontrim}} & \textbf{97.08} & 87.73 & \underline{93.00} & 77.79 & \underline{50.66} & 67.43 & 89.54 & \textbf{74.10} & \textbf{81.78} & \underline{98.0\%} \\
            ZOO-Prune\pub{(CVPR26)\cite{kim2026zooprune}} & \underline{96.80} & \underline{88.09} & \underline{93.00} & \underline{79.12} & 46.85 & 66.79 & 88.81 & 73.10 & 80.91 & 97.3\% \\
            PruneSID\pub{(ICLR26)\cite{fang2026prunesid}} & 96.52 & 85.74 & 91.82 & 77.69 & 49.67 & 66.98 & 89.03 & 73.05 & 81.00 & 97.0\% \\
            \rowcolor{rowblue}\textbf{\method{}} & 96.66 & \textbf{88.45} & \textbf{93.08} & 78.61 & 49.01 & 67.30 & \underline{89.70} & 73.16 & 81.39 & 97.8\% \\
            \rowcolor{headergray}\multicolumn{11}{c}{\textsc{Retain 25\% Tokens ($\downarrow$\,75\%)}} \\
            Random & 72.42 & 63.90 & 68.71 & 28.66 & 14.07 & 23.09 & 51.82 & 37.96 & 44.85 & 53.5\% \\
            Uniform & 82.87 & 70.76 & 77.59 & 36.13 & 19.70 & 29.85 & 66.59 & 48.64 & 57.57 & 64.9\% \\
            DivPrune\pub{(CVPR25)\cite{alvar2025divprune}} & 84.96 & 75.99 & 81.05 & 55.58 & 29.30 & 45.54 & 70.73 & 55.40 & 63.02 & 75.8\% \\
            CDPruner\pub{(NeurIPS25)\cite{zhang2025cdpruner}} & 75.77 & 58.66 & 68.32 & 49.85 & 21.36 & 38.96 & 46.00 & 29.00 & 37.45 & 57.8\% \\
            VisPruner\pub{(ICCV25)\cite{zhang2025vispruner}} & 88.02 & \underline{83.03} & 85.85 & \underline{67.76} & \underline{45.20} & \underline{59.14} & \underline{81.48} & \textbf{63.03} & \underline{72.20} & 87.7\% \\
            PruMerge+\pub{(ICCV25)\cite{shang2025prumerge}} & 82.03 & 77.98 & 80.27 & 52.81 & 37.25 & 46.87 & 75.55 & 54.29 & 64.86 & 76.9\% \\
            TRIM\pub{(COLING25)\cite{song2025trim}} & 71.17 & 74.01 & 72.41 & 35.72 & 19.70 & 29.60 & 67.99 & 44.00 & 55.93 & 62.3\% \\
            FastV\pub{(ECCV24)\cite{chen2024fastv}} & 86.35 & 80.87 & 83.96 & 61.31 & 41.72 & 53.83 & 77.34 & 59.71 & 68.48 & 83.0\% \\
            VisionTrim\pub{(ICLR26)\cite{yu2026visiontrim}} & 91.23 & 82.49 & \underline{87.42} & \underline{67.76} & \textbf{47.68} & \textbf{60.09} & 81.03 & 61.54 & 71.23 & \underline{88.3\%} \\
            ZOO-Prune\pub{(CVPR26)\cite{kim2026zooprune}} & 89.28 & 80.87 & 85.61 & 55.58 & 26.16 & 44.34 & 78.46 & 60.10 & 69.23 & 79.4\% \\
            PruneSID\pub{(ICLR26)\cite{fang2026prunesid}} & \underline{91.64} & 80.69 & 86.87 & 54.76 & 26.82 & 44.09 & 78.18 & 61.04 & 69.56 & 79.8\% \\
            \rowcolor{rowblue}\textbf{\method{}} & \textbf{94.15} & \textbf{85.38} & \textbf{90.33} & \textbf{68.88} & 41.06 & 58.25 & \textbf{82.65} & \underline{62.37} & \textbf{72.45} & \textbf{88.9\%} \\
            \bottomrule
        \end{tabular}
    }
\end{table}

\\
\noindent\textbf{Mind2Web Metric Breakdown.}
The pooled Mind2Web columns in Tab.~\ref{tab:main-multi} distinguish target selection from operation prediction.
At the mild budget, \method{} reaches $53.02\%$ element accuracy and $45.52\%$ Step~SR, compared with $47.48\%$ and $39.65\%$ for uniform.
At the tight budget, the corresponding scores are $39.41\%$ and $34.20\%$ for \method{}, against $26.32\%$ and $20.61\%$ for uniform.
In contrast, operation F1 stays close between the two methods, at $84.28\%$ versus $84.35\%$ under the mild budget and $83.90\%$ versus $83.49\%$ under the tight budget.
The larger differences therefore occur in element accuracy and joint step success, rather than operation F1.
Detailed breakdowns can help us understand the robustness of our proposed \method{} under specific scenarios.
\\
\noindent\textbf{Cross-scale Evaluation.}
We next compare the same benchmark metrics on GUI-Owl-1.5-2B.
\noindent\textit{(1) Single-step transfer.}
As shown in Tab.~\ref{tab:supp-2b-single}, \method{} reaches $55.11\%$, $17.33\%$, and $38.17\%$ on ScreenSpot-v2, ScreenSpot-Pro, and MMBench-GUI at $\keepr=10\%$.
At $\keepr=5\%$, it reaches $38.29\%$, $6.83\%$, and $21.93\%$.
It ranks first on all three benchmark averages at both budgets, retaining $48.0\%$ and $28.2\%$ of dense performance on average.
However, the category columns show a narrower advantage than the benchmark ranks alone suggest. On ScreenSpot-Pro at $\keepr=5\%$, \method{} leads FastV on text targets at $6.86\%$ versus $5.53\%$, but trails on icon targets at $6.79\%$ versus $8.28\%$.
Its overall lead is only $0.25$ points, from $6.83\%$ versus $6.58\%$.
Thus, ranking first on the smaller model does not imply either high absolute accuracy or a lead on every target category.
\noindent\textit{(2) Multi-step transfer.}
Tab.~\ref{tab:supp-2b-multi} reports $36.08\%$, $37.92\%$, and $56.53\%$ Step~SR for \method{} on OmniGUI, Mind2Web, and AndroidControl at the mild budget.
The tight-budget scores are $31.88\%$, $26.52\%$, and $53.72\%$.
These retain $91.4\%$ and $77.7\%$ of dense performance on average across the three benchmarks.
\method{} leads OmniGUI at both budgets, but FastV remains ahead on Mind2Web at $38.98\%$ and $27.62\%$, while PruneSID remains ahead on AndroidControl at $57.03\%$ and $54.13\%$.
The cross-scale result therefore preserves competitiveness without preserving the 8B lead on every multi-step benchmark.
\begin{figure}[t]
\centering
\includegraphics[width=0.92\linewidth,height=0.40\textheight,keepaspectratio]{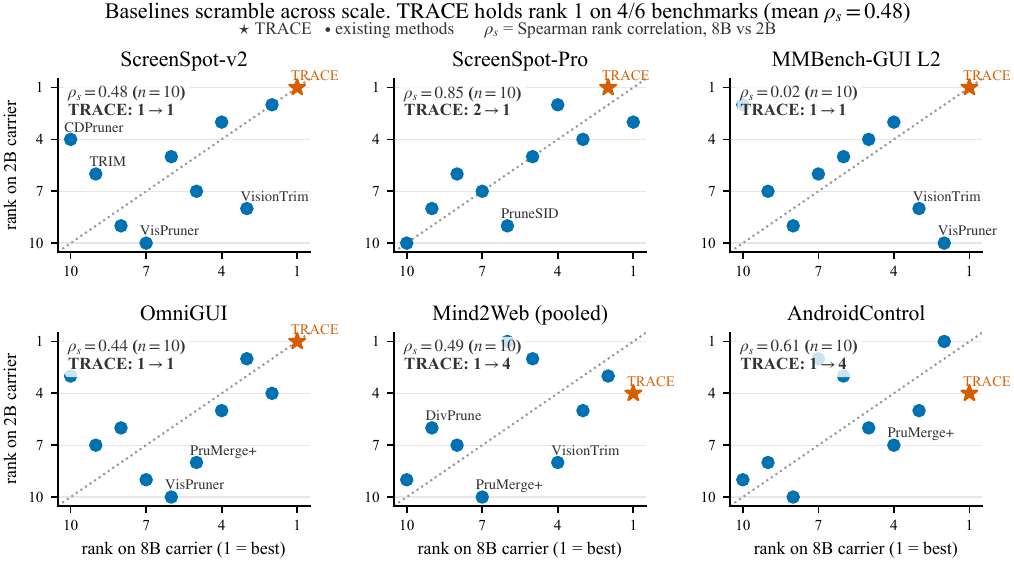}
\caption{Cross-scale rank comparison between \textbf{GUI-Owl-1.5-8B} and \textbf{GUI-Owl-1.5-2B}. Each panel shows the ten-method comparison plotted for one benchmark. Stars mark \method{}, dots mark existing methods, and the dotted diagonal denotes equal ranks.}
\label{fig:rank-stability}
\end{figure}
\noindent\textit{(3) Rank stability.}
Fig.~\ref{fig:rank-stability} separates the rank of \method{} from changes in the overall method ordering.
In the plotted comparisons, \method{} remains first on ScreenSpot-v2, MMBench-GUI, and OmniGUI, and moves from second to first on ScreenSpot-Pro.
It moves from first to fourth on pooled Mind2Web and AndroidControl.
Tab.~\ref{tab:scale-rep} provides the complementary budget-level view with twelve methods, rather than the ten plotted in the figure, and lists the three Mind2Web splits separately.
In that table, \method{} remains first on ScreenSpot-v2 and MMBench-GUI at all four keep ratios and on OmniGUI at both current/history budgets.
However, its 2B rank is fourth on each mild-budget Mind2Web split and ranges from second to third at the tight budget.
The full ordering can change even when the top-ranked method does not. For example, the table reports Spearman $\rho=0.109$ and $0.091$ on MMBench-GUI at $\keepr=10\%$ and $5\%$, while \method{} remains first.
These results distinguish stability of the leading method from agreement among all selectors.
\begin{table}[t]
    \centering
    \caption{Single-step grounding with \textbf{GUI-Owl-1.5-2B} at $\keepr{=}10\%$ and $5\%$.}
    \label{tab:supp-2b-single}
    \adjustbox{max width=\textwidth,max totalheight=0.70\textheight,center}{%
        \scriptsize
        \setlength{\tabcolsep}{4.5pt}
        \renewcommand{\arraystretch}{0.96}
        \begin{tabular}{l ccc ccc ccc c}
            \toprule
            & \multicolumn{3}{c}{\textbf{ScreenSpot-v2}} & \multicolumn{3}{c}{\textbf{ScreenSpot-Pro}} & \multicolumn{3}{c}{\textbf{MMBench-GUI L2}} & \\
            \cmidrule(lr){2-4}\cmidrule(lr){5-7}\cmidrule(lr){8-10}
            \textbf{Method} & Text & Icon & \textbf{Avg.} & Text & Icon & \textbf{Avg.} & Basic & Adv. & \textbf{Avg.} & \textbf{Avg.\,(\%)} \\
            \midrule
            \rowcolor{headergray}\multicolumn{11}{c}{\textsc{GUI-Owl-1.5-2B: Upper Bound (100\% Tokens)}} \\
            GUI-Owl-1.5-2B & 93.31 & 86.82 & 90.49 & 67.66 & 42.22 & 57.94 & 83.10 & 60.71 & 71.84 & 100.0\% \\
            \rowcolor{headergray}\multicolumn{11}{c}{\textsc{Retain 10\% Tokens ($\downarrow$\,90\%)}} \\
            Random & 27.30 & 29.78 & 28.38 & 7.78 & 5.13 & 6.77 & 22.55 & 13.06 & 17.78 & 22.6\% \\
            Uniform & 25.35 & 33.03 & 28.69 & 9.72 & 4.64 & 7.78 & 23.67 & 11.57 & 17.58 & 23.2\% \\
            DivPrune\pub{(CVPR25)\cite{alvar2025divprune}} & 32.31 & 43.86 & 37.34 & 7.06 & 5.30 & 6.39 & 23.39 & 16.16 & 19.76 & 26.6\% \\
            CDPruner\pub{(NeurIPS25)\cite{zhang2025cdpruner}} & 32.03 & 44.95 & 37.66 & 7.57 & 4.64 & 6.45 & \underline{33.86} & \underline{18.59} & \underline{26.18} & 29.7\% \\
            VisPruner\pub{(ICCV25)\cite{zhang2025vispruner}} & 18.66 & 30.69 & 23.90 & 8.80 & 6.79 & 8.03 & 18.47 & 10.24 & 14.33 & 20.1\% \\
            PruMerge+\pub{(ICCV25)\cite{shang2025prumerge}} & 24.51 & 28.52 & 26.26 & 7.37 & 5.63 & 6.70 & 19.75 & 12.17 & 15.94 & 20.9\% \\
            TRIM\pub{(COLING25)\cite{song2025trim}} & 26.46 & 40.25 & 32.47 & 3.79 & 4.30 & 3.98 & 25.35 & 14.22 & 19.76 & 23.4\% \\
            FastV\pub{(ECCV24)\cite{chen2024fastv}} & 27.16 & 33.39 & 29.87 & 15.46 & \textbf{16.06} & \underline{15.69} & 26.69 & 16.10 & 21.37 & \underline{29.9\%} \\
            VisionTrim\pub{(ICLR26)\cite{yu2026visiontrim}} & 24.65 & 36.46 & 29.80 & \underline{16.27} & 12.42 & 14.80 & 25.80 & 12.12 & 18.92 & 28.3\% \\
            ZOO-Prune\pub{(CVPR26)\cite{kim2026zooprune}} & 34.82 & \underline{45.13} & 39.31 & 6.04 & 5.30 & 5.76 & 25.52 & 16.60 & 21.04 & 27.6\% \\
            PruneSID\pub{(ICLR26)\cite{fang2026prunesid}} & \underline{40.53} & 44.58 & \underline{42.30} & 6.04 & 3.64 & 5.12 & 26.52 & 16.60 & 21.54 & 28.5\% \\
            \rowcolor{rowblue}\textbf{\method{}} & \textbf{50.97} & \textbf{60.47} & \textbf{55.11} & \textbf{19.34} & \underline{14.07} & \textbf{17.33} & \textbf{48.18} & \textbf{28.28} & \textbf{38.17} & \textbf{48.0\%} \\
            \rowcolor{headergray}\multicolumn{11}{c}{\textsc{Retain 5\% Tokens ($\downarrow$\,95\%)}} \\
            Random & 12.26 & 17.51 & 14.54 & 1.94 & 1.32 & 1.71 & 10.63 & 6.09 & 8.35 & 10.2\% \\
            Uniform & 12.40 & 14.98 & 13.52 & 2.87 & 2.15 & 2.59 & 10.41 & 5.76 & 8.07 & 10.2\% \\
            DivPrune\pub{(CVPR25)\cite{alvar2025divprune}} & 15.32 & 21.48 & 18.00 & 2.76 & 1.49 & 2.28 & 10.18 & 6.97 & 8.57 & 11.9\% \\
            CDPruner\pub{(NeurIPS25)\cite{zhang2025cdpruner}} & \underline{17.13} & \underline{28.70} & \underline{22.17} & 2.87 & 2.32 & 2.66 & \underline{16.79} & \underline{9.35} & \underline{13.05} & \underline{15.8\%} \\
            VisPruner\pub{(ICCV25)\cite{zhang2025vispruner}} & 10.17 & 16.25 & 12.81 & 2.56 & 1.99 & 2.34 & 8.23 & 5.15 & 6.68 & 9.2\% \\
            PruMerge+\pub{(ICCV25)\cite{shang2025prumerge}} & 9.33 & 15.52 & 12.03 & 2.66 & 1.82 & 2.34 & 9.01 & 4.98 & 6.98 & 9.0\% \\
            TRIM\pub{(COLING25)\cite{song2025trim}} & 12.95 & 23.10 & 17.37 & 0.92 & 2.65 & 1.58 & 12.37 & 6.92 & 9.63 & 11.8\% \\
            FastV\pub{(ECCV24)\cite{chen2024fastv}} & 6.96 & 14.98 & 10.46 & \underline{5.53} & \textbf{8.28} & \underline{6.58} & 10.07 & 5.09 & 7.57 & 11.2\% \\
            VisionTrim\pub{(ICLR26)\cite{yu2026visiontrim}} & 14.76 & 20.58 & 17.30 & \underline{5.53} & 2.65 & 4.43 & 13.21 & 5.70 & 9.43 & 13.3\% \\
            ZOO-Prune\pub{(CVPR26)\cite{kim2026zooprune}} & 15.04 & 25.27 & 19.50 & 1.43 & 1.82 & 1.58 & 11.53 & 9.19 & 10.35 & 12.9\% \\
            PruneSID\pub{(ICLR26)\cite{fang2026prunesid}} & 15.32 & 25.99 & 19.97 & 1.64 & 2.15 & 1.83 & 12.98 & 8.36 & 10.66 & 13.4\% \\
            \rowcolor{rowblue}\textbf{\method{}} & \textbf{33.43} & \textbf{44.58} & \textbf{38.29} & \textbf{6.86} & \underline{6.79} & \textbf{6.83} & \textbf{28.93} & \textbf{15.00} & \textbf{21.93} & \textbf{28.2\%} \\
            \bottomrule
        \end{tabular}
    }
\end{table}

\begin{table}[t]
    \centering
    \caption{Multi-step agents with \textbf{GUI-Owl-1.5-2B}. The table reports native auxiliary metrics and Step SR for each multi-step benchmark under the matched current and history budgets.}
    \label{tab:supp-2b-multi}
    \adjustbox{max width=\textwidth,max totalheight=0.70\textheight,center}{%
        \scriptsize
        \setlength{\tabcolsep}{4.5pt}
        \renewcommand{\arraystretch}{0.96}
        \begin{tabular}{l cc ccc ccc c}
            \toprule
            & \multicolumn{2}{c}{\textbf{OmniGUI}} & \multicolumn{3}{c}{\textbf{Mind2Web}} & \multicolumn{3}{c}{\textbf{AndroidControl}} & \\
            \cmidrule(lr){2-3}\cmidrule(lr){4-6}\cmidrule(lr){7-9}
            \textbf{Method} & Type & \textbf{Step~SR} & Ele.Acc & Op.F1 & \textbf{Step~SR} & Type & Ground. & \textbf{Step~SR} & \textbf{Avg.\,(\%)} \\
            \midrule
            \rowcolor{headergray}\multicolumn{10}{c}{\textsc{GUI-Owl-1.5-2B: Upper Bound (100\% Tokens)}} \\
            GUI-Owl-1.5-2B & 57.97 & 39.70 & 52.74 & 84.34 & 44.23 & 83.62 & 72.61 & 57.93 & 100.0\% \\
            \rowcolor{headergray}\multicolumn{10}{c}{\textsc{Retain 50\% Current $+$ 10\% History Tokens}} \\
            Random & 54.74 & 32.31 & 36.68 & 84.03 & 29.60 & 81.94 & 68.13 & 54.88 & 81.0\% \\
            Uniform & \textbf{56.49} & 33.51 & 39.49 & 84.17 & 32.44 & 82.43 & 69.13 & 55.54 & 84.5\% \\
            DivPrune\pub{(CVPR25)\cite{alvar2025divprune}} & 56.26 & 33.71 & 43.97 & 84.11 & 35.99 & 79.48 & 69.57 & 54.62 & 86.9\% \\
            CDPruner\pub{(NeurIPS25)\cite{zhang2025cdpruner}} & 53.58 & 33.83 & 40.13 & 84.19 & 32.51 & 81.77 & 69.26 & 54.88 & 84.5\% \\
            VisPruner\pub{(ICCV25)\cite{zhang2025vispruner}} & 52.64 & 31.34 & 44.93 & 84.17 & 36.90 & 82.85 & 69.22 & 55.93 & 86.3\% \\
            PruMerge+\pub{(ICCV25)\cite{shang2025prumerge}} & 53.62 & 32.81 & 38.55 & 84.18 & 31.18 & \textbf{83.62} & 68.37 & 55.91 & 83.2\% \\
            TRIM\pub{(COLING25)\cite{song2025trim}} & 52.14 & 31.57 & 43.14 & 83.95 & 35.83 & 82.50 & 68.63 & 55.73 & 85.6\% \\
            FastV\pub{(ECCV24)\cite{chen2024fastv}} & 53.50 & 33.79 & \underline{46.45} & 84.11 & \textbf{38.98} & \underline{83.53} & 70.59 & 56.71 & \underline{90.4\%} \\
            VisionTrim\pub{(ICLR26)\cite{yu2026visiontrim}} & 53.54 & 33.13 & 42.30 & 84.25 & 34.83 & 83.28 & 70.16 & \underline{56.77} & 86.7\% \\
            ZOO-Prune\pub{(CVPR26)\cite{kim2026zooprune}} & 55.05 & 33.83 & 46.41 & 84.04 & 38.47 & 82.34 & \textbf{71.11} & 56.42 & 89.9\% \\
            PruneSID\pub{(ICLR26)\cite{fang2026prunesid}} & 54.20 & \underline{34.95} & \textbf{46.80} & 83.97 & \underline{38.81} & 83.22 & \underline{71.01} & \textbf{57.03} & \textbf{91.4\%} \\
            \rowcolor{rowblue}\textbf{\method{}} & \underline{56.42} & \textbf{36.08} & 45.48 & 84.21 & 37.92 & 82.44 & 70.90 & 56.53 & \textbf{91.4\%} \\
            \rowcolor{headergray}\multicolumn{10}{c}{\textsc{Retain 25\% Current $+$ 5\% History Tokens}} \\
            Random & 51.79 & 24.77 & 20.74 & 84.00 & 16.21 & 79.52 & 56.50 & 47.43 & 60.3\% \\
            Uniform & 53.54 & 25.66 & 20.34 & 83.58 & 15.92 & 81.61 & 63.28 & 52.38 & 63.7\% \\
            DivPrune\pub{(CVPR25)\cite{alvar2025divprune}} & \underline{54.16} & 26.52 & 26.66 & 84.22 & 20.52 & 78.24 & 62.31 & 50.30 & 66.7\% \\
            CDPruner\pub{(NeurIPS25)\cite{zhang2025cdpruner}} & 51.48 & 27.10 & 24.31 & 84.01 & 18.20 & 79.88 & 62.81 & 50.19 & 65.3\% \\
            VisPruner\pub{(ICCV25)\cite{zhang2025vispruner}} & 50.16 & 22.55 & 25.04 & 83.99 & 18.85 & 81.36 & 56.68 & 48.59 & 61.1\% \\
            PruMerge+\pub{(ICCV25)\cite{shang2025prumerge}} & 50.82 & 23.17 & 19.12 & 83.77 & 14.68 & 81.28 & 54.41 & 47.35 & 57.8\% \\
            TRIM\pub{(COLING25)\cite{song2025trim}} & 50.12 & 25.35 & 28.78 & 83.83 & 23.58 & 80.51 & 60.86 & 51.28 & 68.6\% \\
            FastV\pub{(ECCV24)\cite{chen2024fastv}} & 49.88 & 25.51 & \textbf{32.63} & 84.12 & \textbf{27.62} & 81.94 & 63.54 & 51.71 & 72.0\% \\
            VisionTrim\pub{(ICLR26)\cite{yu2026visiontrim}} & 49.11 & 23.72 & 23.90 & 83.56 & 17.80 & \underline{82.13} & 58.29 & 49.43 & 61.8\% \\
            ZOO-Prune\pub{(CVPR26)\cite{kim2026zooprune}} & 52.60 & 26.17 & 29.25 & 84.02 & 22.78 & 80.53 & \underline{65.90} & 53.06 & 69.7\% \\
            PruneSID\pub{(ICLR26)\cite{fang2026prunesid}} & 51.59 & \underline{27.53} & 32.00 & 84.09 & 25.68 & \textbf{82.50} & 65.64 & \textbf{54.13} & \underline{73.6\%} \\
            \rowcolor{rowblue}\textbf{\method{}} & \textbf{54.98} & \textbf{31.88} & \underline{32.44} & 84.18 & \underline{26.52} & 80.66 & \textbf{66.53} & \underline{53.72} & \textbf{77.7\%} \\
            \bottomrule
        \end{tabular}
    }
\end{table}

\begin{table}[t]
    \centering
    \caption{Cross-scale replication with \textbf{GUI-Owl-1.5-2B}. Spearman $\rho$ between the 8B and 2B method orderings on each grid, with the rank of \method{} at both scales.}
    \label{tab:scale-rep}
    \small
    \begin{tabular}{ll ccc c}
        \toprule
        \textbf{Benchmark} & \textbf{Budget} & \textbf{\#Methods} & \textbf{Spearman $\rho$} & \textbf{Rank@8B} & \textbf{Rank@2B} \\
        \midrule
        ScreenSpot-v2 & $r{=}50\%$ & 12 & 0.698 & 1 & 1 \\
        ScreenSpot-v2 & $r{=}25\%$ & 12 & 0.545 & 1 & 1 \\
        ScreenSpot-v2 & $r{=}10\%$ & 12 & 0.503 & 1 & 1 \\
        ScreenSpot-v2 & $r{=}5\%$ & 12 & 0.329 & 1 & 1 \\
        ScreenSpot-Pro & $r{=}50\%$ & 12 & 0.874 & 4 & 4 \\
        ScreenSpot-Pro & $r{=}25\%$ & 12 & 0.867 & 3 & 3 \\
        ScreenSpot-Pro & $r{=}10\%$ & 12 & 0.706 & 2 & 1 \\
        ScreenSpot-Pro & $r{=}5\%$ & 12 & 0.701 & 1 & 1 \\
        MMBench-GUI L2 & $r{=}50\%$ & 12 & 0.689 & 4 & 1 \\
        MMBench-GUI L2 & $r{=}25\%$ & 12 & 0.524 & 1 & 1 \\
        MMBench-GUI L2 & $r{=}10\%$ & 12 & 0.109 & 1 & 1 \\
        MMBench-GUI L2 & $r{=}5\%$ & 12 & 0.091 & 1 & 1 \\
        OmniGUI & $c{=}50\%$, $h{=}10\%$ & 12 & 0.488 & 1 & 1 \\
        OmniGUI & $c{=}25\%$, $h{=}5\%$ & 12 & 0.336 & 1 & 1 \\
        Mind2Web Cross-Task & $c{=}50\%$, $h{=}10\%$ & 12 & 0.559 & 1 & 4 \\
        Mind2Web Cross-Task & $c{=}25\%$, $h{=}5\%$ & 12 & 0.399 & 1 & 2 \\
        Mind2Web Cross-Domain & $c{=}50\%$, $h{=}10\%$ & 12 & 0.664 & 2 & 4 \\
        Mind2Web Cross-Domain & $c{=}25\%$, $h{=}5\%$ & 12 & 0.594 & 1 & 2 \\
        Mind2Web Cross-Website & $c{=}50\%$, $h{=}10\%$ & 12 & 0.694 & 1 & 4 \\
        Mind2Web Cross-Website & $c{=}25\%$, $h{=}5\%$ & 12 & 0.308 & 1 & 3 \\
        AndroidControl & $c{=}50\%$, $h{=}10\%$ & 12 & 0.716 & 1 & 4 \\
        AndroidControl & $c{=}25\%$, $h{=}5\%$ & 12 & 0.445 & 4 & 2 \\
        \bottomrule
    \end{tabular}
\end{table}

\\
\noindent\textbf{Target-category Breakdown.}
Finally, the category columns in Tabs.~\ref{tab:supp-single} and~\ref{tab:main-single} show which targets contribute to the benchmark averages.
On 8B ScreenSpot-Pro at $\keepr=25\%$, \method{} leads VisionTrim on text targets at $68.88\%$ versus $67.76\%$, but trails on icon targets at $41.06\%$ versus $47.68\%$.
This text--icon contrast accompanies the lower overall score of $58.25\%$ versus $60.09\%$.
At $\keepr=5\%$, \method{} still trails VisionTrim on icons at $11.26\%$ versus $13.74\%$, but its text accuracy of $25.90\%$ versus $20.98\%$ is accompanied by a higher benchmark average.
The crossover in Fig.~\ref{fig:budget-curves}(b) therefore does not indicate that \method{} overtakes VisionTrim on both target types.
On MMBench-GUI at the same tight budget, the advantage is present in both reported categories. \method{} reaches $40.51\%$ on Basic and $21.36\%$ on Adv., compared with FastV's $25.46\%$ and $15.61\%$.
Together, these breakdowns locate the gains in the reported categories and metrics for comprehensive understanding.
\subsection{Ablations and Visual Evidence}
\label{app:supp-ablation}

\begin{figure}[t]
\centering
\includegraphics[width=0.99\linewidth,height=0.40\textheight,keepaspectratio,trim=0 367 0 0,clip]{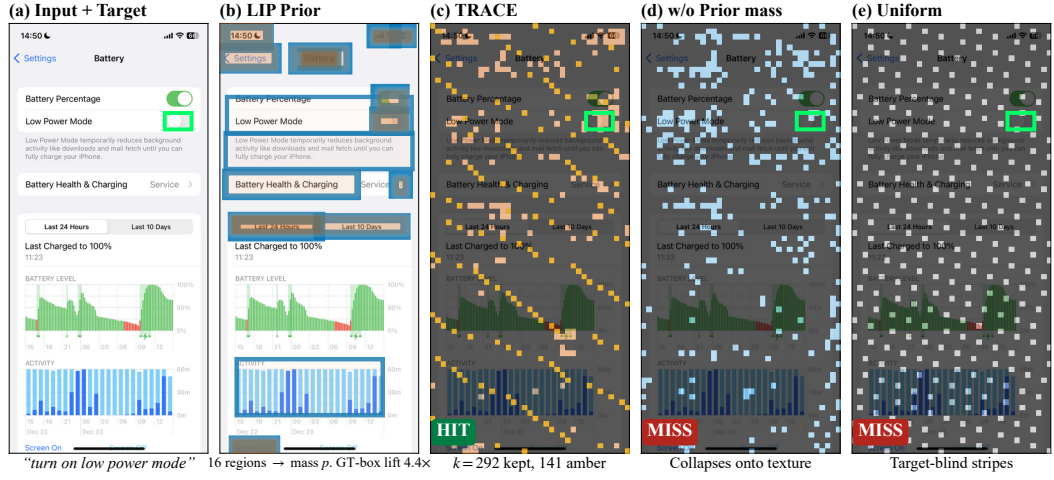}
\caption{Selection outcomes with \textbf{GUI-Owl-1.5-8B} at $\keepr=10\%$. Panels (a) through (e) show the input, the \lmp{} field, and the keeps of \method{}, no-prior, and uniform.}
\label{fig:keepmaps}
\end{figure}

\begin{figure}[t]
\centering
\includegraphics[width=0.80\linewidth]{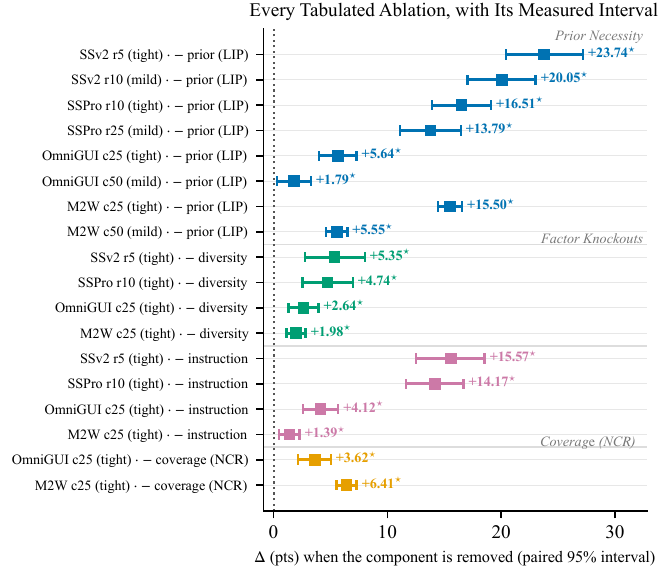}
\caption{Ablation forest of different components in \method{} with \textbf{GUI-Owl-1.5-8B}.}
\label{fig:ablation-forest}
\end{figure}

\begin{figure}[t]
\centering
\includegraphics[width=0.92\linewidth]{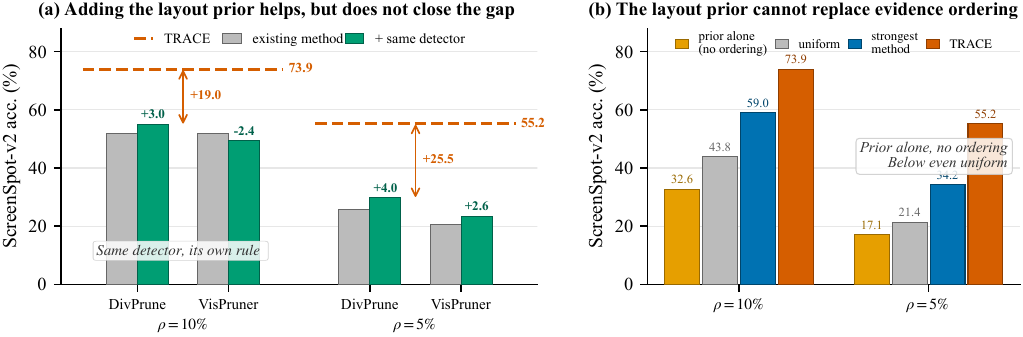}
\caption{Layout-prior transfer and ordering ablation with
\textbf{GUI-Owl-1.5-8B} on ScreenSpot-v2.
(a) Layout-prior transfer analysis under
matched budgets. Dashed lines mark
\method{}, and arrows indicate the gap between \method{} and the best
prior-augmented selector.
(b) Comparison of prior-only top-$k$, uniform sampling, the strongest existing
selector, and \method{} under the same budgets.}
\label{fig:parity}
\end{figure}

\begin{figure}[t]
\centering
\includegraphics[width=\linewidth]{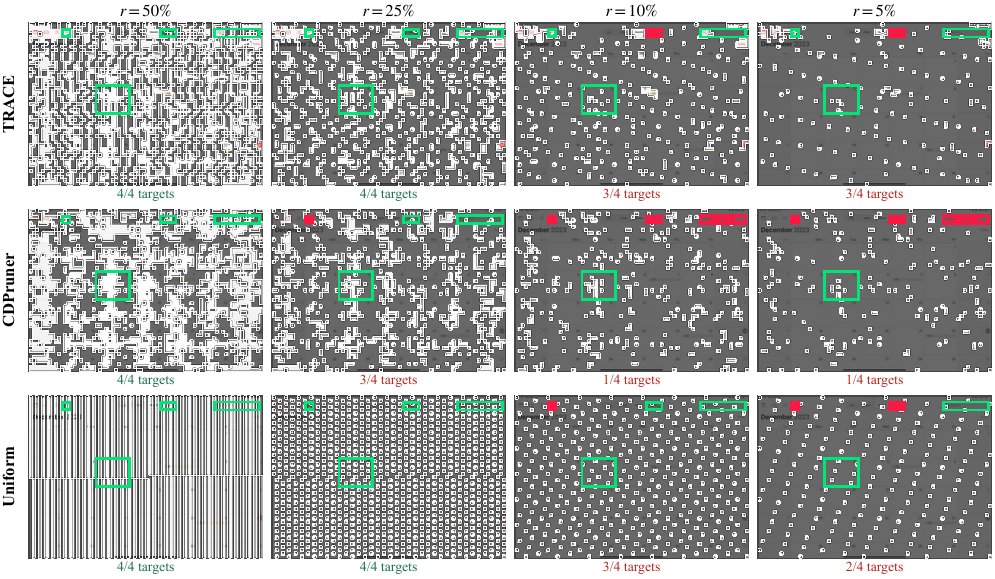}
\caption{Budget tightening on one screen with \textbf{GUI-Owl-1.5-8B}.}
\label{fig:g-budget}
\end{figure}

\begin{figure}[t]
\centering
\includegraphics[width=\linewidth]{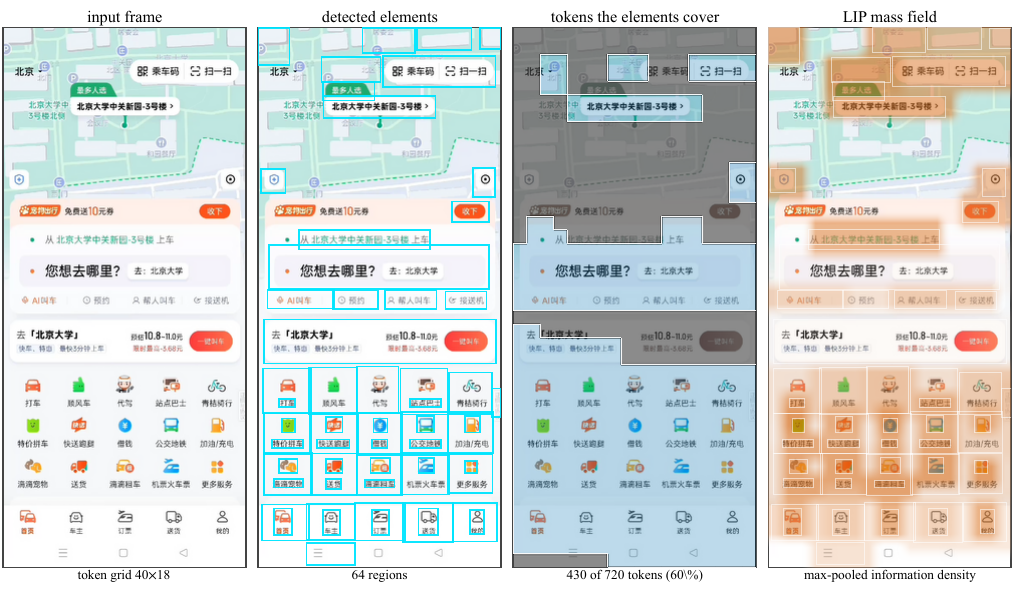}
\caption{From detected elements to a token-level prior with \textbf{GUI-Owl-1.5-8B}.}
\label{fig:g-detector}
\end{figure}

\begin{figure}[t]
\centering
\includegraphics[width=\linewidth]{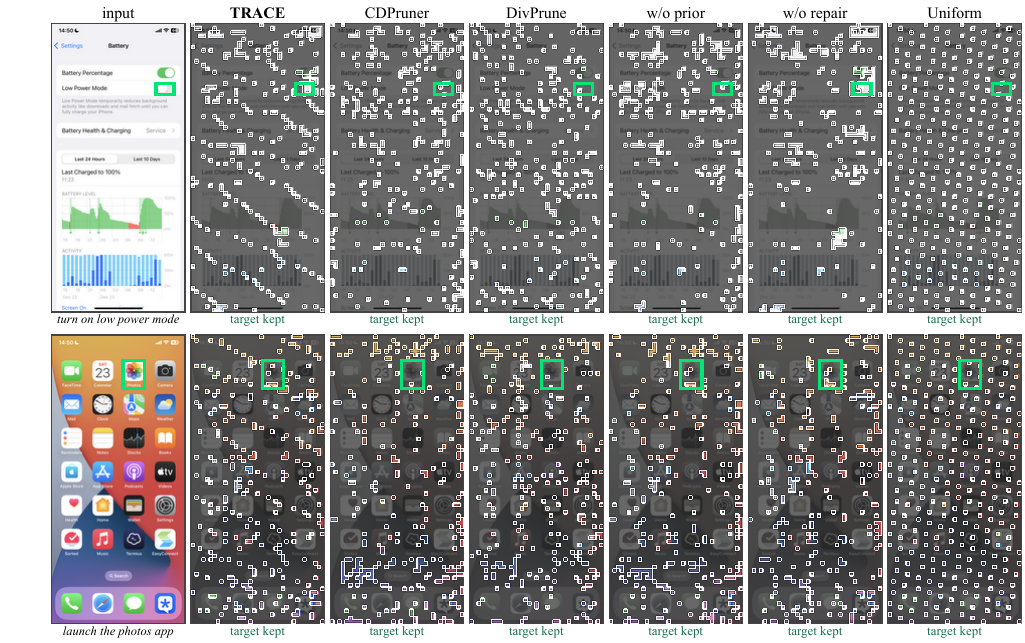}
\caption{Every selector at $\keepr=10\%$ on identical frames with \textbf{GUI-Owl-1.5-8B}.}
\label{fig:g-selectors}
\end{figure}

The benchmark breakdowns identify the settings where \method{} gains or loses accuracy. We next examine how the selection changes under component removal and how those changes appear in the retained patches.
The ablations compare task scores, while the visualizations expose target coverage and the spatial distribution of the keep.
Detailed analysis is provided as follows.
\\
\noindent\textbf{Keep Maps.}
\label{app:viz}
Fig.~\ref{fig:keepmaps} compares three token selections on the same frame at $\keepr=10\%$. The first two panels establish the visual reference. Panel (a) shows the original screen and its target. Panel (b) shows the query-independent \lmp{} field over that screen. The remaining panels show the tokens retained by \method{} and the two controls.
This arrangement connects the layout prior to the resulting selection. The target remains covered by \method{} and is marked HIT. Both the no-prior arm and uniform sampling are marked MISS. The example therefore shows a target-coverage difference at a matched budget.
We next examine how this spatial allocation changes with the budget. Fig.~\ref{fig:keep-maps} in the main text introduces a second screen. Fig.~\ref{fig:g-budget} extends that comparison across four selectors on the same screen. The two views serve complementary purposes. The first identifies which selection covers a given target. The second tracks the retained spatial structure as the budget contracts.
\\
\noindent\textbf{Coverage Recall.}
Area coverage does not distinguish background patches from operable elements. The recall diagnostic behind Fig.~\ref{fig:modules} instead asks whether an element retains at least one token.
The reported recall of \method{} is $90.06\%$ on the $120$-frame dump.
The detector defines the elements used in this diagnostic. The recall therefore measures coverage of its proposals rather than coverage of every true interface element. A surviving token also need not contain enough detail to identify the element correctly.
We therefore use this statistic to interpret the spatial footprint of the keep. Task accuracy remains a separate measure of whether the retained evidence supports a correct prediction.
\\
\noindent\textbf{Coverage Coupling.}
The paired repair comparison associated with Fig.~\ref{fig:modules} records $44$ rescued targets and $5$ lost targets.
Repair reallocates a fixed token budget. It can therefore recover one target at the expense of another. The paired counts expose both outcomes instead of reducing them to an average coverage change.
Recoveries outnumber losses on these frames. The five losses nevertheless show that repair does not benefit every image.
Tab.~\ref{tab:ablation-main} supplies the corresponding task-level comparison with the prior and ordering enabled. Adding repair raises tight-budget ScreenSpot-v2 accuracy from $38.36\%$ to $55.19\%$. The same addition raises tight-budget ScreenSpot-Pro accuracy from $28.15\%$ to $37.63\%$.
Thus, spatial repair recovers more targets on dense GUI screens, and the same addition raises accuracy on both grounding benchmarks under the prior and ordering. 
\\
\noindent\textbf{Ablation Summary.}
Fig.~\ref{fig:ablation-forest} collects the component-removal comparisons. Positive $\Delta$ denotes the accuracy lost when the named component is removed.
The prior-removal deltas on ScreenSpot-v2 grow from $20.05$ points at $\keepr=10\%$ to $23.74$ points at $\keepr=5\%$. On ScreenSpot-Pro, they grow from $13.79$ points at $\keepr=25\%$ to $16.51$ points at $\keepr=10\%$.
The factor knockouts then separate the instruction term from the diversity term. Removing the instruction term costs $15.57$ points on tight ScreenSpot-v2. Removing diversity costs $5.35$ points under the same setting. The corresponding ScreenSpot-Pro losses are $14.17$ and $4.74$ points.
Both terms contribute to the reported score. The larger instruction-term losses show that their contributions are not interchangeable.
The figure also reports a $6.41$-point Mind2Web delta for the removed NCR, demonstrating the contribution of the spatial coverage repair under the multi-step setting.
Together, these results fully demonstrate the effectiveness of each component on the overall performance of our proposed \method{}.
\label{app:mmbench}
\\
\noindent\textbf{MMBench Prior-strength Interaction.}
Tab.~\ref{tab:alpha-sens} reports the prior-strength sweep for MMBench-GUI at $\keepr=25\%$.
The scores are $71.06\%$, $72.15\%$, $72.34\%$, and $72.45\%$ across the four displayed strengths.
The largest improvement occurs before the strongest setting. Increasing $\bar{\alpha}$ from $4$ to $6$ adds only $0.11$ points.
The score increases throughout this sweep. Most of the gain is already present before the final increase in prior strength. This experiment varies prior weighting rather than coverage repair. It therefore does not isolate the contribution of spatial coverage repair.
\\
\noindent\textbf{Layout Prior Comparison.}
\label{app:parity}
Only \method{} uses an external detector in the main tables. This creates an information asymmetry between selectors. The layout prior comparison tests whether giving each selector the same layout prior closes the accuracy gap.
To remove this information asymmetry, we graft the same OmniParser interaction-density field into the official scoring rule of each compatible baseline at its best prior strength.
Each compatible baseline receives its own best prior strength. This gives the baseline a favorable use of the shared signal. An empty detection set reduces each arm to its official rule.
Fig.~\ref{fig:parity} summarizes the layout prior results on ScreenSpot-v2. Tab.~\ref{tab:layout-prior} reports the corresponding accuracies under Own rule and + prior. Best $\alpha$ records the selected strength for each baseline. Transfer measures the change from Own rule to + prior. Gap to \method{} measures how far the + prior score falls below \method{}. Both differences are reported in percentage points (pp).
\begin{table}[t]
    \centering
    \caption{Layout prior comparison with \textbf{GUI-Owl-1.5-8B} on ScreenSpot-v2.}
    \label{tab:layout-prior}
    \adjustbox{max width=\textwidth,center}{%
        \footnotesize
        \renewcommand{\arraystretch}{0.9}
        \begin{tabular}{l c c c c c c}
            \toprule
            \textbf{Selector} & \textbf{Budget} & \textbf{Own rule} & \textbf{+ prior} & \textbf{Best }$\alpha$ & \textbf{Transfer (pp)} & \textbf{Gap to \method{} (pp)} \\
            \midrule
            PruneSID  & $10\%$ & $58.96\%$ & $60.53\%$ & $8$  & $+1.57$  & $13.37$ \\
            PruneSID  & $5\%$  & $34.20\%$ & $34.36\%$ & $8$  & $+0.16$  & $20.83$ \\
            DivPrune  & $10\%$ & $51.97\%$ & $54.95\%$ & $2$  & $+2.98$  & $18.95$ \\
            DivPrune  & $5\%$  & $25.71\%$ & $29.72\%$ & $2$  & $+4.01$  & $25.47$ \\
            VisPruner & $10\%$ & $51.73\%$ & $49.29\%$ & $2$  & $-2.44$  & $24.61$ \\
            VisPruner & $5\%$  & $20.68\%$ & $30.35\%$ & $16$ & $+9.67$  & $24.84$ \\
            CDPruner  & $10\%$ & $33.18\%$ & $47.09\%$ & $2$  & $+13.91$ & $26.81$ \\
            CDPruner  & $5\%$  & $15.88\%$ & $27.99\%$ & $2$  & $+12.11$ & $27.20$ \\
            \midrule
            Prior-only top-$k$ & $10\%$ & \text{N/A} & $32.63\%$ & \text{N/A} & \text{N/A} & $41.27$ \\
            Prior-only top-$k$ & $5\%$ & \text{N/A} & $17.06\%$ & \text{N/A} & \text{N/A} & $38.13$ \\
            \bottomrule
        \end{tabular}
    }
\end{table}

\\
Specifically, Tab.~\ref{tab:layout-prior} shows that the layout prior improves most baselines. The best compatible result nevertheless trails \method{} by $13.37$ points at the mild budget and $20.83$ points at the tight budget.
The different responses also show that the detector prior cannot be treated as a universal replacement for each selector's scoring rule.
The prior-only top-$k$ rows in Tab.~\ref{tab:layout-prior} provide another perspective on the layout prior comparison.
These rows select tokens directly from the prior without evidence ordering. They reach $32.63\%$ at the mild budget and $17.06\%$ at the tight budget. These scores fall below the complete method by $41.27$ and $38.13$ points.
The shared prior improves several selectors, but the remaining gaps show that it does not replace evidence ordering.
\begin{figure}[t]
    \centering
    \includegraphics[width=\linewidth]{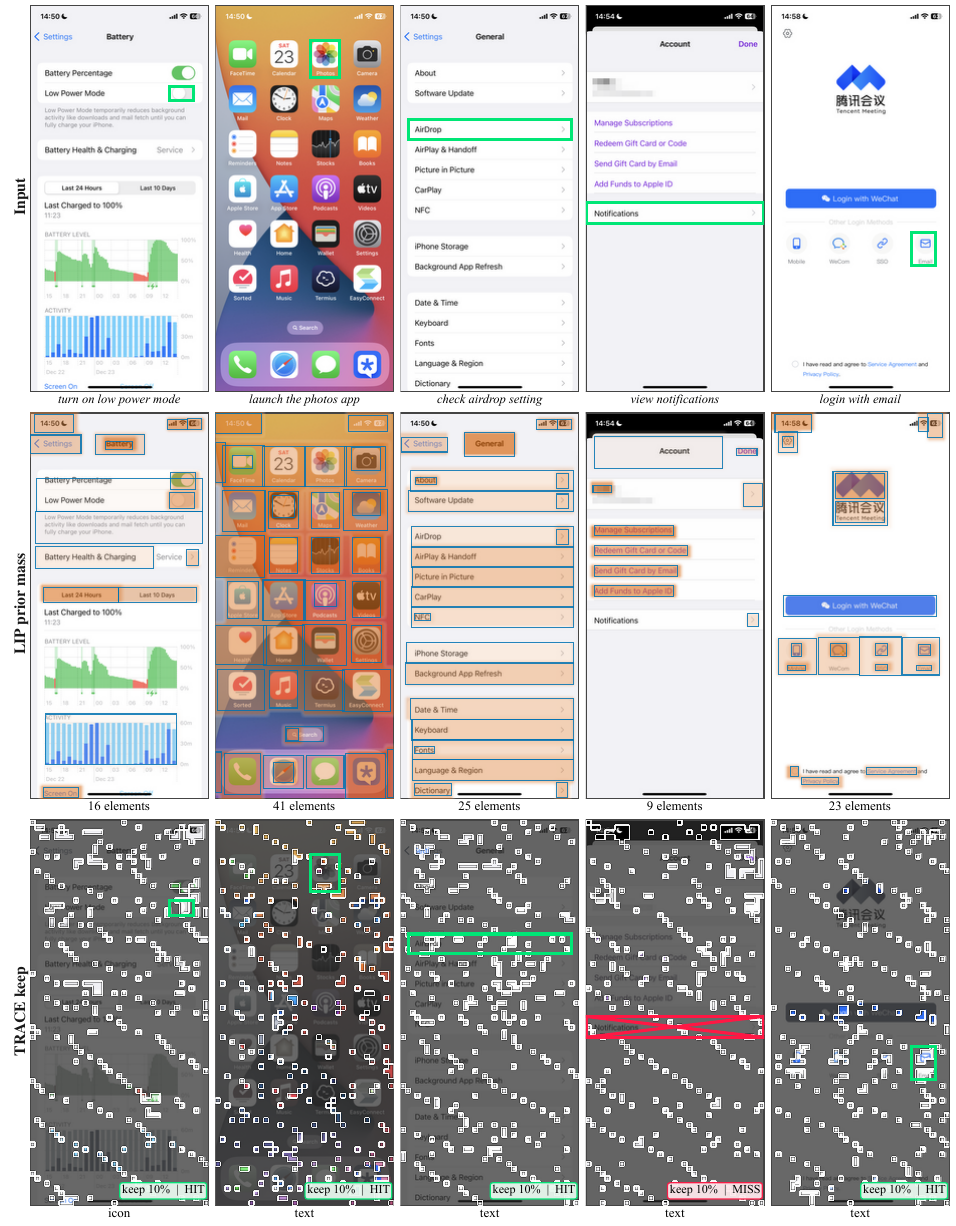}
    \caption{Keep gallery with \textbf{GUI-Owl-1.5-8B} at $\keepr=10\%$.}
    \label{fig:g-gallery}
    \end{figure}
\\
\noindent\textbf{Keep Gallery.}
\label{app:gallery}
Fig.~\ref{fig:g-gallery} presents a five-column by three-row gallery at $\keepr=10\%$. 
From left to right, the columns correspond to Battery Low Power Mode, the home screen target “launch the photos app,” General AirDrop, Account Notifications, and Tencent Meeting email login. 
The top row shows the input screens, the middle row visualizes the detector and its layout prior, and the bottom row shows the retained token field. 
The detector identifies 16, 41, 25, 9, and 23 elements in the five examples, respectively.
Four examples are marked as HIT and one as MISS. 
The Battery example retains the low-power-mode control, while the home-screen example preserves the Photos target. 
The AirDrop example retains the relevant settings row, and the Tencent Meeting example preserves the email-login region. 
The Notifications example is the only MISS because the tiny notification control is not sufficiently represented in the retained tokens. 
This gallery therefore illustrates both the intended behavior and a concrete failure case for small targets rather than suggesting that the layout prior succeeds uniformly.
Across the five examples, the middle row concentrates mass on detected interface regions, while the bottom row distributes native token marks over these regions and their nearby context. 
The reported element counts refer to detector proposals rather than retained-token counts, and the green boxes indicate the requested targets rather than dataset-level accuracy. 
These visualizations show how layout evidence guides token allocation while also exposing an important limitation. 
A detected element may still be too small to support reliable grounding.
\\
\noindent\textbf{Budget Tightening.}
Fig.~\ref{fig:g-budget} holds one screen fixed while the budget tightens from $50\%$ to $5\%$.
The coverage-repaired keep retains a wider spatial skeleton than instruction-conditioned scoring at the same budget.
The target-count annotations make the comparison more specific than the total colored area. At the tightest budget \method{} retains three of the four marked targets. The instruction-conditioned comparison retains one.
The screen and marked targets remain fixed throughout the comparison. The change therefore reflects the allocation of a limited token set rather than a change in screen difficulty. The target counts describe this screen alone and are not dataset-wide accuracy rates.
The comparison shows how spatial coverage changes under matched budgets. The selectors also differ in other components, so the paired ablations provide the task-level evidence for repair.
\\
\noindent\textbf{Detector to Prior.}
Fig.~\ref{fig:g-detector} maps detected boxes onto the token grid before constructing the prior field.
The displayed frame contains $64$ regions on a $40\times18$ grid. Their union covers $430$ of the $720$ tokens, or about $60\%$ of this frame.
The final panel shows a mass field rather than a binary keep mask. Detection assigns spatial weights to the candidate tokens. The ordering and coverage rules then use these weights to select native tokens under the budget.
This distinction matters because detected regions can occupy more tokens than the keep permits. The detector alone does not specify how to allocate the remaining capacity among those regions.
\\
\noindent\textbf{Selector Comparison.}
Fig.~\ref{fig:g-selectors} compares selectors at $\keepr=10\%$ on identical frames and matched token counts.
The columns show \method{}, CDPruner, DivPrune, the no-prior and no-repair arms, and uniform.
Every displayed selection is annotated as target kept. The comparison therefore concerns the surrounding spatial footprint rather than a target-retention advantage. It shows how the rules distribute their remaining capacity after retaining the target.
The component-removal columns expose changes in the distribution of retained patches. Their task-level consequences must be read from the ablation results rather than inferred from the appearance of the keep alone.
Fig.~\ref{fig:parity} complements this visual comparison by holding the detector output fixed. Its remaining accuracy gaps show that access to the same layout field does not make the selection rules equivalent.
\subsection{Serving Efficiency}
\label{app:efficiency}
\begin{figure}[t]
\centering
\includegraphics[width=0.46\linewidth]{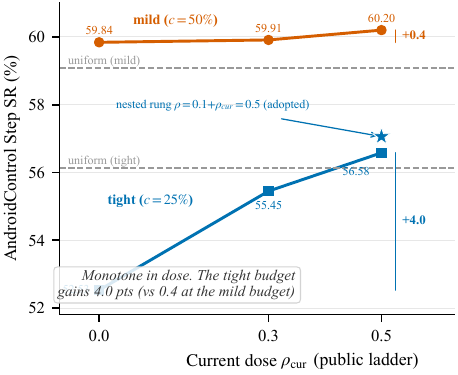}\hfill
\includegraphics[width=0.50\linewidth]{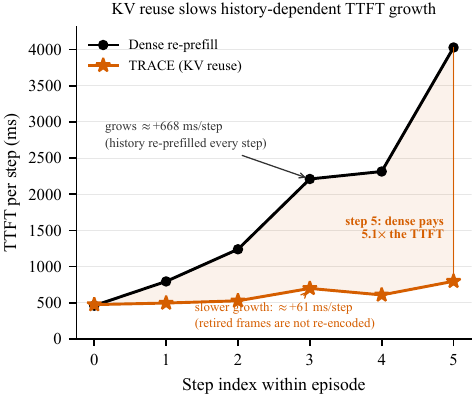}
\caption{Current dose and within-episode TTFT with \textbf{GUI-Owl-1.5-8B}. The left panel shows the evaluation on AndroidControl. The right panel shows the evaluation on OmniGUI.}
\label{fig:dose-ttft}
\end{figure}

This subsection measures the serving benefit of lifecycle reuse and separates it from selector cost.
\\
\noindent\cnum{1}~\textit{End-to-end Serving Result.}
Tab.~\ref{tab:main-efficiency} compares the complete serving path on OmniGUI under the mild budget. 
The dense baseline processes 4,892 visual tokens and 73,385 GFLOPs, resulting in 1,102.6 ms TTFT and 2,958.2 ms end-to-end latency with a Step~SR of 52.45.
The pruned methods reduce the reported input to about 2,011 tokens and 30,170 GFLOPs while reducing the visual KV cache from 697\,MB to 292\,MB. 
Here, In.Tok denotes the input-token count reported by the table and should not be interpreted as the number of visual tokens alone.
The latency ranking reflects the trade-off between selection overhead and the resulting reduction in prefill cost. FastV reports no separate selection cost, yet still uses 38,731 GFLOPs and an In.Tok value of 4,892. 
Its TTFT is 850.7 ms and its end-to-end latency is 2,723.2 ms. Thus, Sel.=0 indicates that no separate selector cost is charged in the table rather than that FastV performs no internal scoring or computation. Among methods that retain 2,011 tokens, TRIM incurs only 10.0 ms of selection time and achieves 721.1 ms TTFT. In contrast, PruneSID spends 740.0 ms on selection, which raises its TTFT to 1,506.9 ms and its end-to-end latency to 3,542.4 ms. ZOO-Prune also incurs 69.5 ms of selection time and reaches 3,015.7 ms end-to-end latency despite achieving a Step~SR of 48.17.
Our method combines a 67.1 ms selection cost with the lowest TTFT of 453.9 ms and the lowest end-to-end latency of 2,295.8 ms. Compared with dense serving, it reduces TTFT by 648.7 ms and end-to-end latency by 662.4 ms while achieving a Step~SR of 48.91, which retains 93.3\% of the dense score. 
This result shows that the serving benefit does not come from pruning alone. 
The method reduces the prefill workload while keeping selection overhead on the critical path substantially below the cost incurred by methods such as PruneSID. 
More broadly, the results indicate that end-to-end serving latency depends on both the cost of selection and the amount of visual state that can be reused after pruning.
Once the selection cost is higher than the prefill benefit, the serving cost instead increases.
\begin{table}[t]
    \centering
    \caption{Efficiency comparison with \textbf{GUI-Owl-1.5-8B} on OmniGUI at the mild budget.}
    \label{tab:main-efficiency}
    \adjustbox{max width=\textwidth,center}{%
        \footnotesize
        \setlength{\tabcolsep}{5pt}
        \renewcommand{\arraystretch}{0.96}
        \begin{tabular}{l cc c cc c cc}
            \toprule
            \textbf{Method} & \textbf{In.Tok} & \textbf{GFLOPs} & \textbf{Sel.\,(ms)} & \textbf{TTFT\,(ms)} & \textbf{E2E\,(ms)} & \textbf{KV\,(MB)} & \textbf{Step~SR} & \textbf{Ret.\,(\%)} \\
            \midrule
            \textcolor{gray}{Baseline (full)} & \textcolor{gray}{4892} & \textcolor{gray}{73385} & \textcolor{gray}{0} & \textcolor{gray}{1102.6} & \textcolor{gray}{2958.2} & \textcolor{gray}{697} & \textcolor{gray}{52.45} & \textcolor{gray}{100.0\%} \\
            DivPrune\pub{(CVPR25)\cite{alvar2025divprune}} & \underline{2011} & \textbf{30170} & 67.0 & 774.1 & 2847.7 & \textbf{292} & 45.76 & 87.2\% \\
            CDPruner\pub{(NeurIPS25)\cite{zhang2025cdpruner}} & \underline{2011} & \textbf{30170} & 149.2 & 847.6 & 2954.7 & \textbf{292} & 41.45 & 79.0\% \\
            VisPruner\pub{(ICCV25)\cite{zhang2025vispruner}} & \underline{2011} & \textbf{30170} & 83.0 & 845.6 & 2954.2 & \textbf{292} & 46.11 & 87.9\% \\
            PruMerge+\pub{(ICCV25)\cite{shang2025prumerge}} & \underline{2011} & \textbf{30170} & 15.8 & 775.7 & 2833.4 & \textbf{292} & 46.50 & 88.7\% \\
            TRIM\pub{(COLING25)\cite{song2025trim}} & \underline{2011} & \textbf{30170} & \underline{10.0} & \underline{721.1} & 2853.6 & \textbf{292} & 45.84 & 87.4\% \\
            FastV\pub{(ECCV24)\cite{chen2024fastv}} & 4892 & \underline{38731} & \textbf{0} & 850.7 & \underline{2723.2} & \underline{315} & 46.89 & 89.4\% \\
            VisionTrim\pub{(ICLR26)\cite{yu2026visiontrim}} & \underline{2011} & \textbf{30170} & 10.7 & 773.5 & 2892.6 & \textbf{292} & 45.72 & 87.2\% \\
            ZOO-Prune\pub{(CVPR26)\cite{kim2026zooprune}} & \underline{2011} & \textbf{30170} & 69.5 & 972.5 & 3015.7 & \textbf{292} & \underline{48.17} & \underline{91.8\%} \\
            PruneSID\pub{(ICLR26)\cite{fang2026prunesid}} & \textbf{1939} & \textbf{30170} & 740.0 & 1506.9 & 3542.4 & \textbf{292} & 47.40 & 90.4\% \\
            \rowcolor{rowblue}\textbf{\method{}} & \underline{2011} & \textbf{30170} & 67.1 & \textbf{453.9} & \textbf{2295.8} & \textbf{292} & \textbf{48.91} & \textbf{93.3\%} \\
            \bottomrule
        \end{tabular}
    }
\end{table}

\\
\noindent\cnum{2}~\textit{Attribution of the Gain.}
Tab.~\ref{tab:efficiency} separates cache reuse from token selection in a second and strictly serial measurement round on the same serving path.
The dense and complete-method TTFT values are $1116.8$ and $452.7$ ms, respectively, rather than the $1102.6$ and $453.9$ ms reported in Tab.~\ref{tab:main-efficiency}. We therefore make all comparisons within the same measurement round.
Applying \kvr{} alone reduces TTFT to $487.4$ ms without pruning the visual cache, so the visual KV entry remains at $100\%$. Adding the complete selector further reduces TTFT to $452.7$ ms at the mild budget and $397.8$ ms at the tight budget. At the same time, visual KV usage decreases by $58.1\%$ and $68.1\%$, respectively.
The LLM prefill time decreases from $328.6$ ms with reuse alone to $140.6$ and $101.9$ ms after selection. The corresponding selector-plus-prior costs are $67.4$ and $47.0$ ms. These measurements expose the two sides of the selection trade-off. Selection introduces additional computation before language-model processing, but it also shortens the visual sequence processed by the language model. Because some component operations overlap on the serving path, their timings cannot be directly summed to predict the final latency. The realized benefit is therefore best evaluated from the measured end-to-end latency.
The prior-plus-ordering arm provides a reference for the incremental cost of repair.
Adding repair changes the mild-budget TTFT from $451.7$ to $452.7$ ms, while the tight-budget TTFT changes from $399.1$ to $397.8$ ms. 
These differences indicate that repair introduces negligible TTFT overhead.
\\
\noindent\cnum{3}~\textit{Accuracy and latency are separate outcomes.}
Tab.~\ref{tab:efficiency} also shows why TTFT and total latency must be distinguished. Tightening the budget reduces full-method TTFT from $452.7$ to $397.8$\ ms. End-to-end latency changes much less from $2298.9$ to $2291.5$\ ms.
The decode totals differ as well, at $1711.0$ and $1765.6$\ ms. A smaller prefill therefore does not imply the same proportional saving in total response time.
Together with the accuracy breakdowns, these measurements characterize a budget trade-off rather than an unqualified preference for the tightest setting.
\begin{table}[t]
    \centering
    \caption{Per-step latency breakdown with \textbf{GUI-Owl-1.5-8B} on OmniGUI, in milliseconds. This breakdown is measured in a separate strictly serial round on the same serving path.}
    \label{tab:efficiency}
    \footnotesize
    \setlength{\tabcolsep}{4.5pt}
    \renewcommand{\arraystretch}{1.0}
    \begin{tabular}{l cc cc cc}
        \toprule
        & & & \multicolumn{2}{c}{\textbf{Mild} ($\keepc{=}50/\keeph{=}10\%$)} & \multicolumn{2}{c}{\textbf{Tight} ($\keepc{=}25/\keeph{=}5\%$)} \\
        \cmidrule(lr){4-5}\cmidrule(lr){6-7}
        \textbf{Component} & Baseline (full) & \kvr{} & \lmp{}+\nwe{} & \method{} full & \lmp{}+\nwe{} & \method{} full \\
        \midrule
        Vision encode + prefill & 649.6 & 263.6 & 324.8 & 329.5 & 305.5 & 305.1 \\
        Pruning (selector + prior) & 0.0 & 0.0 & 60.7 & 67.4 & 40.5 & 47.0 \\
        LLM prefill & 452.6 & 328.6 & 141.9 & 140.6 & 102.6 & 101.9 \\
        \textbf{TTFT} & 1116.8 & 487.4 & 451.7 & \textbf{452.7} & 399.1 & \textbf{397.8} \\
        Decode (total) & 1794.3 & 1810.4 & 1715.8 & 1711.0 & 1846.0 & 1765.6 \\
        \textbf{End-to-end} & 3014.7 & 2399.8 & 2304.7 & \textbf{2298.9} & 2374.9 & \textbf{2291.5} \\
        Visual KV cache & 100\% & 100\% & $-$58.1\% & $-$58.1\% & $-$68.1\% & $-$68.1\% \\
        \bottomrule
    \end{tabular}
\end{table}

\\
\noindent\textbf{Dose and Within-episode Cost.}
The two panels of Fig.~\ref{fig:dose-ttft} report separate measurements on separate benchmarks.
The left panel plots AndroidControl Step~SR against the current-frame dose. The mild-budget scores rise from $59.84\%$ through $59.91\%$ to $60.20\%$. The tight-budget curve reaches $55.45\%$ at dose $0.3$ and $56.58\%$ at dose $0.5$.
The star additionally enables the nested history rung and should not be read as another point on the current-dose-only curve. This separates the current-frame dose response from the adopted joint setting.
The right panel follows OmniGUI TTFT across successive steps within one episode. It does not vary the history-window length independently.
Both paths start near the same latency. The dense curve then rises much more sharply as steps accumulate. Its TTFT reaches $5.1\times$ that of \method{} at step $5$.
The widening gap is consistent with avoiding repeated encoding of retired frames. Reuse does not make latency constant. Its curve also rises overall and fluctuates between individual steps.
This visualization demonstrates the importance of cache reuse.
\section{Transfer and Scope}
\label{app:transfer}
In this appendix, we provide the final evidence containing two transfer experiments followed by an analysis of scope and failure modes. 
Specifically, we first place existing selectors on the nested \kvr{} path to match their serving state. We then transfer \method{} to UI-TARS-1.5-7B with its native prompt and action grammar, without retuning the 8B configuration.
Details are provided below.
\\
\noindent\textbf{Nested-path Transfer.}
Tab.~\ref{tab:transfer} places DivPrune, VisPruner, and ZOO-Prune on the same nested \kvr{} path at matched budgets. \method{} remains ahead by $0.88$--$9.62$ points across the reported settings.
All methods now use the same reusable cache path.
The differences therefore reflect the visual evidence admitted by their selection rules under matched serving conditions.
These results fully demonstrate the effectiveness and efficiency of \method{} on the nested \kvr{} path.

\begin{table}[t]
    \centering
    \caption{Paradigm-transfer comparison with \textbf{GUI-Owl-1.5-8B}.}
    \label{tab:transfer}
    \adjustbox{max width=0.92\textwidth,center}{%
        \footnotesize
        \setlength{\tabcolsep}{6pt}
        \renewcommand{\arraystretch}{0.98}
        \begin{tabular}{ll cc cc}
            \toprule
            & & \multicolumn{2}{c}{\textbf{OmniGUI (Step~SR)}} & \multicolumn{2}{c}{\textbf{Mind2Web (Step~SR)}} \\
            \cmidrule(lr){3-4}\cmidrule(lr){5-6}
            \textbf{Selector} & \textbf{Serving path} & mild & tight & mild & tight \\
            \midrule
            \rowcolor{rowblue}\textbf{\method{}} & \kvr{} (ours) & \textbf{48.91} & \textbf{43.08} & \textbf{45.52} & \textbf{34.20} \\
            \midrule
            DivPrune & native re-prefill & 45.76 & 37.17 & 40.69 & 24.28 \\
            & \kvr{} (ported) & 46.27 & 35.58 & 40.51 & 24.58 \\
            & \emph{margin of \method{}} & $+2.64$ & $+7.50$ & $+5.01$ & $+9.62$ \\
            \midrule
            VisPruner & native re-prefill & 46.11 & 39.54 & 44.45 & 30.67 \\
            & \kvr{} (ported) & 46.89 & 39.85 & 44.57 & 30.66 \\
            & \emph{margin of \method{}} & $+2.02$ & $+3.23$ & $+0.95$ & $+3.54$ \\
            \midrule
            ZOO-Prune & native re-prefill & 48.17 & 40.01 & 44.98 & 27.90 \\
            & \kvr{} (ported) & 47.51 & 39.42 & 44.64 & 28.31 \\
            & \emph{margin of \method{}} & $+1.40$ & $+3.66$ & $+0.88$ & $+5.89$ \\
            \bottomrule
        \end{tabular}
    }
\end{table}

\noindent\textbf{Transfer to UI-TARS.}
\label{app:uitars}
Tab.~\ref{tab:uitars-full} evaluates \method{} with UI-TARS's native prompt and action grammar. At the reported retention settings, \method{} leads ScreenSpot-Pro at $25\%$ and $5\%$ with average accuracies of $16.76\%$ and $3.04\%$. It also leads ScreenSpot-v2 at $25\%$ and $10\%$. The ranking is not uniform across benchmarks. DivPrune remains ahead on MMBench-GUI, while PruneSID leads several higher-retention settings. The transfer advantage is clearest at tighter budgets.
\begin{table}[t]
    \centering
    \caption{Complete single-step results with \textbf{UI-TARS-1.5-7B}.}
    \label{tab:uitars-full}
    \adjustbox{max width=\textwidth,max totalheight=0.70\textheight,center}{%
        \scriptsize
        \setlength{\tabcolsep}{4.5pt}
        \renewcommand{\arraystretch}{0.96}
        \begin{tabular}{l ccc ccc ccc c}
            \toprule
            & \multicolumn{3}{c}{\textbf{ScreenSpot-v2}} & \multicolumn{3}{c}{\textbf{ScreenSpot-Pro}} & \multicolumn{3}{c}{\textbf{MMBench-GUI L2}} & \\
            \cmidrule(lr){2-4}\cmidrule(lr){5-7}\cmidrule(lr){8-10}
            \textbf{Method} & Text & Icon & \textbf{Avg.} & Text & Icon & \textbf{Avg.} & Basic & Adv. & \textbf{Avg.} & \textbf{Ret.\,(\%)} \\
            \midrule
            \rowcolor{headergray}\multicolumn{11}{c}{\textsc{UI-TARS-1.5-7B: Upper Bound (100\% Tokens)}} \\
            UI-TARS-1.5-7B & 93.73 & 84.84 & 89.86 & 57.42 & 17.55 & 42.19 & 85.06 & 62.59 & 73.76 & 100.0\% \\
            \rowcolor{headergray}\multicolumn{11}{c}{\textsc{Retain 50\% Tokens ($\downarrow$\,50\%)}} \\
            Random & 81.34 & 73.10 & 77.75 & 31.53 & 9.44 & 23.09 & 66.87 & 45.82 & 56.29 & 72.5\% \\
            Uniform & 83.84 & 73.65 & 79.40 & 34.39 & 8.77 & 24.60 & 68.33 & 46.43 & 57.32 & 74.8\% \\
            DivPrune\pub{(CVPR25)\cite{alvar2025divprune}} & \textbf{90.11} & 81.23 & 86.24 & \underline{44.93} & 14.40 & \underline{33.27} & \underline{75.49} & 53.46 & \underline{64.41} & \underline{87.4\%} \\
            CDPruner\pub{(NeurIPS25)\cite{zhang2025cdpruner}} & 85.65 & 74.37 & 80.74 & 34.49 & 11.42 & 25.68 & 64.35 & 45.88 & 55.06 & 75.1\% \\
            TRIM\pub{(COLING25)\cite{song2025trim}} & 77.58 & 74.01 & 76.02 & 29.17 & 10.60 & 22.07 & 66.70 & 40.18 & 53.37 & 69.8\% \\
            PruneSID\pub{(ICLR26)\cite{fang2026prunesid}} & \underline{89.00} & \textbf{83.03} & \underline{86.40} & \textbf{46.06} & \textbf{15.07} & \textbf{34.22} & \textbf{78.12} & \textbf{53.79} & \textbf{65.89} & \textbf{88.9\%} \\
            \rowcolor{rowblue}\textbf{\method{}} & \textbf{90.11} & \underline{82.67} & \textbf{86.87} & 42.89 & \underline{14.57} & 32.07 & 73.64 & \underline{53.51} & 63.52 & 86.3\% \\
            \rowcolor{headergray}\multicolumn{11}{c}{\textsc{Retain 25\% Tokens ($\downarrow$\,75\%)}} \\
            Random & 52.51 & 46.93 & 50.08 & 9.01 & 4.14 & 7.15 & 33.74 & 22.58 & 28.13 & 36.9\% \\
            Uniform & 53.06 & 43.14 & 48.74 & 5.53 & 2.81 & 4.49 & 29.55 & 21.42 & 25.46 & 33.1\% \\
            DivPrune\pub{(CVPR25)\cite{alvar2025divprune}} & \textbf{72.28} & 66.79 & \underline{69.89} & \underline{18.53} & \textbf{8.94} & \underline{14.86} & \textbf{50.64} & \underline{34.31} & \textbf{42.43} & \underline{56.8\%} \\
            CDPruner\pub{(NeurIPS25)\cite{zhang2025cdpruner}} & 64.35 & 53.07 & 59.43 & 16.68 & 6.79 & 12.90 & 36.54 & 25.51 & 31.00 & 46.2\% \\
            TRIM\pub{(COLING25)\cite{song2025trim}} & 47.35 & 51.08 & 48.98 & 8.70 & 4.64 & 7.15 & 38.11 & 19.76 & 28.88 & 36.9\% \\
            PruneSID\pub{(ICLR26)\cite{fang2026prunesid}} & \underline{69.08} & \textbf{68.59} & 68.87 & 16.68 & 8.11 & 13.41 & \underline{49.52} & 31.60 & 40.51 & 54.4\% \\
            \rowcolor{rowblue}\textbf{\method{}} & \textbf{72.28} & \underline{67.69} & \textbf{70.28} & \textbf{21.70} & \underline{8.77} & \textbf{16.76} & 46.84 & \textbf{35.81} & \underline{41.29} & \textbf{58.0\%} \\
            \rowcolor{headergray}\multicolumn{11}{c}{\textsc{Retain 10\% Tokens ($\downarrow$\,90\%)}} \\
            Random & 17.41 & 15.52 & 16.59 & 2.05 & 1.49 & 1.83 & 10.02 & 6.53 & 8.26 & 11.3\% \\
            Uniform & 19.50 & 16.06 & 18.00 & 1.94 & 1.32 & 1.71 & 10.52 & 6.14 & 8.32 & 11.8\% \\
            DivPrune\pub{(CVPR25)\cite{alvar2025divprune}} & \underline{36.07} & \textbf{37.73} & \underline{36.79} & \underline{4.71} & \underline{1.99} & \underline{3.67} & \textbf{19.19} & \textbf{12.51} & \textbf{15.83} & \underline{23.7\%} \\
            CDPruner\pub{(NeurIPS25)\cite{zhang2025cdpruner}} & 31.20 & 28.52 & 30.03 & 4.30 & 1.82 & 3.35 & 13.21 & 8.58 & 10.88 & 18.7\% \\
            TRIM\pub{(COLING25)\cite{song2025trim}} & 12.67 & 21.84 & 16.67 & 2.56 & 1.16 & 2.02 & 12.53 & 5.15 & 8.82 & 11.8\% \\
            PruneSID\pub{(ICLR26)\cite{fang2026prunesid}} & 32.73 & \underline{35.38} & 33.88 & 3.58 & 1.49 & 2.78 & \underline{19.03} & 12.23 & \underline{15.61} & 21.8\% \\
            \rowcolor{rowblue}\textbf{\method{}} & \textbf{37.47} & \textbf{37.73} & \textbf{37.58} & \textbf{8.39} & \textbf{2.65} & \textbf{6.20} & 16.40 & \underline{12.45} & 14.41 & \textbf{25.4\%} \\
            \rowcolor{headergray}\multicolumn{11}{c}{\textsc{Retain 5\% Tokens ($\downarrow$\,95\%)}} \\
            Random & 7.38 & 6.86 & 7.15 & 0.72 & 0.33 & 0.57 & 4.70 & 3.10 & 3.90 & 4.9\% \\
            Uniform & 7.52 & 7.40 & 7.47 & 0.92 & 0.33 & 0.70 & 5.04 & 2.88 & 3.95 & 5.1\% \\
            DivPrune\pub{(CVPR25)\cite{alvar2025divprune}} & \underline{15.88} & \textbf{20.40} & \textbf{17.85} & \underline{2.35} & \underline{0.99} & \underline{1.83} & \textbf{7.83} & \underline{5.20} & \textbf{6.51} & \underline{11.0\%} \\
            CDPruner\pub{(NeurIPS25)\cite{zhang2025cdpruner}} & 15.18 & 15.34 & 15.25 & 1.23 & 0.83 & 1.08 & 6.44 & 4.48 & 5.45 & 9.0\% \\
            TRIM\pub{(COLING25)\cite{song2025trim}} & 5.99 & 9.75 & 7.63 & 0.82 & 0.33 & 0.63 & 5.20 & 1.66 & 3.42 & 4.9\% \\
            PruneSID\pub{(ICLR26)\cite{fang2026prunesid}} & 11.00 & \underline{19.13} & 14.54 & 1.64 & 0.66 & 1.27 & \underline{7.39} & 3.87 & 5.62 & 8.9\% \\
            \rowcolor{rowblue}\textbf{\method{}} & \textbf{16.57} & 18.23 & \underline{17.30} & \textbf{3.99} & \textbf{1.49} & \textbf{3.04} & 6.10 & \textbf{5.81} & \underline{5.95} & \textbf{11.5\%} \\
            \bottomrule
        \end{tabular}
    }
\end{table}

\\
\noindent\textbf{Configuration Check.}
Tab.~\ref{tab:uitars-cal} checks the transferred configuration with the 8B protocol. The scores change by $0.13$--$0.63$ points, and the method rankings remain unchanged. The comparison is therefore stable without retuning the full configuration, which demonstrates the transferability.
\begin{table}[t]
    \centering
    \caption{Configuration selection with \textbf{UI-TARS-1.5-7B} over the declared grid $\bar{\alpha}\in\{2,4\}$ and current dose $\rhocur\in\{.10,.30,.50\}$, with the selected setting in \textbf{bold}.}
    \label{tab:uitars-cal}
    \adjustbox{max width=\textwidth,center}{%
        \footnotesize
        \setlength{\tabcolsep}{5pt}
        \renewcommand{\arraystretch}{0.98}
        \begin{tabular}{ll cccc c}
            \toprule
            \textbf{Benchmark} & \textbf{Budget} & $\bar{\alpha}{=}2,\rhocur{=}.10$ & $\bar{\alpha}{=}2,\rhocur{=}.30$ & $\bar{\alpha}{=}2,\rhocur{=}.50$ & $\bar{\alpha}{=}4,\rhocur{=}.10$ & \textbf{Selected} \\
            \midrule
            MMB-L2 & 5\% & \textbf{5.95} & 5.73 & 5.70 & 5.29 & $\bar{\alpha}{=}2,\rhocur{=}.10$ \\
            MMB-L2 & 10\% & \textbf{14.41} & 13.80 & 13.61 & 13.97 & $\bar{\alpha}{=}2,\rhocur{=}.10$ \\
            MMB-L2 & 25\% & 41.04 & 41.04 & 38.81 & \textbf{41.29} & $\bar{\alpha}{=}4,\rhocur{=}.10$ \\
            MMB-L2 & 50\% & 63.27 & 63.47 & \textbf{63.52} & 63.05 & $\bar{\alpha}{=}2,\rhocur{=}.50$ \\
            SS-Pro & 50\% & 31.94 & \textbf{32.07} & 31.69 & 31.94 & $\bar{\alpha}{=}2,\rhocur{=}.30$ \\
            SS-v2 & 5\% & 16.67 & 17.06 & 13.84 & \textbf{17.30} & $\bar{\alpha}{=}4,\rhocur{=}.10$ \\
            \bottomrule
        \end{tabular}
    }
\end{table}

\\
\noindent\textbf{Multi-step Transfer.}
\label{app:uitars-multi}
Every arm uses UI-TARS's native action grammar on the adopted \kvr{} path. On the $1521$ of $2572$ OmniGUI steps with target locations, \method{} gains $1.05$ and $1.18$ points over the strongest baseline across the two budgets. On pooled Mind2Web, the margins are $2.80$ and $4.92$ points.
The gains appear on both the mobile OmniGUI actions and the web Mind2Web actions. All methods receive the same visual-context scope, while \method{} reuses cached rows from prior frames instead of re-encoding them. The transfer therefore preserves the same evidence-selection and cache-reuse comparison under a different backbone and action grammar.
\begin{table}[t]
    \centering
    \caption{Complete multi-step transfer with \textbf{UI-TARS-1.5-7B}.}
    \label{tab:uitars-multi}
    \adjustbox{max width=\textwidth,center}{%
        \footnotesize
        \setlength{\tabcolsep}{5pt}
        \renewcommand{\arraystretch}{0.98}
        \begin{tabular}{l ccc cc c}
            \toprule
            & \multicolumn{3}{c}{\textbf{OmniGUI}} & \multicolumn{2}{c}{\textbf{Mind2Web}} & \\
            \cmidrule(lr){2-4}\cmidrule(lr){5-6}
            \textbf{Method} & Type & Ground. & \textbf{Step~SR} & Ele.\,Acc & \textbf{Step~SR} & \textbf{Ret.\,(\%)} \\
            \midrule
            \rowcolor{headergray}\multicolumn{7}{c}{\textsc{UI-TARS-1.5-7B: Upper Bound (100\% Tokens)}} \\
            UI-TARS-1.5-7B & 79.22 & 62.82 & 49.77 & 53.23 & 42.94 & 100.0\% \\
            \rowcolor{headergray}\multicolumn{7}{c}{\textsc{Mild ($\keepc{=}50\%$, $\keeph{=}10\%$)}} \\
            Random & 77.38 & 49.11 & 38.00 & 35.89 & 27.69 & 70.4\% \\
            Uniform & \textbf{78.90} & 51.00 & 40.24 & 33.60 & 25.98 & 70.7\% \\
            DivPrune\pub{(CVPR25)\cite{alvar2025divprune}} & 77.32 & \underline{54.93} & \underline{42.47} & \underline{41.66} & \underline{32.33} & \underline{80.3\%} \\
            CDPruner\pub{(NeurIPS25)\cite{zhang2025cdpruner}} & 76.13 & 47.06 & 35.83 & 35.59 & 27.16 & 67.6\% \\
            PruneSID\pub{(ICLR26)\cite{fang2026prunesid}} & 76.59 & 54.25 & 41.55 & 39.99 & 31.21 & 78.1\% \\
            \rowcolor{rowblue}\textbf{\method{}} & \underline{78.44} & \textbf{55.49} & \textbf{43.52} & \textbf{44.11} & \textbf{35.13} & \textbf{84.6\%} \\
            \rowcolor{headergray}\multicolumn{7}{c}{\textsc{Tight ($\keepc{=}25\%$, $\keeph{=}5\%$)}} \\
            Random & 74.10 & 29.46 & 21.83 & 11.40 & 8.55 & 31.9\% \\
            Uniform & 76.00 & 30.45 & 23.14 & 8.46 & 6.51 & 30.8\% \\
            DivPrune\pub{(CVPR25)\cite{alvar2025divprune}} & \underline{77.12} & \underline{37.08} & \underline{28.60} & \underline{17.53} & \underline{13.00} & \underline{43.9\%} \\
            CDPruner\pub{(NeurIPS25)\cite{zhang2025cdpruner}} & 72.78 & 32.70 & 23.80 & 16.42 & 12.13 & 38.0\% \\
            PruneSID\pub{(ICLR26)\cite{fang2026prunesid}} & 74.36 & 33.16 & 24.65 & 14.68 & 10.88 & 37.4\% \\
            \rowcolor{rowblue}\textbf{\method{}} & \textbf{78.90} & \textbf{37.75} & \textbf{29.78} & \textbf{22.32} & \textbf{17.92} & \textbf{50.8\%} \\
            \bottomrule
        \end{tabular}
    }
\end{table}

\begin{figure}[t]
    \centering
    \includegraphics[width=0.7\linewidth]{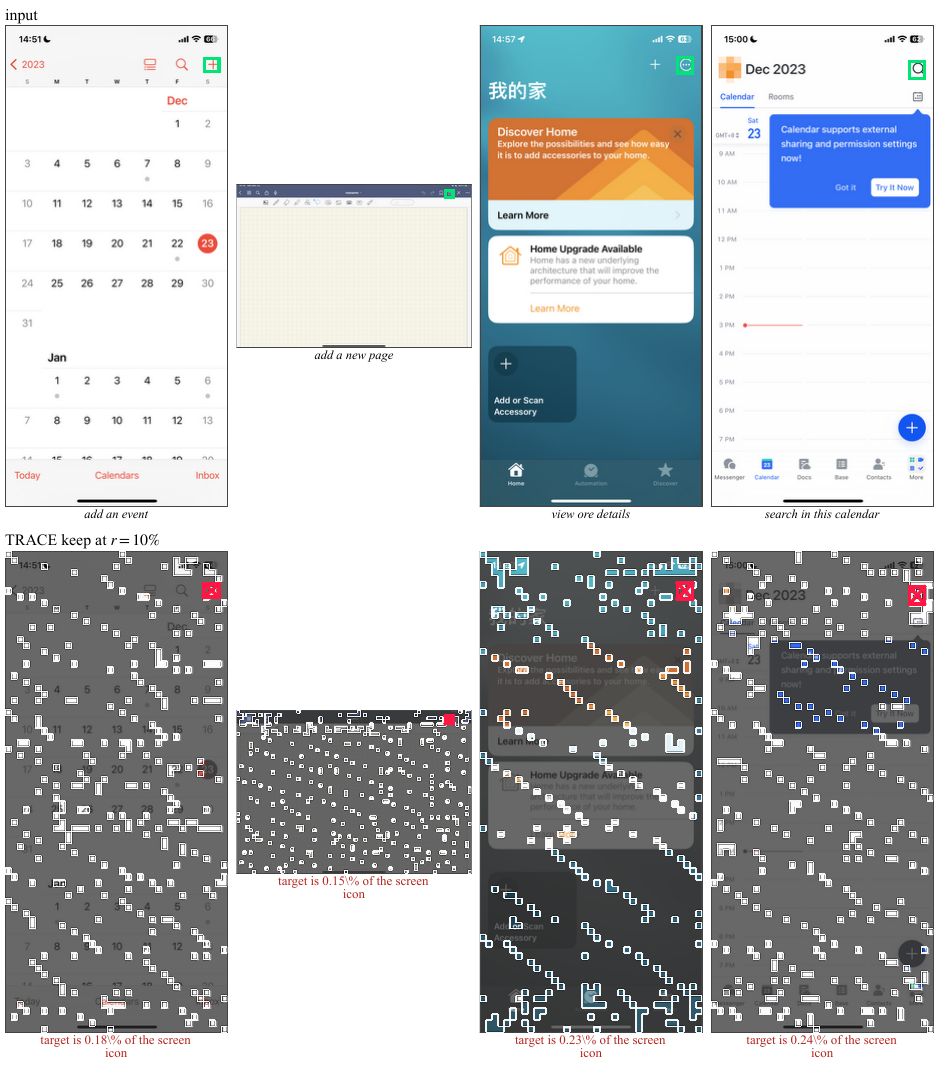}
    \caption{Remaining failure cases with \textbf{GUI-Owl-1.5-8B}.}
    \label{fig:g-failures}
    \end{figure}
\subsection{Limitations and Failure Modes}
\label{app:limits}
In this section, we discuss the trajectory horizon and failure modes of our proposed \method{}.
\\
\noindent\textbf{Trajectory Horizon.}
The evaluated episodes are short relative to the history horizon supported by our method. Most existing GUI benchmarks retain at most two or three historical frames, whereas we extend this horizon to five or six frames. As $H$ grows, the benefit of compact historical evidence reuse becomes increasingly pronounced, as reflected by the serving-cost expression $\mathcal{O}(k_c+Hk_h)$.
However, existing benchmarks rarely require evidence retrieval across long frame intervals. Most tasks can be solved from recent screens, which limits their ability to expose the advantage of longer history reuse. Evaluating this regime requires trajectories with larger temporal gaps between evidence and query screens. 
We leave this systematic evaluation on longer trajectories to future work.
\\
\noindent\textbf{Failure Modes.}
On the recorded ScreenSpot-v2 run at $\keepr=5\%$, \method{} misses $503$ examples that the unpruned model solves. In $93.0\%$ of these failures, a retained token overlaps the target. Removing the target's spatial evidence accounts for only $0.8\%$ of the gap to dense.
The remaining errors occur mainly when the selector retains the target neighborhood but not enough detail for precise localization. The failure is therefore a precision problem under aggressive context reduction rather than a simple failure to reach the target region.
Besides the aforementioned precision problem, Fig.~\ref{fig:g-failures} shows thin or low-contrast widgets whose extent fits within a single patch. These widgets remain difficult to separate from the background after the budget is reduced. For this subset, increasing patch detail or improving the visual representation is more direct than token pruning.

\end{document}